\pdfoutput=1
\documentclass[11pt]{article}

\usepackage{acl} % [review]
\usepackage{subcaption}

\usepackage{times}
\usepackage{latexsym}
\usepackage{hhline}

\usepackage[T1]{fontenc}
\usepackage[utf8]{inputenc}

\usepackage{microtype}
\usepackage[utf8]{inputenc}
\usepackage{graphicx}

\usepackage{hyperref}
\hypersetup{
colorlinks   = true,
citecolor    = blue,
urlcolor    =  blue
}

\usepackage{url}
\usepackage{cleveref}
\crefformat{footnote}{#2\footnotemark[#1]#3}

\newcommand{\footURL}[1]{\footnote{\url{#1}}}

\usepackage{ragged2e}

\usepackage[
singlelinecheck=false % <-- important
]{caption}
\usepackage{multirow}
\usepackage{hhline}
\usepackage{subcaption}

\usepackage{ftnxtra}
\usepackage{fnpos}

\makeatletter
\newread\pin@file
\newcounter{pinlineno}
\newcommand\pin@accu{}
\newcommand\pin@ext{pintmp}
\newcommand*\partialinput [3] {%
  \IfFileExists{#3}{%
    \openin\pin@file #3
    \setcounter{pinlineno}{1}
    \@whilenum\value{pinlineno}<#1 \do{%
      \read\pin@file to\pin@line
      \stepcounter{pinlineno}%
    }
    \addtocounter{pinlineno}{-1}
    \let\pin@accu\empty
    \begingroup
    \endlinechar\newlinechar
    \@whilenum\value{pinlineno}<#2 \do{%
      \readline\pin@file to\pin@line
      \edef\pin@accu{\pin@accu\pin@line}%
      \stepcounter{pinlineno}%
    }
    \closein\pin@file
    \expandafter\endgroup
    \scantokens\expandafter{\pin@accu}%
  }{%
    \errmessage{File `#3' doesn't exist!}%
  }%
}
\makeatother

\usepackage{pdflscape}
\usepackage{svg}

\usepackage{multirow}
\usepackage{pgfplots} 

\usepackage{caption}
\usepackage{tabularx}
\newcolumntype{Y}{>{\centering\arraybackslash}X}
\newcolumntype{Z}{>{\raggedleft\arraybackslash}X}

\usepackage{diagbox}
\usepackage{float}
\usepackage{dblfloatfix}    % To enable figures at the bottom of page
\usepackage{placeins}
\usepackage{makecell}
\usepackage{longtable}
\usepackage{colortbl}

\newcommand{\half}{0.485}
\newcommand{\halfSingle}{0.48}

\newcommand{\fourth}{0.245}
\newcommand{\fourthSingle}{0.23}
\newcommand{\fifth}{0.195}
\newcommand{\sixth}{0.1565}
\newcommand{\tabComp}{0.96}
\newcommand{\tabSmallComp}{0.94}

\newcommand\blfootnote[1]{%
  \begingroup
  \renewcommand\thefootnote{}\footnote{#1}%
  \addtocounter{footnote}{-1}%
  \endgroup
}

\title{\textit{Some Languages are More Equal than Others}: Probing Deeper into the Linguistic Disparity in the NLP World}

 \author{Surangika Ranathunga \and Nisansa de Silva \\
        Department of Computer Science and Engineering \\ University of Moratuwa, Katubedda 10400, Sri Lanka\\
        \texttt{\{surangika, nisansaDds\}@cse.mrt.ac.lk}
        }
        
\begin{document}
\maketitle
\begin{abstract}
Linguistic disparity in the NLP world is a problem that has been widely acknowledged recently. However, different facets of this problem, or the reasons behind this disparity are seldom discussed within the NLP community. This paper provides a comprehensive analysis of the disparity that exists within the languages of the world. We show that simply categorising languages considering data availability may not be always correct. Using an existing language categorisation based on speaker population and vitality, we analyse the distribution of language data resources, amount of NLP/CL research, inclusion in multilingual web-based platforms and the inclusion in pre-trained multilingual models.  We show that many languages do not get covered in these resources or platforms, and even within the languages belonging to the same language group, there is wide disparity.  We analyse the impact of family, geographical location, GDP and the speaker population of languages and provide possible reasons for this disparity, along with some suggestions to overcome the same.
\end{abstract}
\section{Introduction}
Even after more than fifty years since the inception of the fields of Computational Linguistics (CL) and Natural Language Processing (NLP), we still observe a significant bias favouring the so-called \textit{high-resource} languages in the field.\blfootnote{\noindent The paper title is inspired by the quote ``\textit{All animals are equal, but some animals are more equal than others}'' by~\citet{orwell1945animal} which satirically alludes to disparities that exist in places which, ostensibly are supposed to be homogeneous. In this paper, we discuss how the same phenomenon is observed in the broadly used language categorisation systems.} Conversely, this  means that the majority of the 6500+ languages in the world, which have been classified as \textit{low-resource}, have received limited to no attention. This resource poverty is not merely an academic or theoretical issue. It impacts the lives and the well-being of people concerned in a very present and practical manner, and deprives the speakers of low-resource languages from reaping the benefits of NLP in areas such as healthcare~\cite{perez2020upstage}, disaster response~\cite{ray2019keyphrase}, law~\cite{ratnayaka-etal-2020-effective}, and education~\cite{taghipour-ng-2016-neural}.

%There is newfound hope for emergence from obscurity, as
% On the positive side, 
This digital divide between high-resource and low-resource languages (LRLs)\footnote{An LRL is also known as under resourced, low-density, resource-poor, low data, or less-resourced language \cite{besacier2014automatic}} has been brought into the spotlight by many scholars in the field~\cite{bender2019rule, cains2019Geographic, joshi-etal-2020-state, anastasopoulos-etal-2020-endangered}. Consequently, there have been efforts to build data sets covering low-resource languages~\cite{conneau-etal-2018-xnli,ebrahimi-etal-2022-americasnli}, benchmarks~\cite{hu2020xtreme} and techniques that favour low-resource languages~\cite{schwartz-etal-2019-bootstrapping}; all of which, are very promising developments.  However, the problem is not fully solved, and  this disparity should be quantified to understand the gravity of the problem~\cite{khanuja2022evaluating}. Such an understanding is the first step in developing solutions to solve the problem~\cite{grutzner-zahn-rehm-2022-introducing}. % much more to be done. In doing so, having a clear idea of the disparity that exists between the languages in the world with respect to resource availability and other socio-economic conditions is helpful. 

%The `\textit{resourcefulness}' of a language can be analysed with respect to different socio-linguistic aspects.  \citet{besacier2014automatic} identify these factors as: 1)~The existence of a unique writing system, 2)~The amount of presence on the World Wide Web, 3)~The availability of linguistic expertise, and/or 4)~The availability of electronic resources such as corpora (monolingual and parallel), and vocabulary lists.~\citet{singh2008natural}, on the other-hand,  identifies these factors as: 1) The amount of linguistic study, 2) The availability of language resources, 3) The level of computerisation, 4) The availability of language processing tools, and 5) other privileges such as finance and human resource.
%However, we are of the view that for a particular language to come out of its low-resource status, a major role should be played by the researchers who use, or are familiar with that language. To this end, the first step would be to identify the current state of a given low-resource language. 

NLP researchers have mainly considered the availability of electronic data resources as the main descriptor of `\textit{resourcefulness}' of languages. For example, \citet{joshi-etal-2020-state} considered the availability of annotated and raw corpora.~\citet{hedderich-etal-2021-survey}  considered the availability of auxiliary resources such as lexicons in addition.~\citet{faisal-etal-2022-dataset} estimated the level of language speaker representation in dataset content.~\citet{joshi-etal-2020-state} used their criterion to categorise 2485 languages into six groups, based on the availability of unannotated data (number of Wikipedia pages) and the number of annotated  datasets available in the LDC\footnote{\url{https://catalog.ldc.upenn.edu/}} and ELRA\footnote{\url{http://catalog.elra.info/en-us/}}  data repositories. %Figure~\ref{fig:joshiWithoutHugging} shows a recreation of these language categories\footnote{Refer Appendix~\ref{sec:JoshiClass} for class descriptions.}.

%\begin{figure}[!hbt]
%     \centering
%    \includegraphics[width=\half\textwidth]{images/classes_without_HuggingFace_AreaMarked.png}
%    \caption{Recreation of Classes in~\citet{joshi-etal-2020-state}}
%    \label{fig:joshiRe}
%\end{figure}

%According to this categorisation, an astounding 2191 languages fall into \textit{Category 0}- those that have exceptionally low amount of resources. This paints a very grim picture of the linguistic diversity and inclusion in the NLP world. This is not surprising though; this categorisation is based on Wikipedia data as the source of monolingual data, and Wikipedia has articles only in 325 languages including 7 constructed languages such as \textit{Esperanto}\footURL{https://bit.ly/WikiList}. Therefore, inherently, all the other languages automatically get labeled as extremely low resourced.  %However, they considered the data resources available in LDC and LREC as the only source of annotated data. For raw data, they considered the amount of content in Wikipedia.

%What are these extremely low resource languages? where are they being spoken? and above all, why do these languages have to be low resourced and what can be done to take them out of their current status? This paper intends to answer these questions, at least to a certain extent.
However, such a data-centric perspective  tends to overlook other aspects of resourcefulness, such as the inclusion of a language in multilingual web-based platforms such as Facebook, or the inclusion in pre-trained multilingual models such as mBERT~\cite{devlin-etal-2019-bert} and XLM-R~\cite{conneau-etal-2020-unsupervised}. Moreover, such a narrow view does not shed light on how this language disparity could be explained with respect to other socio-economic-linguistic factors such as language family, geographical location or speaker population.

This paper provides a deeper analysis into the less-known facts of the well-known problem of linguistic disparity in the world. We start with an existing language categorisation based on speaker population and vitality (Ethnologue\footURL{https://bit.ly/3kJircB})~\cite{ethnologue}, and analyse the distribution of language data resources, amount of NLP/CL research, inclusion in multilingual web-based platforms and the inclusion in pre-trained multilingual models. \textbf{ We show that simply categorising languages using data availability as done by~\citet{joshi-etal-2020-state} can be misleading.} We also show that many languages are neglected with respect to all the considered criteria, and even within the languages belonging to the same language group, there is wide disparity. We analyse this disparity with respect to the family, geographical location, as well as the speaker population and GDP. We also provide possible reasons for this disparity, along with some recommendations to eradicate the same.%and argue that most these reasons are beyond the control of ACL, as an organization. Based on this argument, we provide a preliminary set of recommendations that may be implemented by various stakeholders, in reducing this disparity across languages.

\section{The 12 Kinds of Languages}
%In order to understand the language disparity at a deeper level with respect to socio-economic-linguistic factors, we  use the Ethnologue language categorisation.

Ethnologue is an annual publication that provides statistics and other information of the living languages in the world. It has 7139 language entries, including dialects. We could identify 6420 unique languages by considering alternate names, dialects, and minor schisms to map to their most prominent entry. The language list we extracted, as well as the selection criteria are in Appendix~\ref{sec:ehtnologue}.

Ethnologue languages are categorised into  12 classes, based on 2 variables: \textit{Population} and \textit{Vitality}. \textit{Population} is ``the estimated number of all users (including both first and second language speakers) in terms of three levels'', the aforementioned three levels being: \textit{large}, \textit{Mid-sized}, and \textit{small}~\cite{ethnologue}. \textit{Vitality} is categorised into four distinct classes: \textit{institutional}, \textit{stable}, \textit{endangered} and \textit{extinct}, according to the Expanded Graded Intergenerational Disruption Scale (EGIDS) grid~\cite{lewis2010assessing}. 

\begin{figure*}[!hbt]
     \centering
      \begin{subfigure}[!hbt]{\fourth\textwidth}
         \centering
         \includegraphics[width=\textwidth]{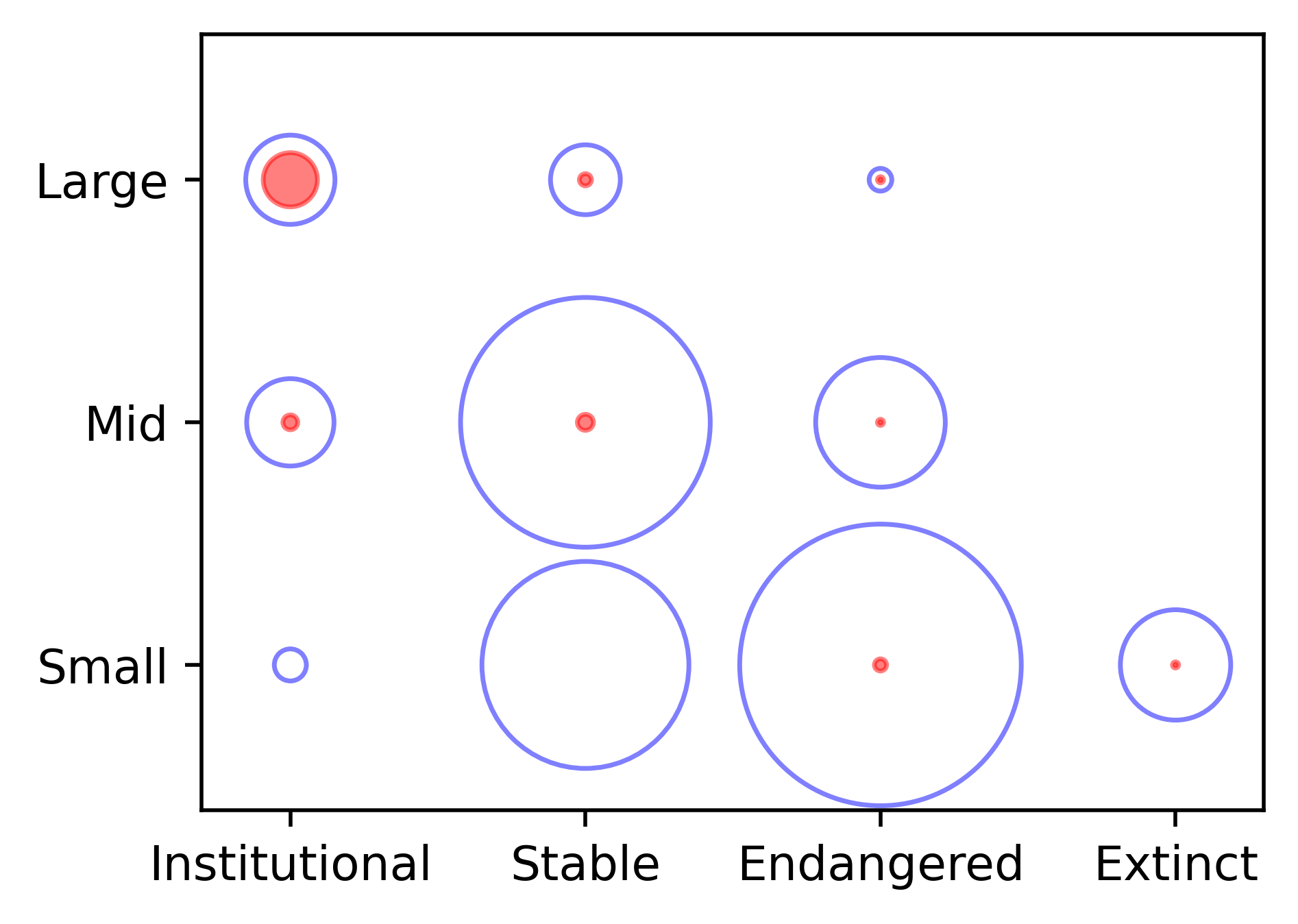}
         \caption{LDC}
    \end{subfigure}
    \begin{subfigure}[!hbt]{\fourth\textwidth}
         \centering
         \includegraphics[width=\textwidth]{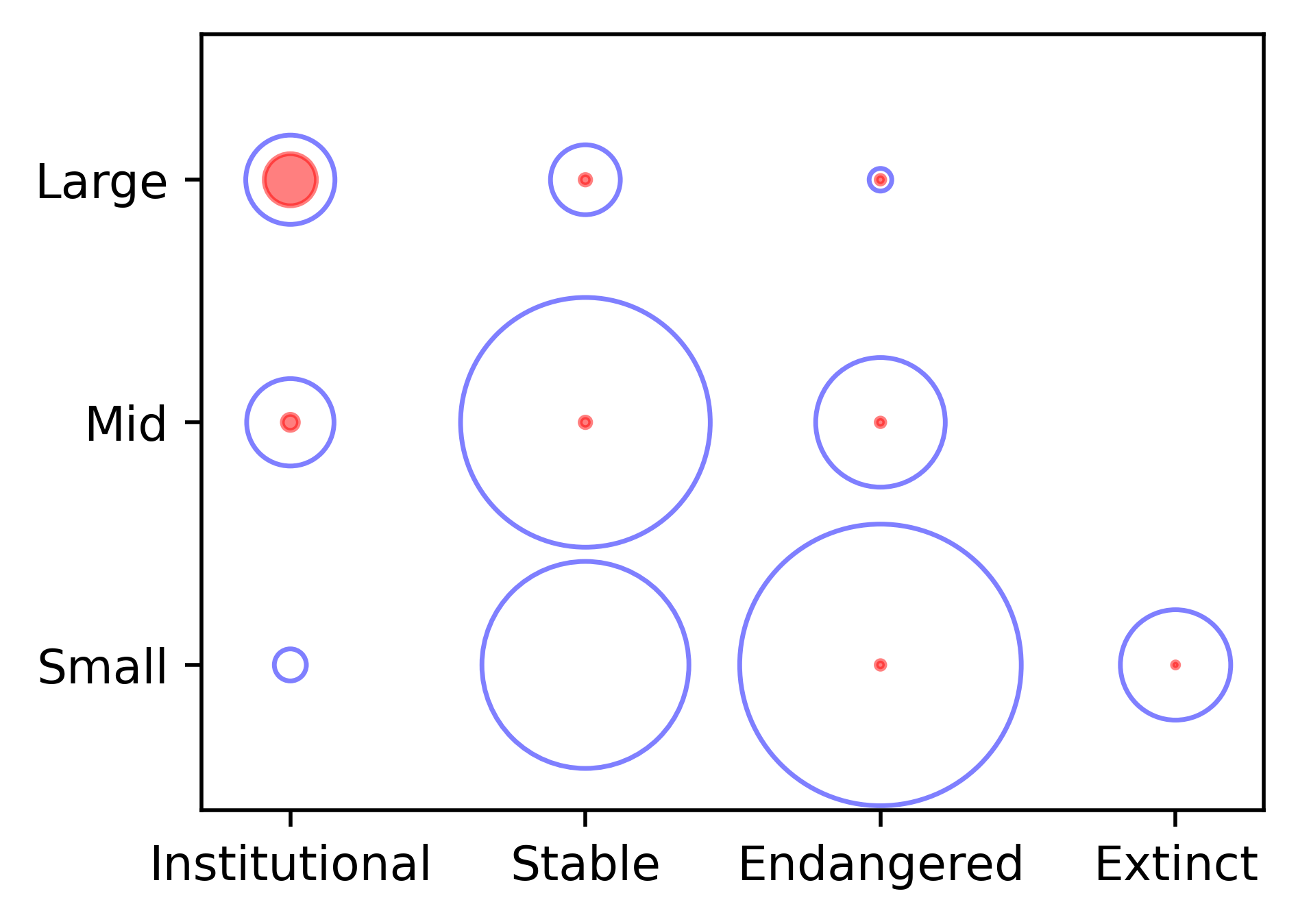}
         \caption{ELRA}
    \end{subfigure}
    \begin{subfigure}[!hbt]{\fourth\textwidth}
         \centering
         \includegraphics[width=\textwidth]{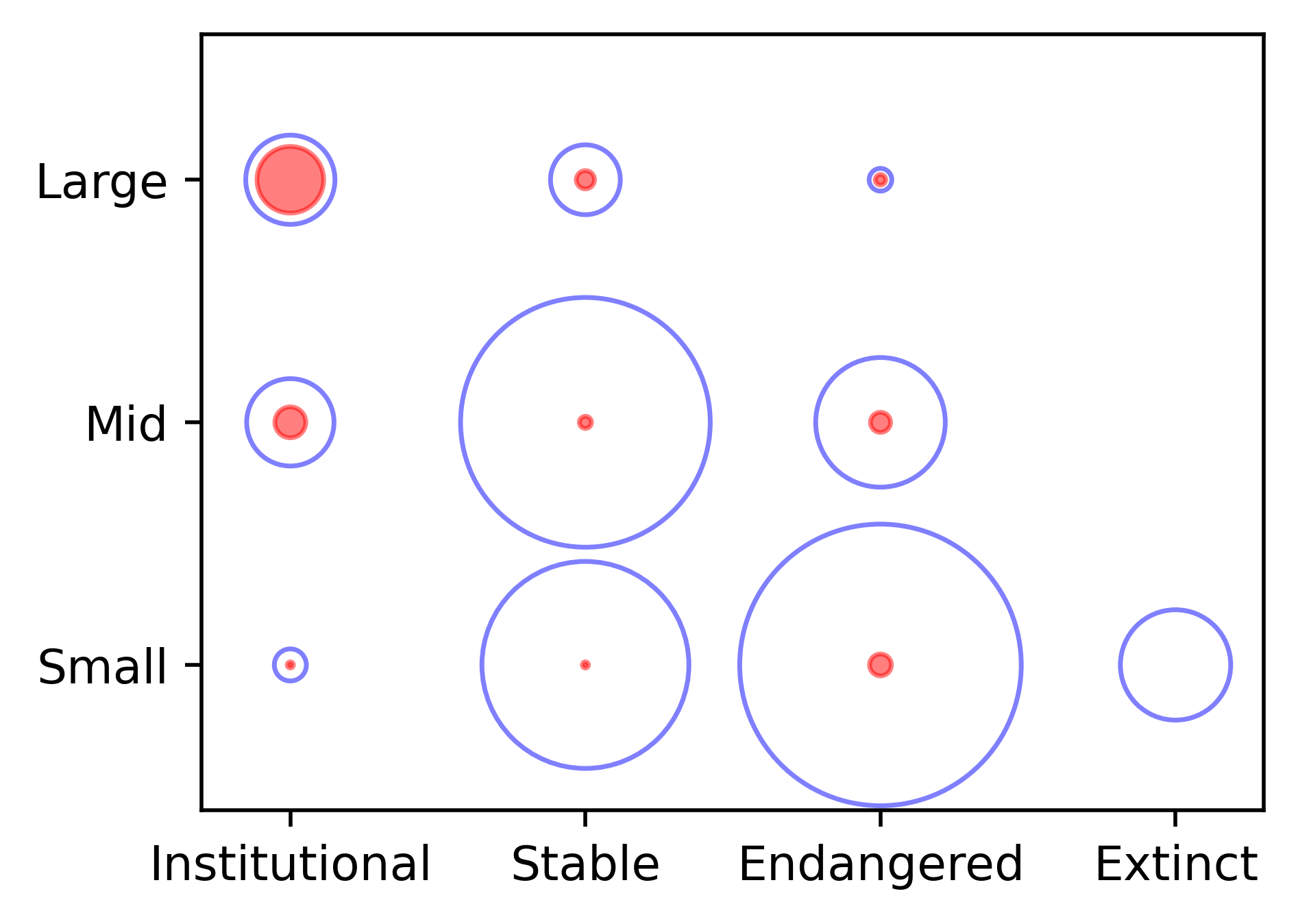}
         \caption{Huggingface}
    \end{subfigure}
     \begin{subfigure}[!hbt]{\fourth\textwidth}
         \centering      \includegraphics[width=\textwidth]{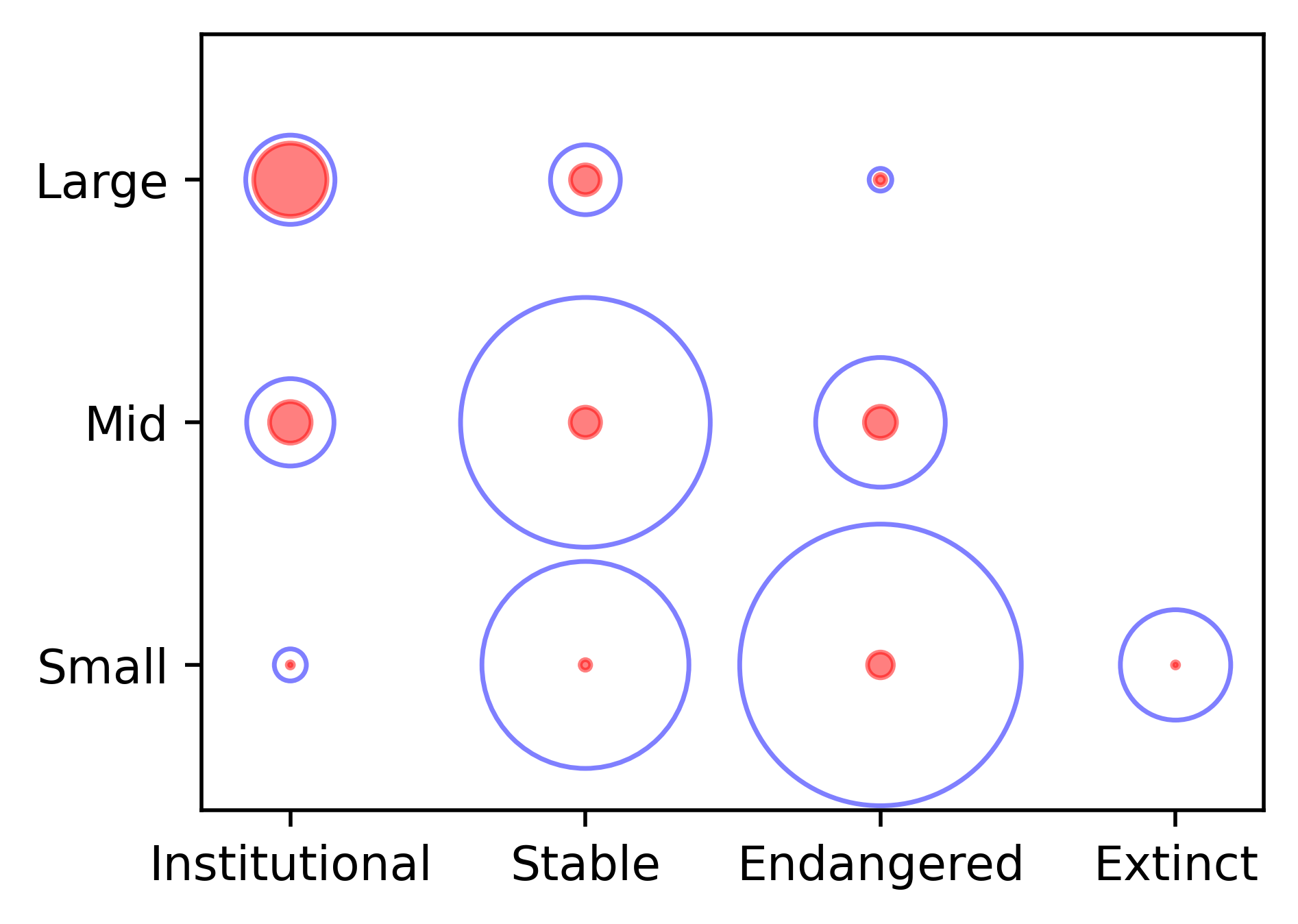}
         \caption{Wikipedia}
    \end{subfigure}
    
    \begin{subfigure}[!hbt]{\fourth\textwidth}
         \centering         \includegraphics[width=\textwidth]{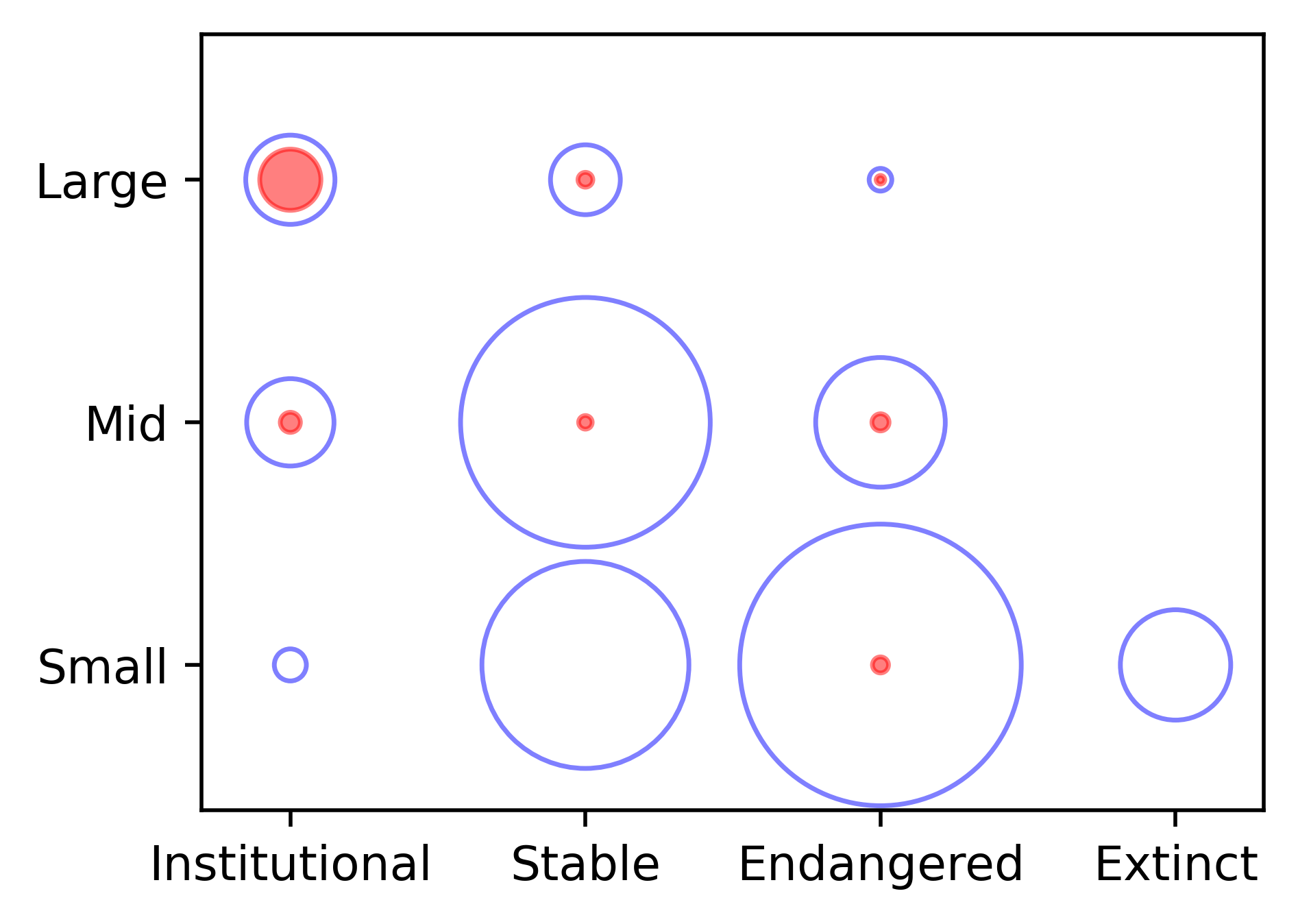}
         \caption{Facebook}
    \end{subfigure}
    \begin{subfigure}[!hbt]{\fourth\textwidth}
         \centering         \includegraphics[width=\textwidth]{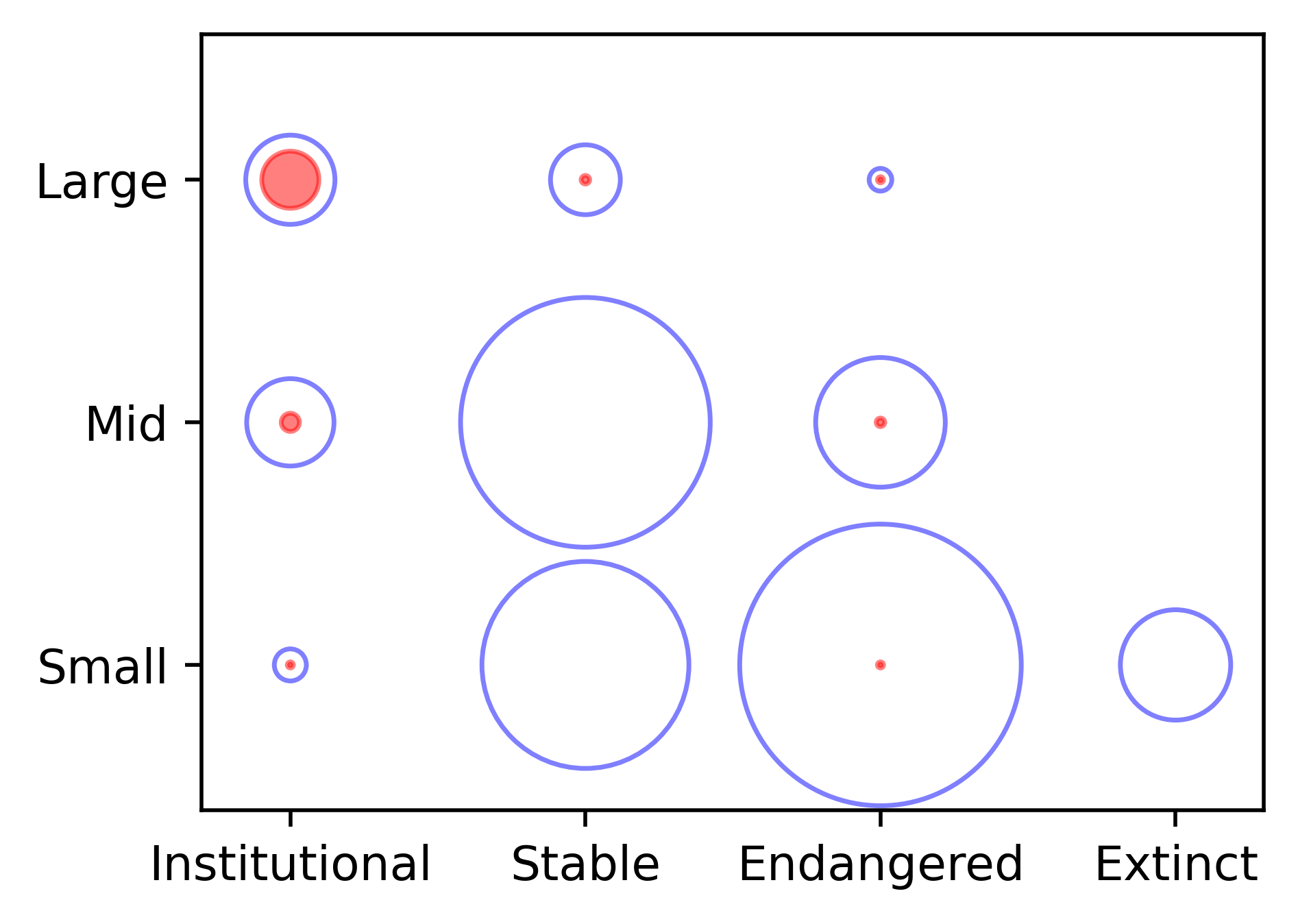}
         \caption{Google Translate}
    \end{subfigure}
    \begin{subfigure}[!hbt]{\fourth\textwidth}
         \centering         \includegraphics[width=\textwidth]{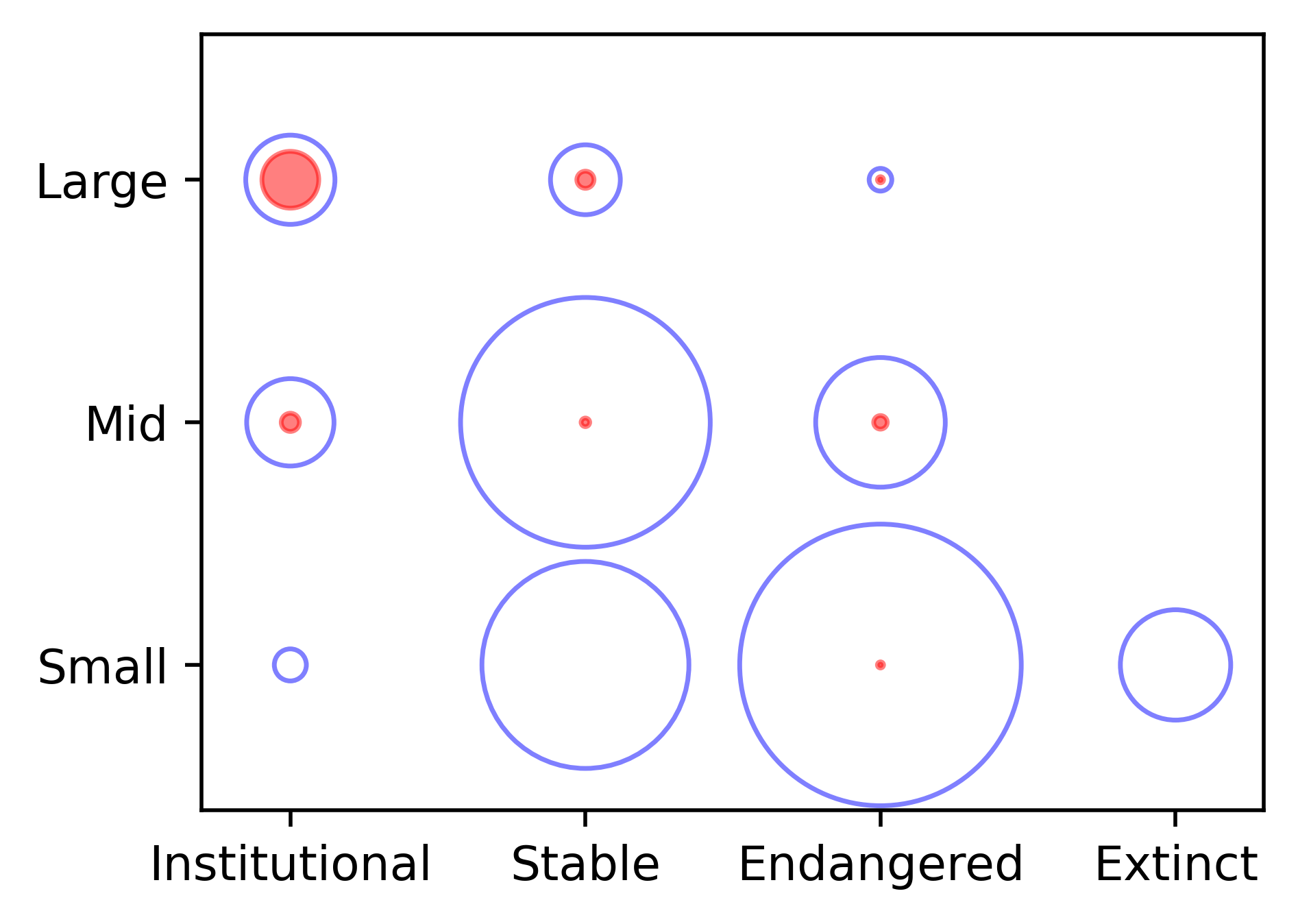}
         \caption{XLMR+mBERT}
    \end{subfigure}
    \begin{subfigure}[!hbt]{\fourth\textwidth}
         \centering         %\includegraphics[width=\textwidth]{images/bubbleplot_ACL count.png}
         \includegraphics[width=\textwidth]{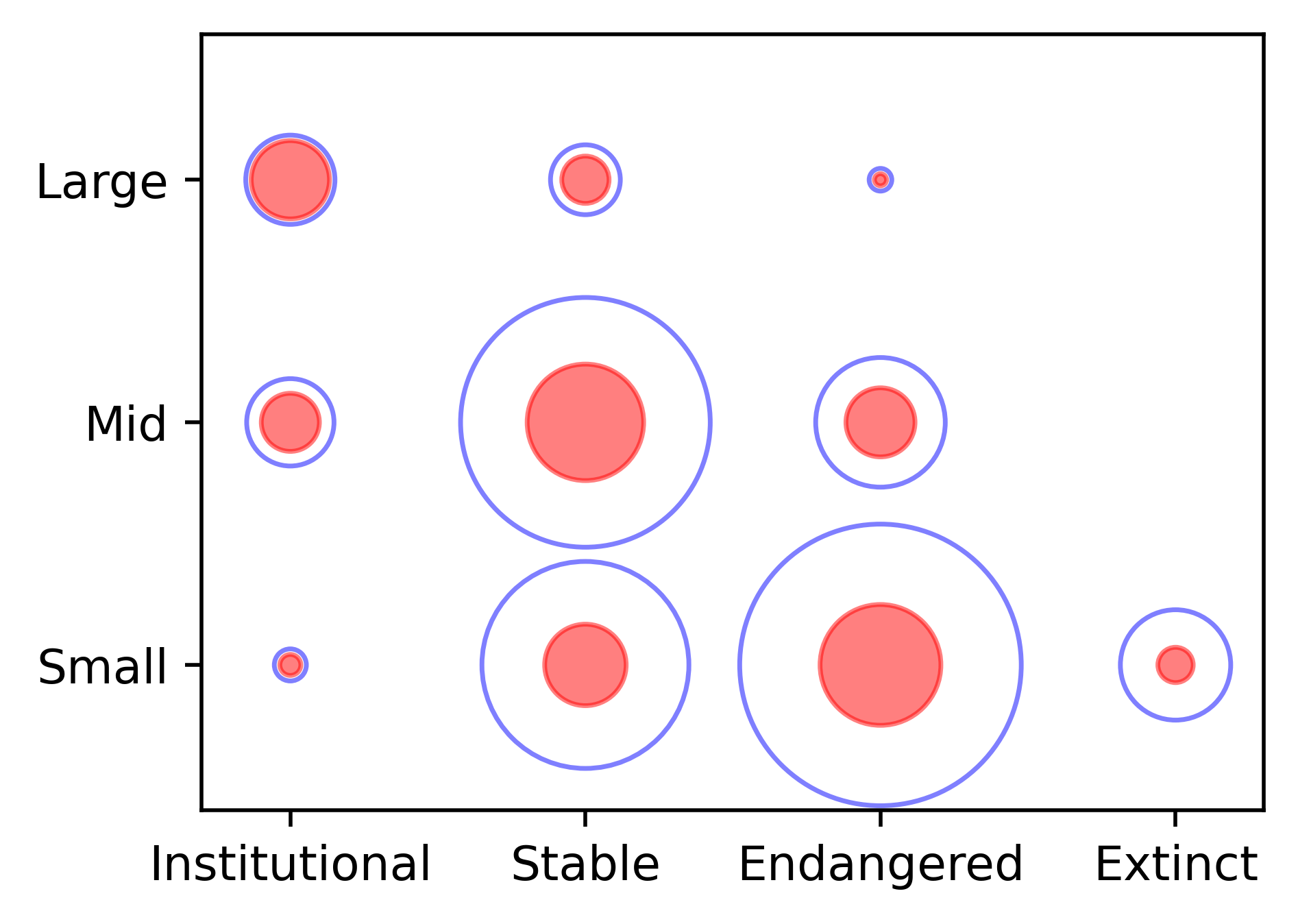}
         \caption{ACL Anthology}
        \label{fig:ethnologue_data_ACL}
    \end{subfigure}
    \caption{The 12 Ethnologue language classes where the size of each outer circle corresponds to the number of languages in that category and the size of each red circle corresponds to the coverage of that class in the relevant resource.}
    \label{fig:ethnologue_data}
\end{figure*}

We plotted the languages in a 12-point grid, according to vitality and number of speaker population. The size of the outer circles corresponds to the number of languages in one category. According to Figure~\ref{fig:ethnologue_data}, a large number of languages are endangered with small speaker populations, or stable but with mid or small speaker population numbers. Note that two classes do not have any representation in this grid. Therefore, hereafter we only refer to the remaining 10 classes.

%\begin{figure}[!hbt]
%     \centering
%    \includegraphics[width=\half\textwidth]{images/bubbleplot.png}
%    \caption{Language distribution across the 12 Ethnologue Classes}
%    \label{fig:Ethnologue}
%\end{figure}

%From this graph, the reason for low resourcefulness of many languages is evident - many languages either do not have any institutional support, or enough speaker population - this inherently leads to the scarcity of digital language resources. When there is no demand for language technologies  due to unfavorable socio-economic conditions in the region or lack of access to technology, or purely due to lack of language resources due to colonial suppression, there would be a dearth of digital language resources and tools~\cite{nekoto2020participatory}. Another reason could be the disconnection between different (indigenous) communities and the documentary linguistics community~\cite{bird2020decolonising}.  According to geographical distribution of endangered languages visualized in Ethnologue\footnote{\url{https://www.ethnologue.com/guides/how-many-languages-Endangered}}, the severity of this problem is even more pronounced. Most of the endangered languages are in the global south - meaning that the institutions would not have the logistics to develop language resources or technologies, nor does the population has enough literacy levels or access to technology.

\section{Resource \& Tool Support Distribution}
\label{sec:resource_dist_intro}
We analyse how languages in the  Ethnologue categories are being treated with respect to data (annotated and un-annotated) availability, inclusion in multilingual web-based platforms and inclusion in pre-trained multilingual models. This dataset was extracted in October-November, 2021. The dataset preparation process is given in Appendix~\ref{sec:dataset}.
%Ideally, this analysis should have been carried out for the availability of language technologies as well, the  EU-funded project European 433 Language Equality (ELE) project ~\cite{meta2021white, grutzner-zahn-rehm-2022-introducing}. However, this would be a daunting task, and is out of the scope of this research. 
%With that restriction, we discuss the available resources and tools in Sections~\ref{sec:uAnno} through~\ref{Sec:Multi}, which is then followed by an aggregated analysis in Section~\ref{sec:Analysis}. 

\subsection{Un-annotated Data Availability}
\label{sec:uAnno}

There are two possible sources: Wikipedia data and CommonCrawl. However, the latter covers only 160 languages\footURL{https://bit.ly/3F9iK87}, compared to the 318 languages in Wikipedia (excluding the 7 constructed languages). Thus, we focus on Wikipedia data as the source of un-annotated data. The CommonCrawl data analysis is briefly reported in Appendix~\ref{sec:Cc}.

\subsection{Annotated Data Availability}
Although~\citet{joshi-etal-2020-state} used \textit{LDC} and \textit{ELRA}  to retrieve the number of annotated datasets, not all datasets in these sources are available for free, and there are membership charges. This can be  quite a disadvantage for researchers working under severe financial constraints. Thus not many  languages have their datasets in these repositories.  In order to highlight that categorising languages while having incomplete information about datasets gives a wrong picture (see Section~\ref{sec:revisiting_language_categorisation}), we selected another public data repository - \textit{Huggingface} data sets\footURL{https://huggingface.co/docs/datasets/}. Huggingface is known to be sparse, and the data has to be accessed via an API. On the positive side,  despite being launched in 2021, it has more datasets than ELRA and LDC. Huggingface  datasets are categorised according to language and task. Many existing datasets, such as those hosted in OPUS~\cite{tiedemann-thottingal-2020-opus}, have been already linked to Huggingface. Other possible data repositories include Zenodo\footURL{https://zenodo.org/} and CLARIN\footURL{https://www.clarin.eu/content/data}. However, these do not have a language-wise categorisation or have a smaller number of datasets.

\subsection{Multilingual Web-based Platforms}
Facebook, Google Translate and Twitter are examples for widely used multilingual web-based platforms.
The availability of a platform interface in the native language of a user encourages them to use that platform to express themselves in the same, and reinforces the legitimacy of a language~\cite{CBC_News}. 
Conversely, the languages that are %supported by these technologies will be more and more used, while those that are
not supported will be less and less used~\cite{bird-2020-decolonising}. For our analysis, we considered the languages covered by Google Translate and the languages supported by Facebook, as these have the widest language coverage (Twitter supports 36 languages). Note that what we are checking here is the availability of a system interface in one's own language. We are not considering the ability to type using a language script.  %\sr{Note that here we are referring to language localization of the product interface, but not the ability to view content in one's script, which should be explained using the unicode support for that language. Unfortunately, it is not straightforward to identify the full list of languages having unicode support\footURL{https://tinyurl.com/yc2pr25r}}.% \sr{We note that Facebook data cannot be extracted for research purposes. However, as mentioned above our point here is, having interfaces in local languages encourages people  to use their language digitally. Also note that here we are not referring to the ability to view content in one's script, which should be explained using the unicode support for that language. Unfortunately, it is not straightforward to identify the full list of languages having unicode support\footURL{https://tinyurl.com/yc2pr25r}. } %
%Figure~\ref{fig:LangGeoWM} and~\ref{fig:LangFamWM} shows the distribution in the perspective of the geological location and the language families.

\subsection{Pre-trained Multilingual Model Coverage}
\label{Sec:Multi}
 \textit{mBERT} (trained with Wikipedia data) and \textit{XLM-R} (trained with CommonCrawl data) are the most popular models as of today. These models are quite effective in zero-shot and few-shot NLP tasks~\cite{hu2020xtreme, lauscher-etal-2020-zero}. They mostly perform better for languages that are included in the pre-training stage~\cite{muller-etal-2021-unseen} and  outperform their monolingual counterparts for low resource languages~\cite{wu-dredze-2020-languages}. Considering the above facts, and noting that training multilingual models is computationally expensive, languages that are included in mBERT and XLM-R would have an edge over those that are not. 

\section{Aggregated Analysis}
\label{sec:Analysis}
\subsection{Overview}
Inner circles in Figure~\ref{fig:ethnologue_data} as well as Tables~\ref{tab:datasets12} and~\ref{tab:ConAndCov} show how the languages from different categories have been included in different types of resources and web-based platforms. Note that the language categorisation shown in the bottom part of Table~\ref{tab:ConAndCov} is newly created by us, according to~\citet{joshi-etal-2020-state}'s categories (see Table~\ref{joshi_classes} in Appendix~\ref{sec:JoshiClass}).

%The size of the red circle on top of each blue circle in Fig~\ref{fig:ethnologue_data} corresponds to the coverage each language category enjoys in \textit{LDC}, \textit{ELRA}, \textit{Hugginface}, \textit{Wikipedia}, \textit{Facebook}, \textit{Google Keyboard}, \textit{XLMR+mBERT} and \textit{ACL anthology}. 
It is evident that language resource creation and technology availability have been mostly centred around institutional languages with high speaker populations, while small and endangered languages have mostly been ignored. 

\begin{table*}[!hbt]
\centering
\small
%\begin{tabular}{lrrrrrrrrrr}
\begin{tabularx}{\textwidth}{|l|ZZ|ZZ|ZZ|ZZ|ZZ|ZZ|}
\hline
\multirow{2}{*}{\centering Class} & \multicolumn{2}{c|}{LDC} & \multicolumn{2}{c|}{ELRA} & \multicolumn{2}{c|}{Huggingface} & %\multicolumn{2}{c}{CC} & 
\multicolumn{2}{c|}{Wikipedia} & \multicolumn{2}{c|}{ACL}\\
\hhline{~----------}
& Count & \% & Count & \% & Count & \% & 
%r & s &
Count & \% & Count & \% \\
\hline

       Small-Extinct & 1 & 0.30 & 1 & 0.30 & 0 & 0.00 & 1 & 0.30 & 12 & 3.61 \\
    Small-Endangered & 4 & 0.19 & 2 & 0.09 & 13 & 0.60 & 18 & 0.83 & 188 & 8.70 \\
        Small-Stable & 0 & 0.00 & 0 & 0.00 & 1 & 0.09 & 3 & 0.26 & 105 & 8.99 \\
 Small-Institutional & 0 & 0.00 & 0 & 0.00 & 1 & 3.57 & 1 & 3.57 & 5 & 17.86 \\
      Mid-Endangered & 1 & 0.22 & 2 & 0.44 & 11 & 2.40 & 28 & 6.11 & 55 & 12.01 \\
          Mid-Stable & 7 & 0.41 & 3 & 0.18 & 4 & 0.24 & 25 & 1.47 & 193 & 11.35 \\
   Mid-Institutional & 4 & 1.92 & 5 & 2.40 & 26 & 12.50 & 46 & 22.12 & 42 & 20.19 \\
    Large-Endangered & 0 & 0.00 & 2 & 14.29 & 3 & 21.43 & 3 & 21.43 & 1 & 7.14 \\
        Large-Stable & 4 & 3.01 & 3 & 2.26 & 9 & 6.77 & 24 & 18.05 & 29 & 21.80 \\
 Large-Institutional & 69 & 31.80 & 64 & 29.49 & 121 & 55.76 & 145 & 66.82 & 134 & 61.75 \\

\hline
\end{tabularx}
\caption{The \textit{Coverage} of the 10 existing Ethnologue language classes in the listed resources. Under each resource, the \textit{Count} column shows the number of languages in the relevant class included in the resource and the \textit{\%} column shows that number as a percentage of the total number of languages in the class.}
\label{tab:datasets12}
%\end{tabular}
\end{table*}

\begin{table*}[!hbt]
\centering
\small
\begin{tabularx}{\textwidth}{|l|l|ZZZ|ZZZ|Z|}
\hline
\multicolumn{2}{|c|}{\multirow{2}{*}{Class}} & \multicolumn{3}{c|}{Contribution} & \multicolumn{3}{c|}{Coverage} & Language\\
\hhline{~~------~}
\multicolumn{2}{|c|}{} & Facebook & Google & X+mB & Facebook & Google & X+mB & Count \\ 
\hline
\hline

\parbox[t]{2mm}{\multirow{10}{*}{\rotatebox[origin=c]{90}{\tiny Ethnologue}}} & Small-Extinct & 0.00 & 0.00 & 0.00 & 0 & 0 & 0 & 332 \\
& Small-Endangered & 4.96 & 0.95 & 0.88 & 0.32 & 0.05 & 0.05 & 2162 \\
& Small-Stable & 0.00 & 0.00 & 0.00 & 0 & 0 & 0 & 1168 \\
& Small-Institutional & 0.00 & 0.95 & 0.00 & 0 & 3.57 & 0 & 28 \\
%& Mid-Extinct & 0.00 & 0.00 & 0.00 & 0 & 0 & 0 & 0 \\
& Mid-Endangered & 5.67 & 1.90 & 4.39 & 1.75 & 0.44 & 1.09 & 458 \\
& Mid-Stable & 3.55 & 0.00 & 1.75 & 0.29 & 0 & 0.12 & 1700 \\
& Mid-Institutional & 7.80 & 8.57 & 7.89 & 5.29 & 4.33 & 4.33 & 208 \\
%& Large-Extinct & 0.00 & 0.00 & 0.00 & 0 & 0 & 0 & 0 \\
& Large-Endangered & 1.42 & 0.95 & 0.88 & 14.29 & 7.14 & 7.14 & 14 \\
& Large-Stable & 4.26 & 1.90 & 7.02 & 4.51 & 1.5 & 6.02 & 133 \\
& Large-Institutional & 72.34 & 84.76 & 77.19 & 47 & 41.01 & 40.55 & 217 \\
\hline
\parbox[t]{2mm}{\multirow{5}{*}{\rotatebox[origin=c]{90}{\tiny\citet{joshi-etal-2020-state}}}} & 0 & 7.80 & 0.00 & 1.75 & 0.18 & 0 & 0.03 & 6134 \\
& 1 & 11.35 & 3.81 & 9.65 & 12.31 & 3.08 & 8.46 & 130 \\
& 2 & 41.13 & 41.90 & 37.72 & 59.79 & 45.36 & 44.33 & 97 \\
& 3 & 19.86 & 27.62 & 26.32 & 93.33 & 96.67 & 100 & 30 \\
& 4 & 14.89 & 20.00 & 18.42 & 95.45 & 95.45 & 95.45 & 22 \\
& 5 & 4.96 & 6.67 & 6.14 & 100 & 100 & 100 & 7 \\
\hline
\hline
\multicolumn{2}{|c|}{Total} & 141 & 105 & 114 & & & & 6420 \\

\hhline{-----~~~-}
\end{tabularx}
\caption{\textit{Contribution} and \textit{Coverage} of the 10 existing Ethnologue language classes and \citet{joshi-etal-2020-state} classes in the listed resources where \textit{X+mB} refers to the union of \textit{XLMR} and \textit{mBERT}. %The \textit{Contribution} columns show the percentage contribution of each class to the total number of languages covered in a given resource. The \textit{Coverage} columns show the percentage of languages in each class covered by the given resource. 
If for Class $C_i$ of total $n_i$ members and a resource $R_j$ of total $m_j$ members, the number of members in $C_i$ present in $R_j$ is given by $u_{i,j}$ then, the contribution is $100(u_{i,j}/m_j)$ and the coverage is $100(u_{i,j}/n_j)$}
\label{tab:ConAndCov}
\end{table*}
\subsection{Data Availability}
Table~\ref{tab:datasets12} shows that Wikipedia has some coverage for all existing categories, including some extinct languages, which may be partly due to research efforts\footnote{\url{https://stanford.io/3mXQK0Z}}~\cite{paranjape2016improving}. However, LDC, ELRA and Huggingface have comparatively less coverage. This is to be expected, as annotated data creation takes a different level of expertise and more time (and money) compared to writing Wikipedia articles, which is more decentralized. 

\subsection{Inclusion in Web-based Platforms and Pre-trained Models}
In Table~\ref{tab:ConAndCov} we observe that Facebook and Google Translate platforms mainly cover institutional languages, with a negligible representation of other languages. The same is observed for the coverage in the pre-trained multilingual models \textit{mBERT} and \textit{XLM-R}, released by Google and Facebook, respectively. Note that such models suffer from  `curse of multilinguality'~\cite{conneau-etal-2020-unsupervised}, and the number of languages in the models has to be bound. 

 %The only languages from the Global North that are missing are Limburgish (a European language without a writing tradition), Tibetan, Nuosu, Zhuang and Southern Min from Chinese/Taiwanese regions. 

\subsection{Impact of Socio-Econo-linguistic Factors}
\begin{figure}[!hbt]
    \centering
    \begin{subfigure}[!hbt]{\halfSingle\textwidth}
         \centering         \includegraphics[width=\textwidth]{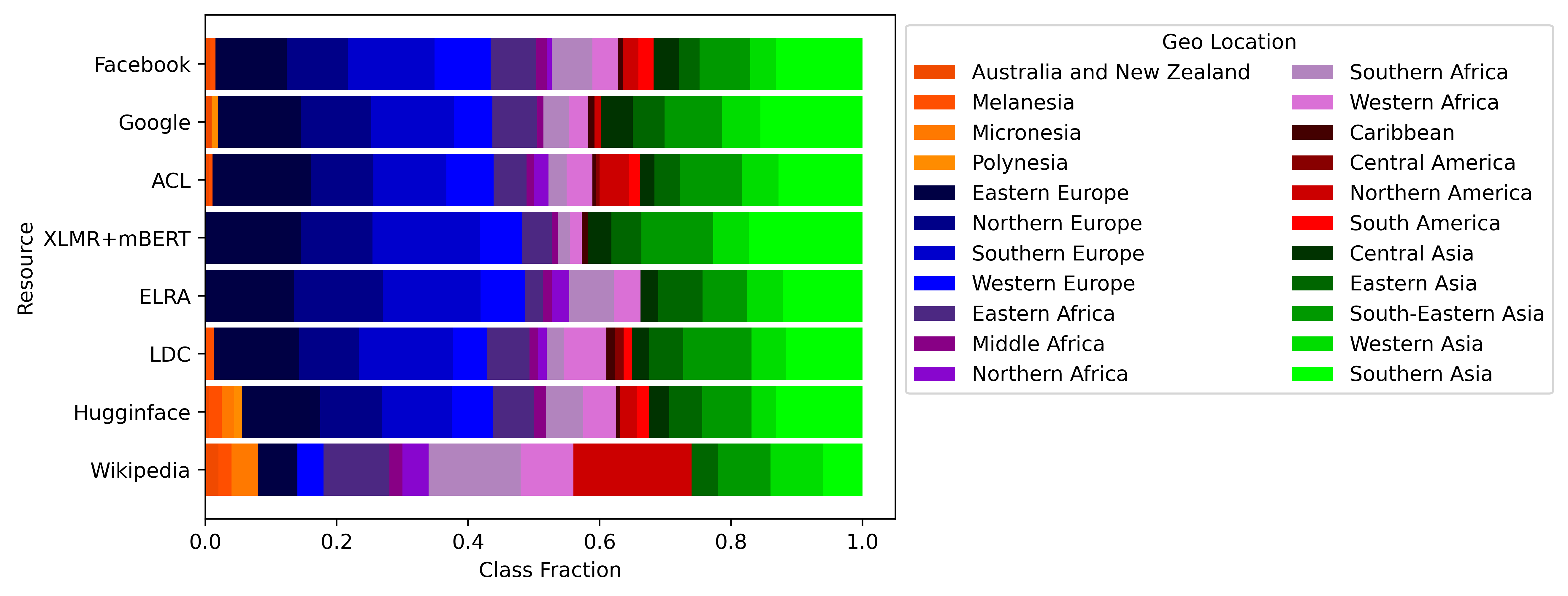}
         \caption{By Geographical Location of the Language Origin}
        \label{fig:LangGeoWM}
    \end{subfigure}
    
    \begin{subfigure}[!hbt]{\halfSingle\textwidth}
         \centering         \includegraphics[width=\textwidth]{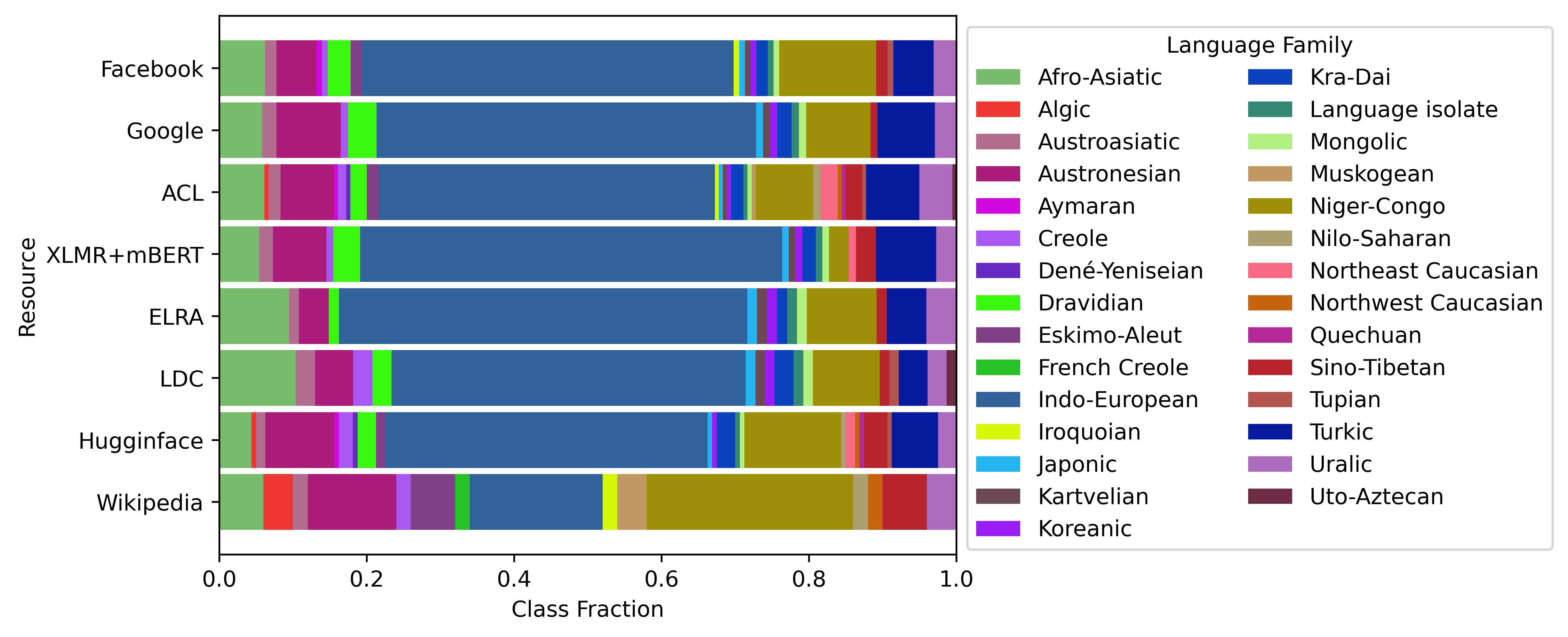}
         \caption{By Language Families}
        \label{fig:LangFamWM}
    \end{subfigure}
    \caption{The Distribution of Resources\footnotemark}
\end{figure}

Figures~\ref{fig:LangGeoWM} and \ref{fig:LangFamWM} visualise the coverage of these different platforms and resources with respect to the geographical location and family of a language. We can see that all these criteria are biased towards the \textit{Indo-European} family and the \textit{Europe} region. This is not surprising, given the emphasis placed on language resource development in the European region~\cite{meta2021white}.

Further analysis on the languages covered by \textit{mBERT} and \textit{XLM-R} models shows that the language selection has indeed been motivated by the speaker population and geographical location. Most of the languages included in these models are \textit{Large-Institutional}. As shown in Figure~\ref{fig:xlmr_coverage} in Appendix~\ref{sec:xlmr_coverage}, 75\% of non-Large-Institutional languages included in either XLM-R or mBERT are from Europe, and the rest are from Asia. All these Asian languages are either \textit{Mid-Institutional} or \textit{Large-Stable}. On the other hand, most of the Large-Institutional languages omitted from these models are in the African region (51\%). This also explains the observation made by~\citet{hu2020xtreme}, where pre-trained multilingual models perform better for Indo-European languages.

Interestingly, Wikipedia has been more democratic compared to other resources, mainly because content creation is de-centralized (More analysis in Appendix~\ref{sec:Wikipedia}). LDC and ELRA data sources are more concentrated in the Europe area. In contrast, Huggingface is more distributed. This affirms the importance of free data repositories.%, as opposed to venues such as LDC and ELRA.
\footnotetext{Larger versions are available in Appendix~\ref{sec:reso}.}

\begin{figure*}[!hbt]
     \centering
     \begin{subfigure}[!hbt]{\fifth\textwidth}
         \centering
         \includegraphics[width=\textwidth]{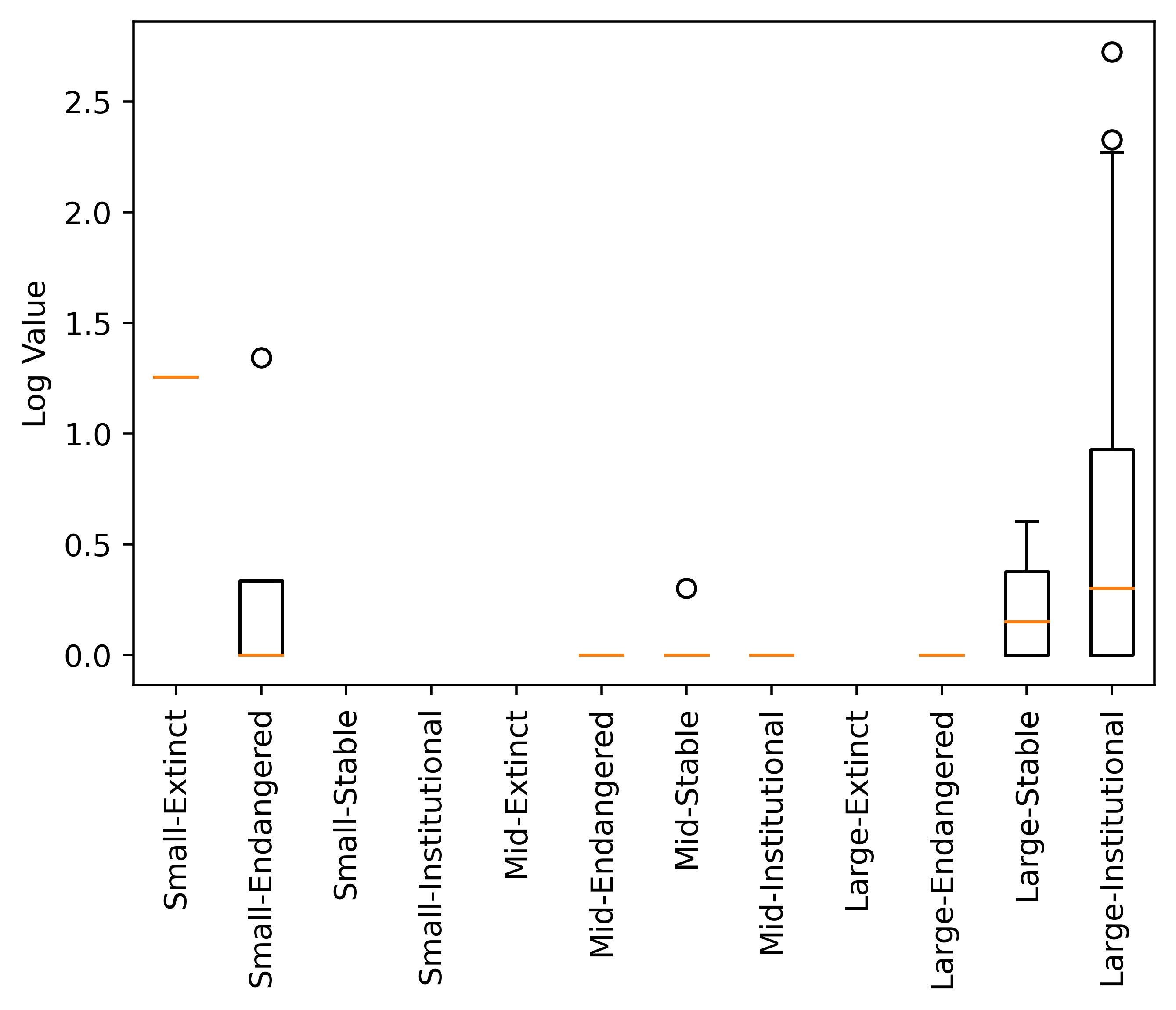}
         \caption{LDC}
    \end{subfigure}
    \begin{subfigure}[!hbt]{\fifth\textwidth}
         \centering         \includegraphics[width=\textwidth]{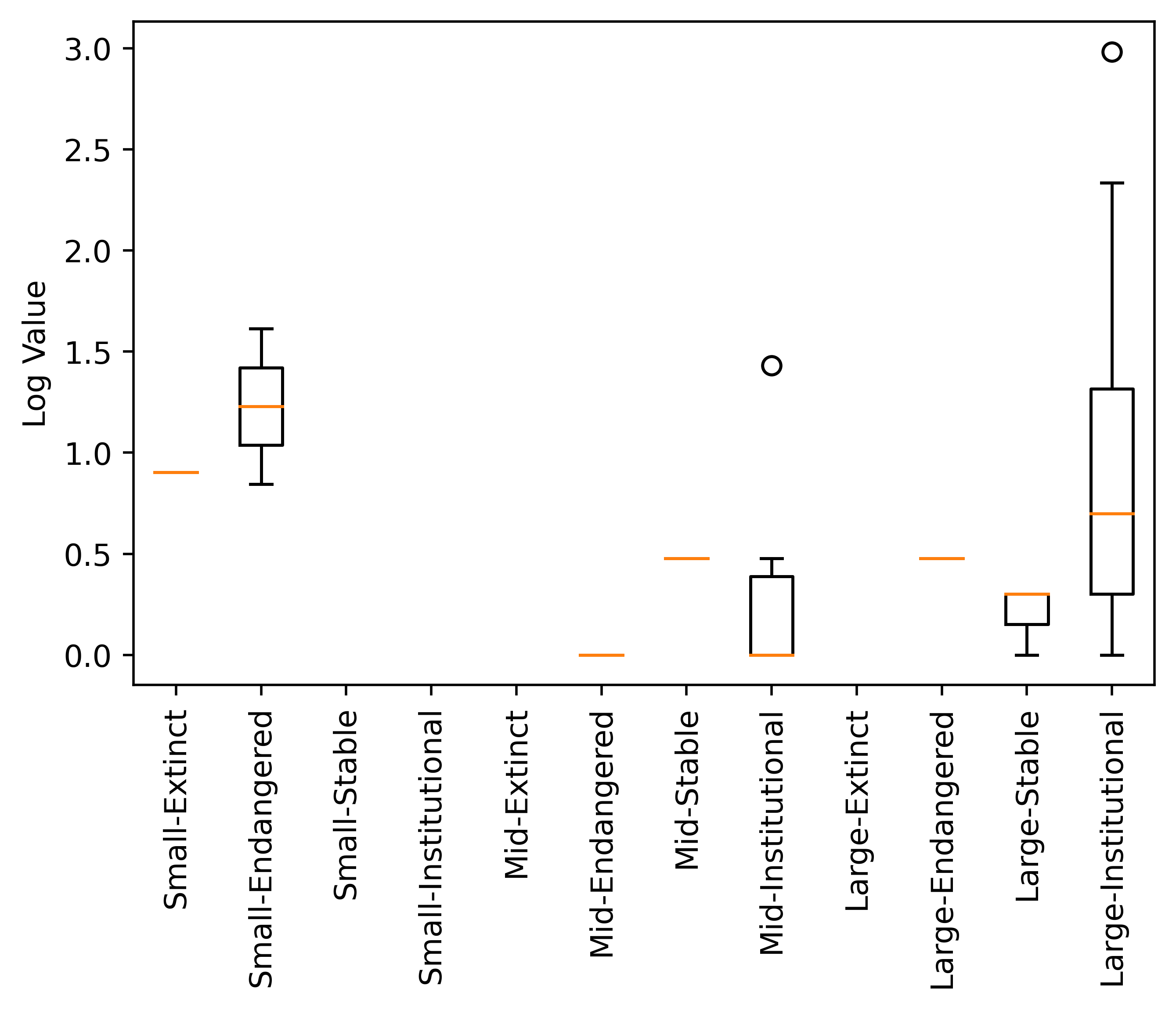}
         \caption{ELRA}
    \end{subfigure}
    \begin{subfigure}[!hbt]{\fifth\textwidth}
         \centering         \includegraphics[width=\textwidth]{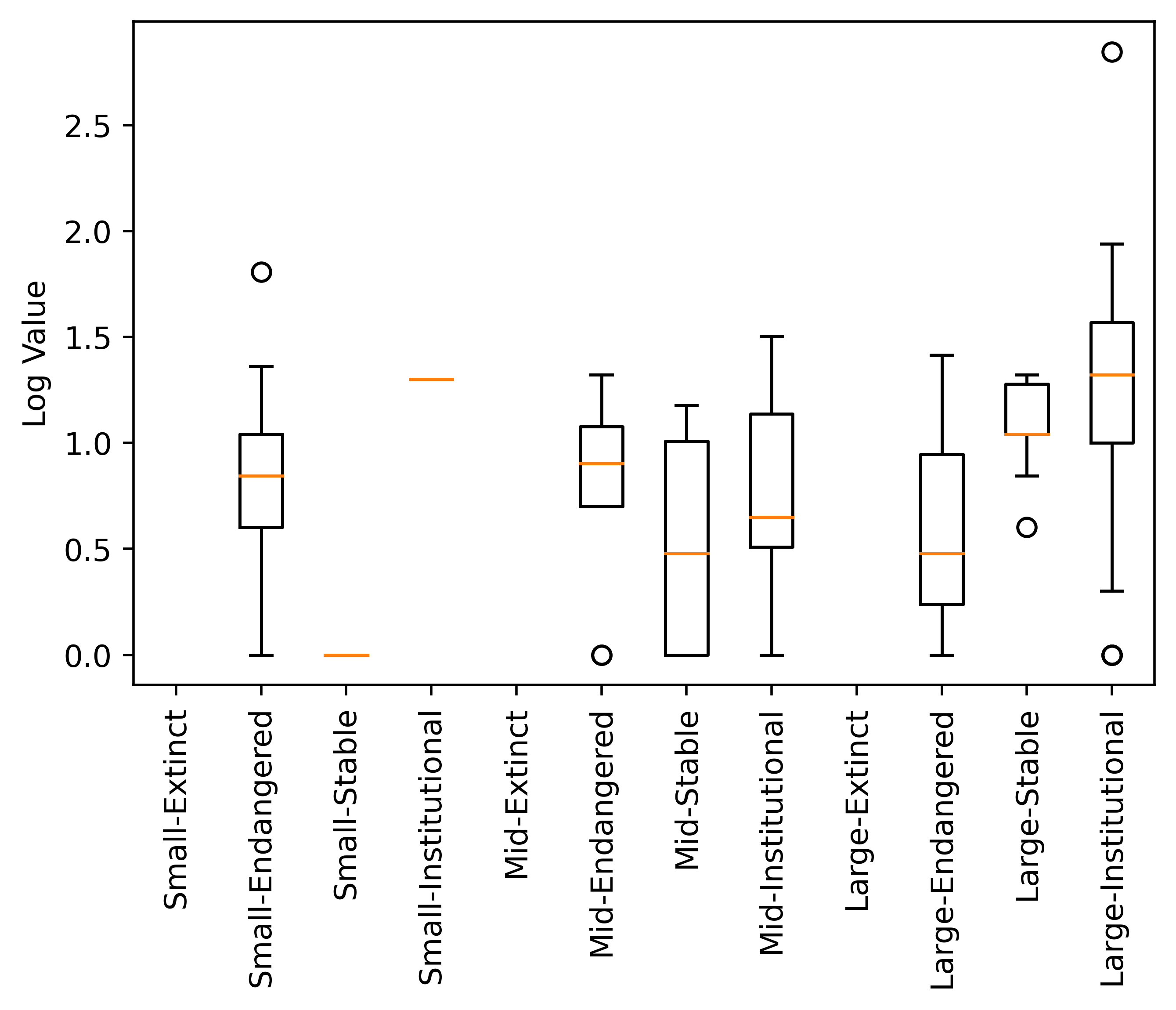}
         \caption{Huggingface}
    \end{subfigure}
    \begin{subfigure}[!hbt]{\fifth\textwidth}
         \centering
         \includegraphics[width=\textwidth]{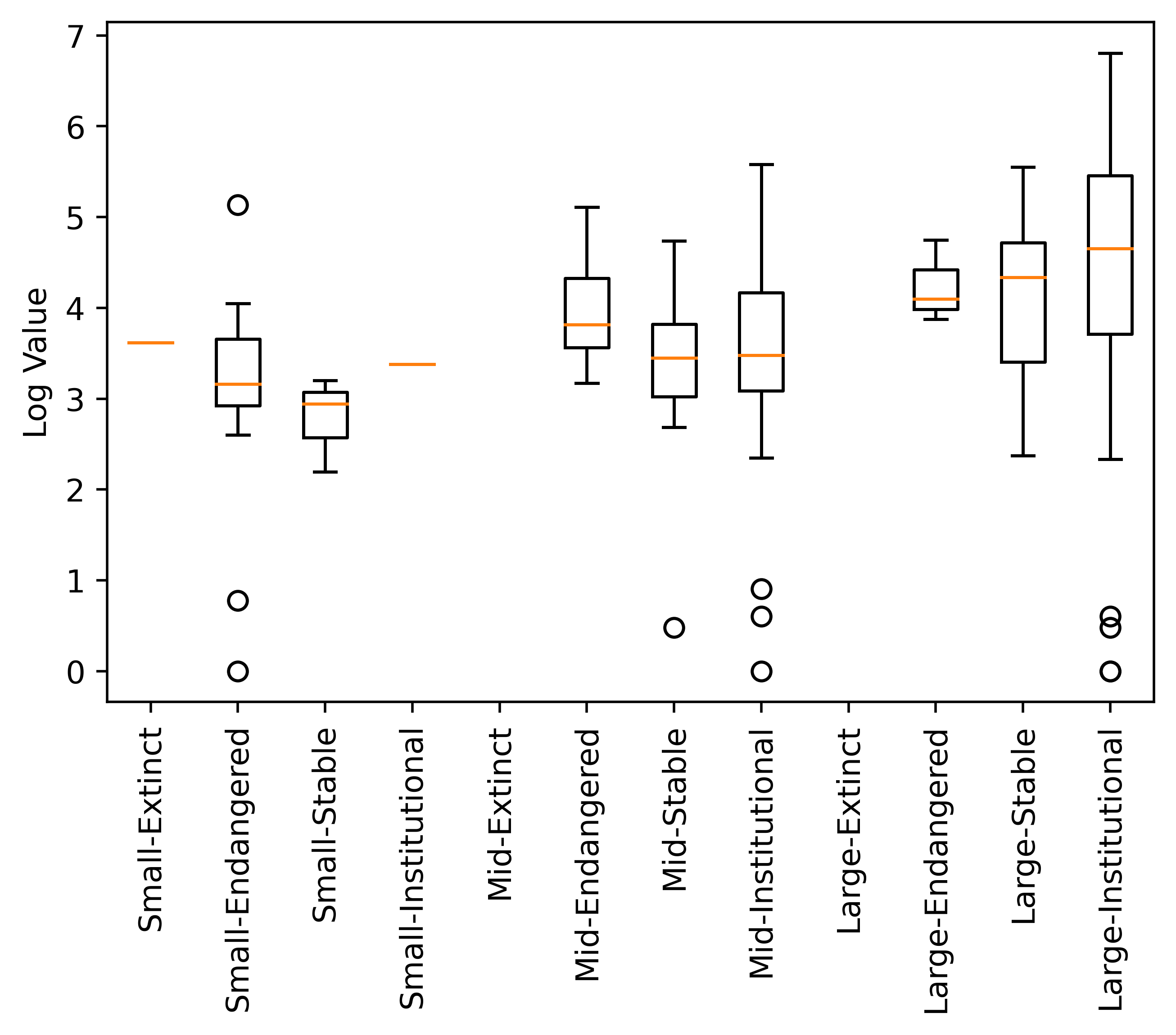}
         \caption{Wikipedia}
         \label{fig:boxWiki}
    \end{subfigure}
  %  \begin{subfigure}[!hbt]{\third\textwidth}
 %        \centering         \includegraphics[width=\textwidth]{images/boxplot_Is in XLMR.png}
 %        \caption{XLMR}
 %   \end{subfigure}
    \begin{subfigure}[!hbt]{\fifth\textwidth}
         \centering
         \includegraphics[width=\textwidth]{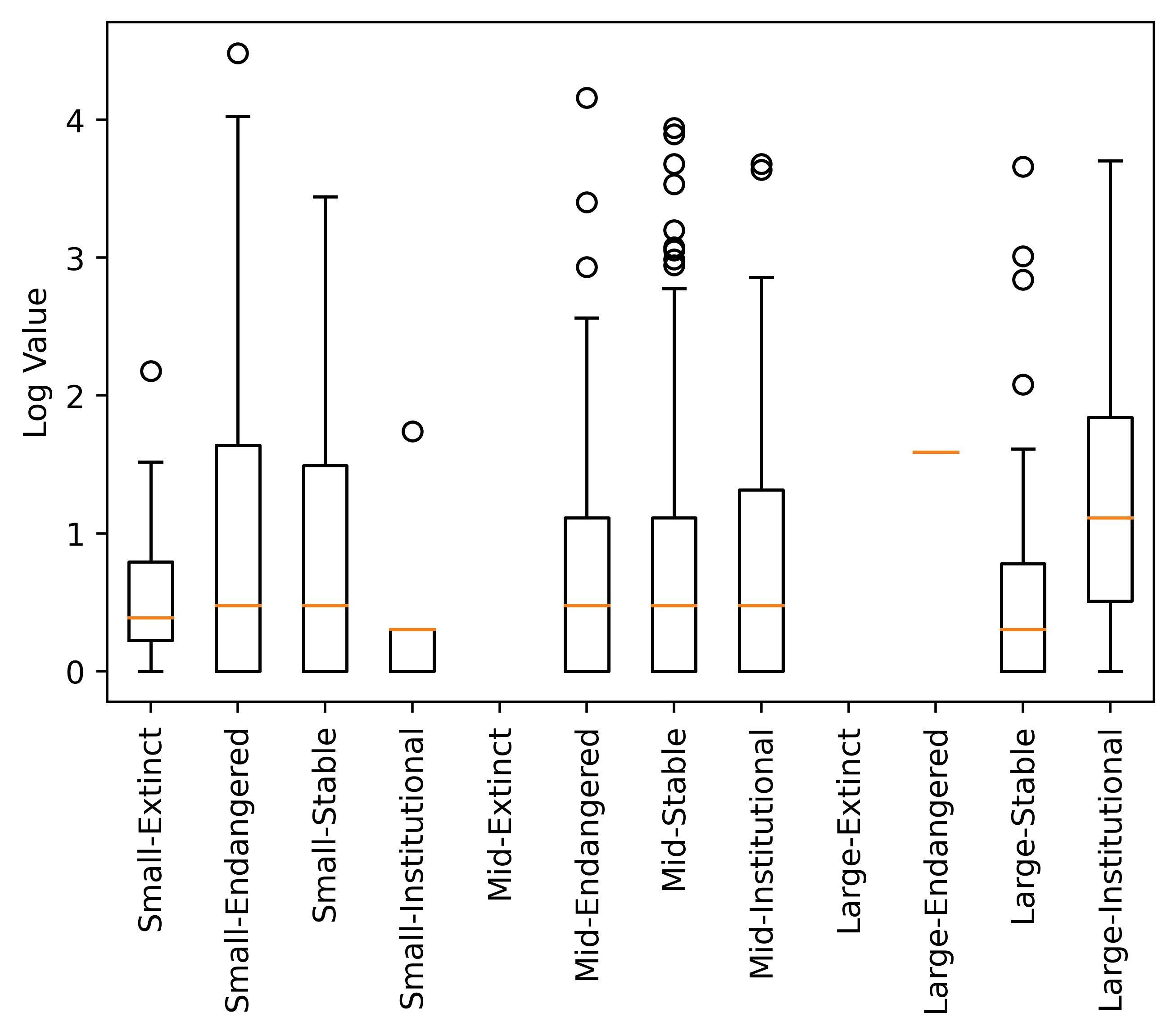}
         \caption{ACL Anthology}
         \label{fig:boxACL}
    \end{subfigure}
    \caption{Boxplots showing the resources where the amounts corresponding to the Ethnologue language classes are countable. (As opposed to Boolean)}
    \label{fig:boxplots}
\end{figure*}

However, Figure~\ref{fig:ethnologue_data} only can be misleading, as the amount of data varies across languages even within the same category. We derived the box plots shown in Figure~\ref{fig:boxplots}, which uncovered a noticeable disparity between language categories. %
Aside from the inter-class disparities, Figure~\ref{fig:boxWiki} especially shows a noticeable variance in Wikipedia data availability within the \textit{Large-Institutional} class. 

\begin{figure}[!hbt]
     \centering
    \includegraphics[width=\half\textwidth]{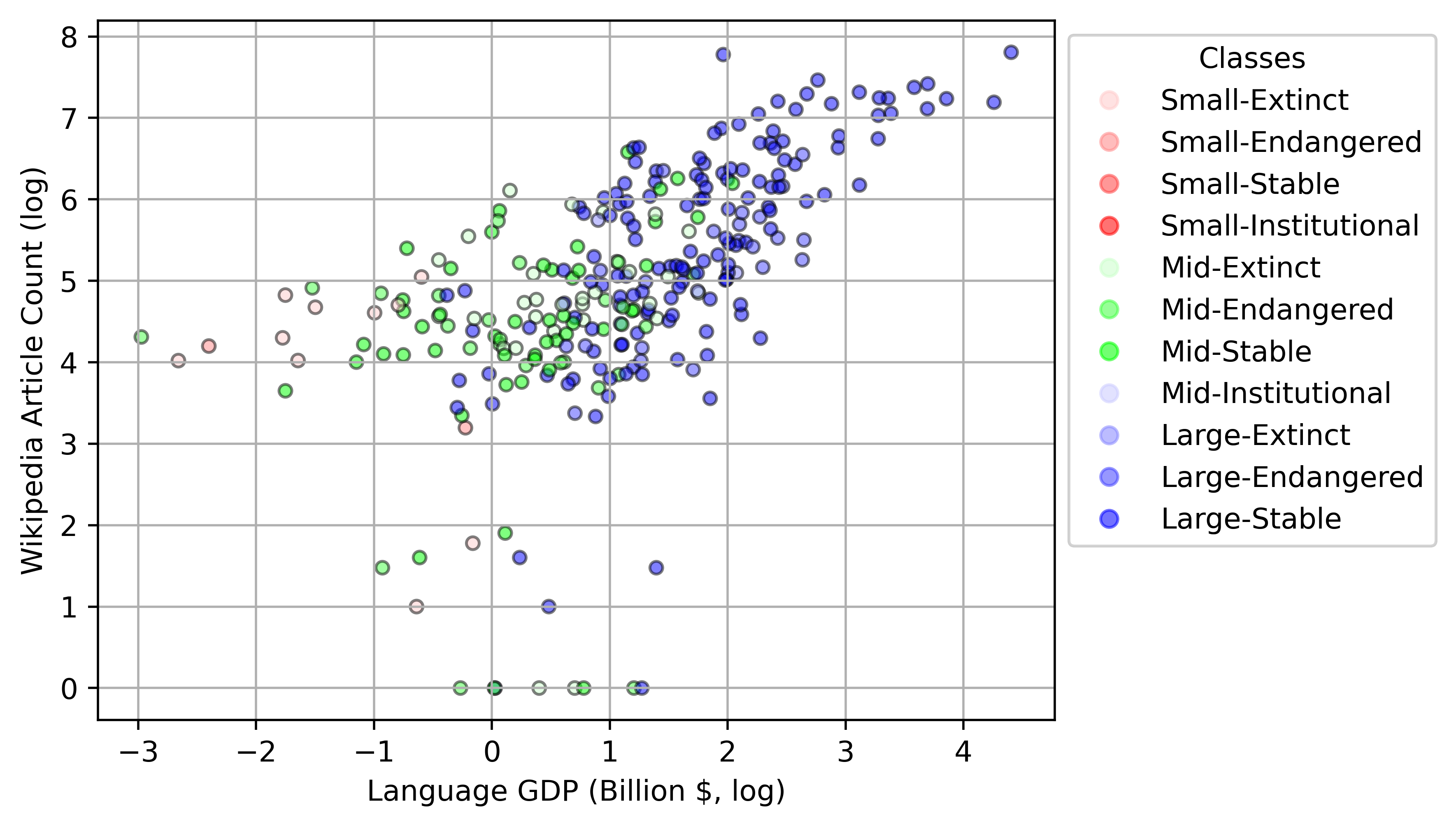}
    \caption{Language GDP in Billions of Dollars (log) vs Wikipedia Article Count (log).}
    \label{fig:GDPvsWiki}
\end{figure}

In order to understand this variance, we plotted the graph shown in Figure~\ref{fig:GDPvsWiki} and used Pearson correlation. As can be seen, the number of Wikipedia articles available has a \textit{moderate correlation} ($0.561474$) to the GDP represented by the speakers of that language\footnote{GDP, population of a country and the percentage of language speakers of a country are mainly extracted from \url{https://www.worlddata.info/}. Missing entries were identified from Wikipedia and Ethnologue. The GDP for a given language is calculated by a variation of~\citet{blasi-etal-2022-systematic} where a GDP of each country is first distributed proportionally among languages spoken as L1 in that country and then the GDP of the language is calculated by summing the aforementioned portions. The colour of each data point is taken according to the class in Ethnologue. %Therefore, data points that violate the colour boundaries along the X-axis are instances where Wikipedia and Ethnologue do not agree.
}.~\citet{blasi-etal-2022-systematic} found a similar correlation, between population and GDP, and the number of research papers per language. %About 70 languages in \textit{Large-Institutional} class do not have a presence in Wikipedia - a majority of these languages are in the African region. This suggests that other than population, economic factors also have an influence. In fact,~\citet{blasi-etal-2022-systematic} showed that GDP of the country of a language is a deciding factor in research publishing. Similarly, in this study,
Here we show that the same GDP impact can be seen in the size of Wikipedia \footnote{An equivalent analysis between population and the number of Wikipedia articles is in Appendix~\ref{sec:PopvsWiki}.}.

\subsection{Task-wise and Size-wise Analysis}
We also carried out a preliminary analysis of NLP task-wise data availability in HuggingFace. Results are shown in Table~\ref{tab:HugginTask} in Appendix~\ref{huggingface_tasks}. Despite this task categorisation being extremely noisy, % with a lot of redundant task labels,  Moreover, 50 of the 149 tasks do not have a single dataset. Still, 
there are some interesting observations. %Translation has the highest number of datasets, may be because the datasets in OPUS have been linked with HuggingFace. Other 
Popular NLP tasks such as translation, text classification, text generation and text retrieval have the highest counts, at least for Large-Institutional category. For all the tasks, dataset availability is the highest for large-Institutional, followed by Mid-Institutional.

As for the size of datasets, we are only aware of OPUS, which records the number of sentences per language. According to the results in Table~\ref{tab:OPUS} in  Appendix~\ref{sec:OPUS}, not only the number of datasets, but the amount of data samples also depends on the language class.

\section{Revisiting Data Availability-based Language Categorisation}
\label{sec:revisiting_language_categorisation}
\begin{figure*}[!hbt]
     \centering
      \begin{subfigure}[!hbt]{0.99\textwidth}
         \centering
         \includegraphics[height=15pt,keepaspectratio]{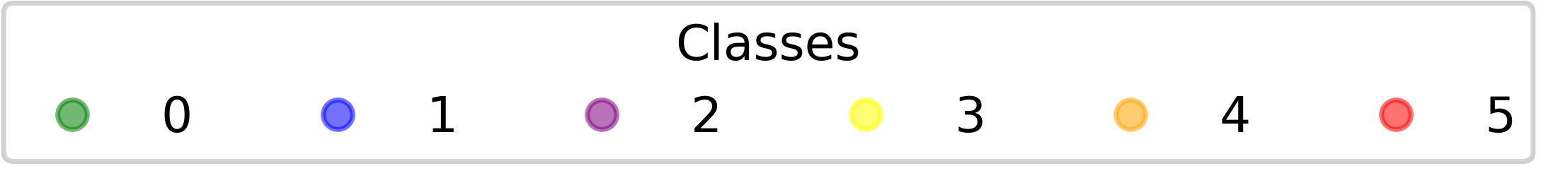}
     \end{subfigure}
     
     \begin{subfigure}[!hbt]{\half\textwidth}
         \centering      \includegraphics[width=\textwidth]{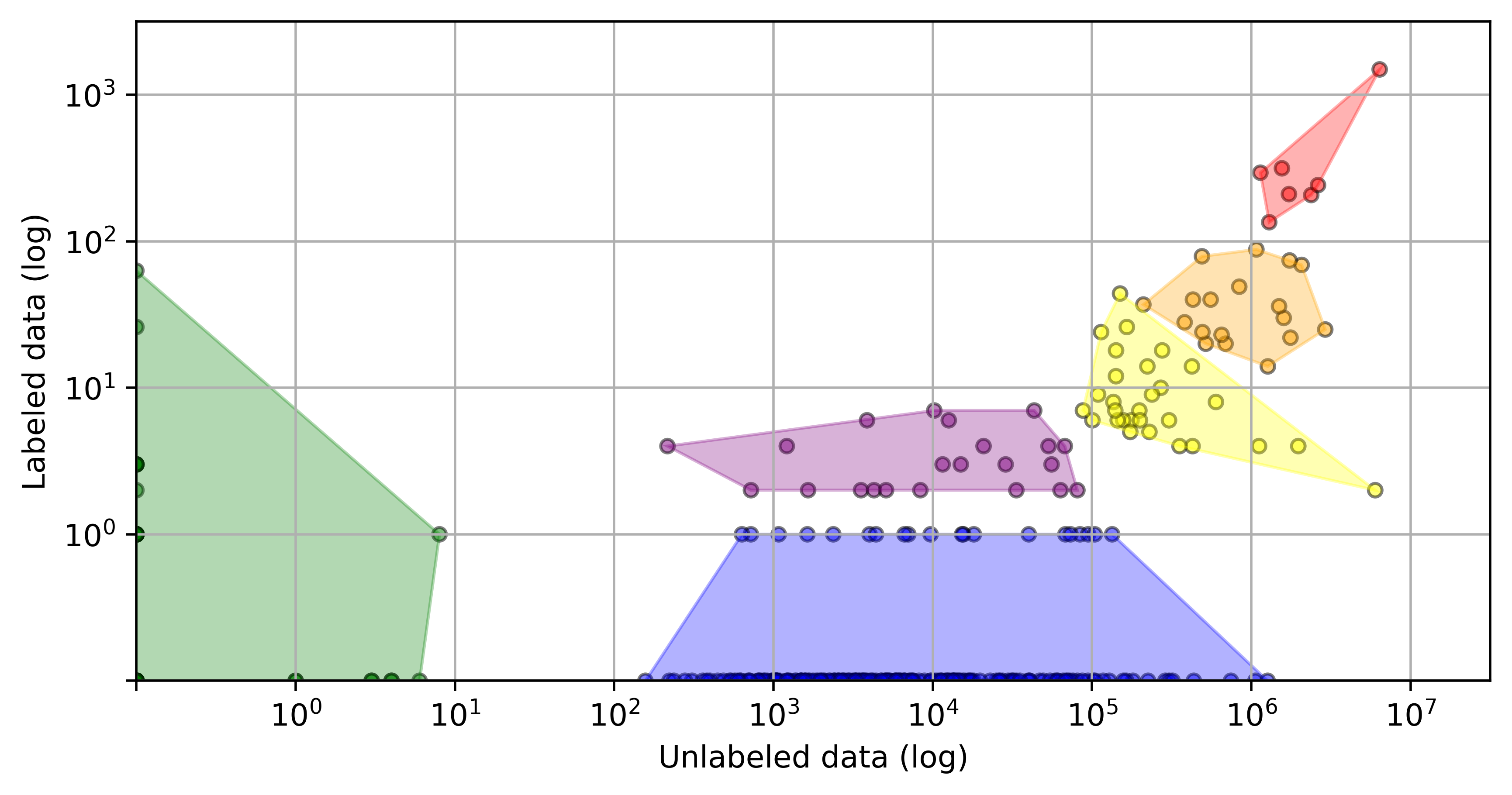}
         \caption{\textit{LDC} and \textit{ELRA} as the annotated sources}       \label{fig:joshiWithoutHugging}
    \end{subfigure}
    \begin{subfigure}[!hbt]{\half\textwidth}
         \centering
         \includegraphics[width=\textwidth]{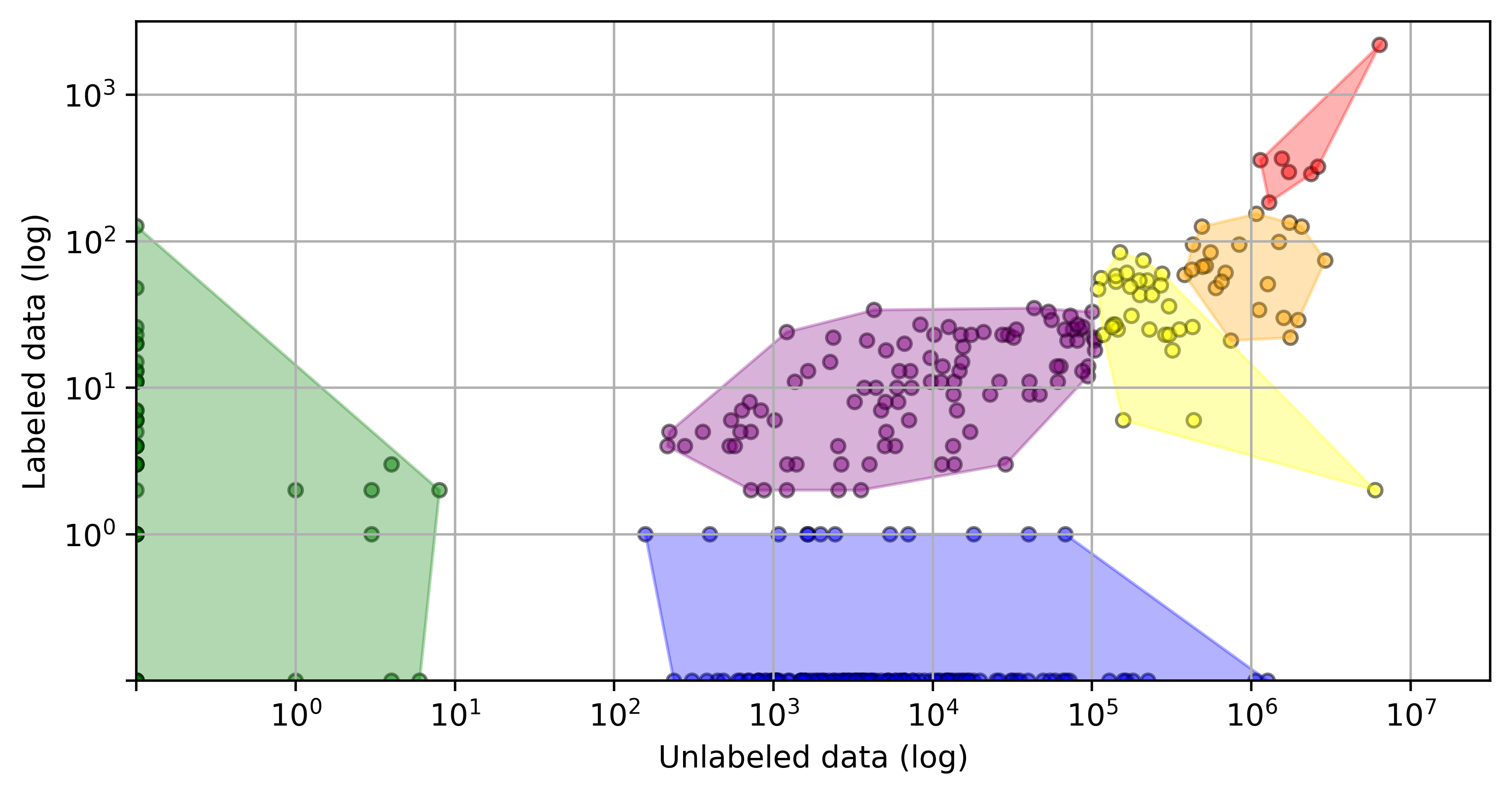}
         \caption{\textit{LDC}, \textit{ELRA}, and \textit{Huggingface} as the annotated sources}
         \label{fig:joshiWithHugging} 
    \end{subfigure}
    \caption{Reconstructing~\citet{joshi-etal-2020-state} language classes with Wikipedia article count as the unannotated source and two configurations of annotated sources.}
    \label{fig:joshiShift}
\end{figure*}

In order to analyse the robustness of using annotated data availability to categorise languages, we recreated~\citet{joshi-etal-2020-state}'s language category plot. We plot the availability of annotated data in LDC and ELRA against the unannotated wiki data in~\ref{fig:joshiWithoutHugging}\footnote{Different to~\cite{joshi-etal-2020-state}, we considered the number of \textit{Wikipedia articles}, as considering \textit{pages} could be misleading due to admin-pages such as user pages and talk pages.}. In~\ref{fig:joshiWithHugging} we plot the same graph including the HuggingFace datasets as well.

\begin{figure*}[!hbt]
     \centering
  %   \begin{subfigure}[!hbt]{\fourth\textwidth}
 %        \centering
 %        \includegraphics[width=\textwidth]{images/Sinhala_code_available_with_Axis.png}
 %        \caption{Code Availability Count}
%    \end{subfigure}
%    \begin{subfigure}[!hbt]{\fourth\textwidth}
%         \centering
 %        \includegraphics[width=\textwidth]{images/Sinhala_data_available_with_Axis.png}
 %        \caption{Data Availability Count}
%    \end{subfigure}
%    \begin{subfigure}[!hbt]{\fourth\textwidth}
 %        \centering
%         \includegraphics[width=\textwidth]{images/Sinhala_tool_publicly_available_with_Axis.png}
 %        \caption{Tool Availability Count}
  %  \end{subfigure}
  %  \begin{subfigure}[!hbt]{\fourth\textwidth}
 %        \centering
%         \includegraphics[width=\textwidth]{images/Sinhala_language_with_Axis.png}
%         \caption{Used Language Count}
 %   \end{subfigure}
   % 
    \begin{subfigure}[!hbt]{\fourth\textwidth}
         \centering
         \includegraphics[width=\textwidth]{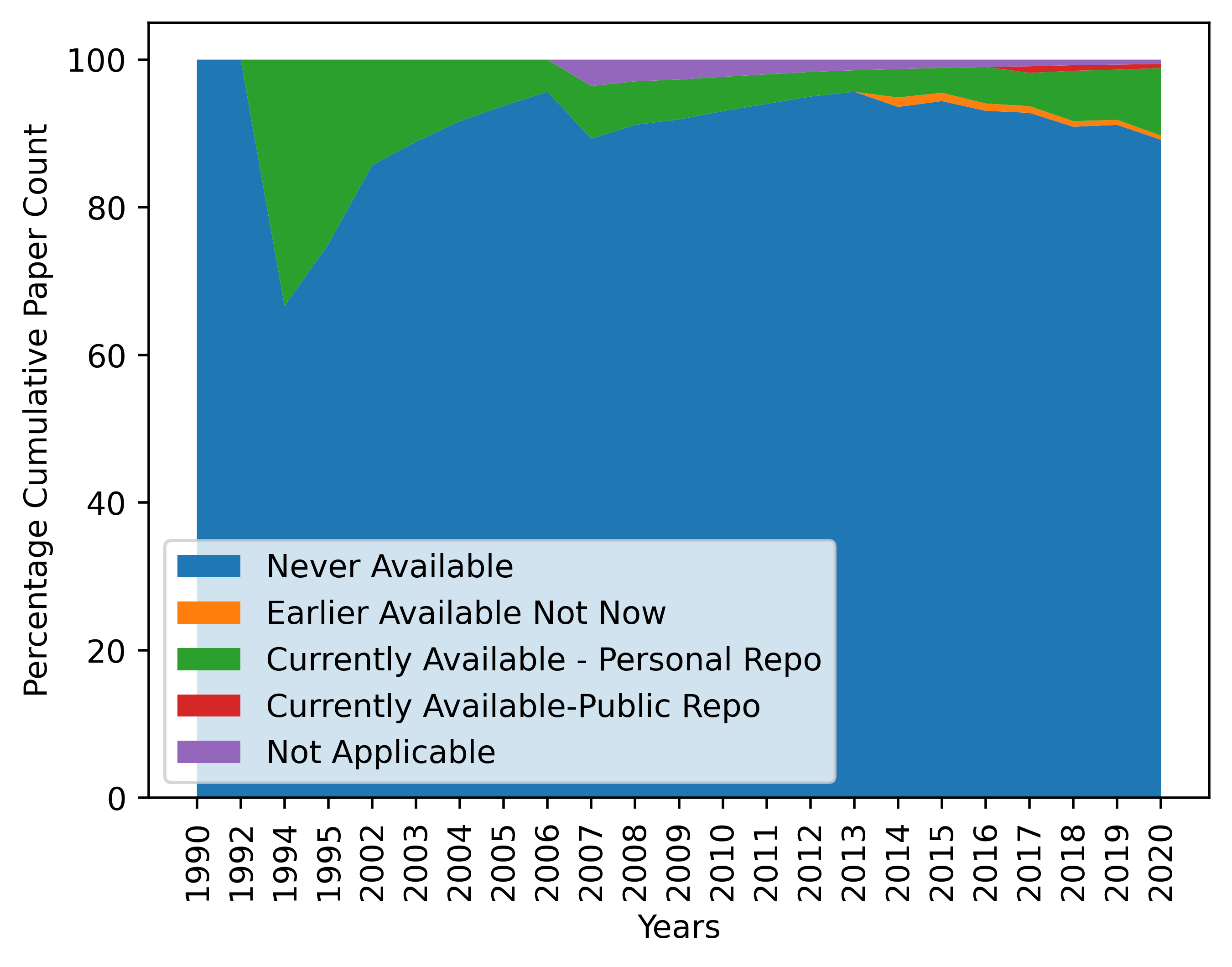}
         \caption{Code Availability}
    \end{subfigure}
    \begin{subfigure}[!hbt]{\fourth\textwidth}
         \centering
         \includegraphics[width=\textwidth]{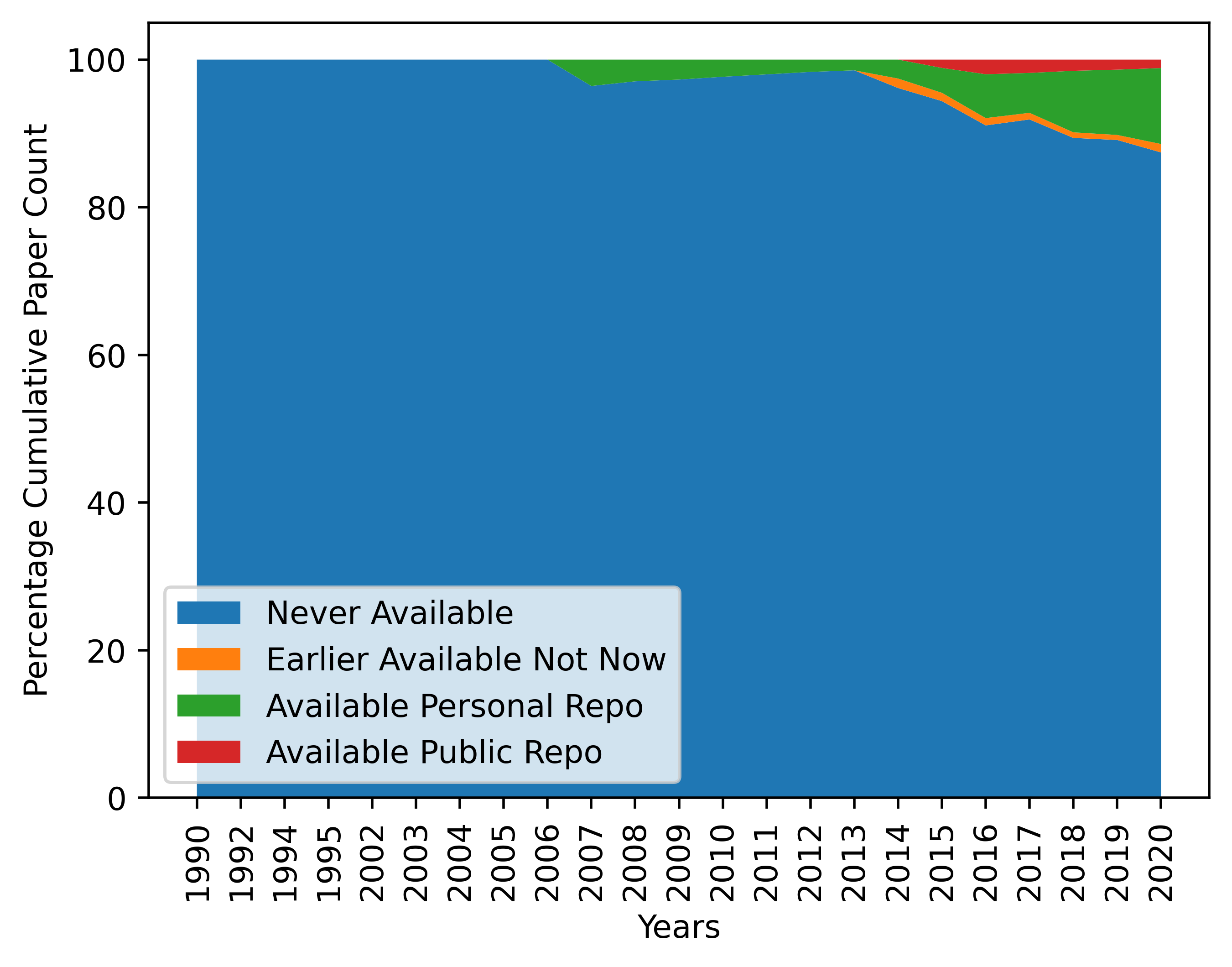}
         \caption{Data Availability}
    \end{subfigure}
    \begin{subfigure}[!hbt]{\fourth\textwidth}
         \centering
         \includegraphics[width=\textwidth]{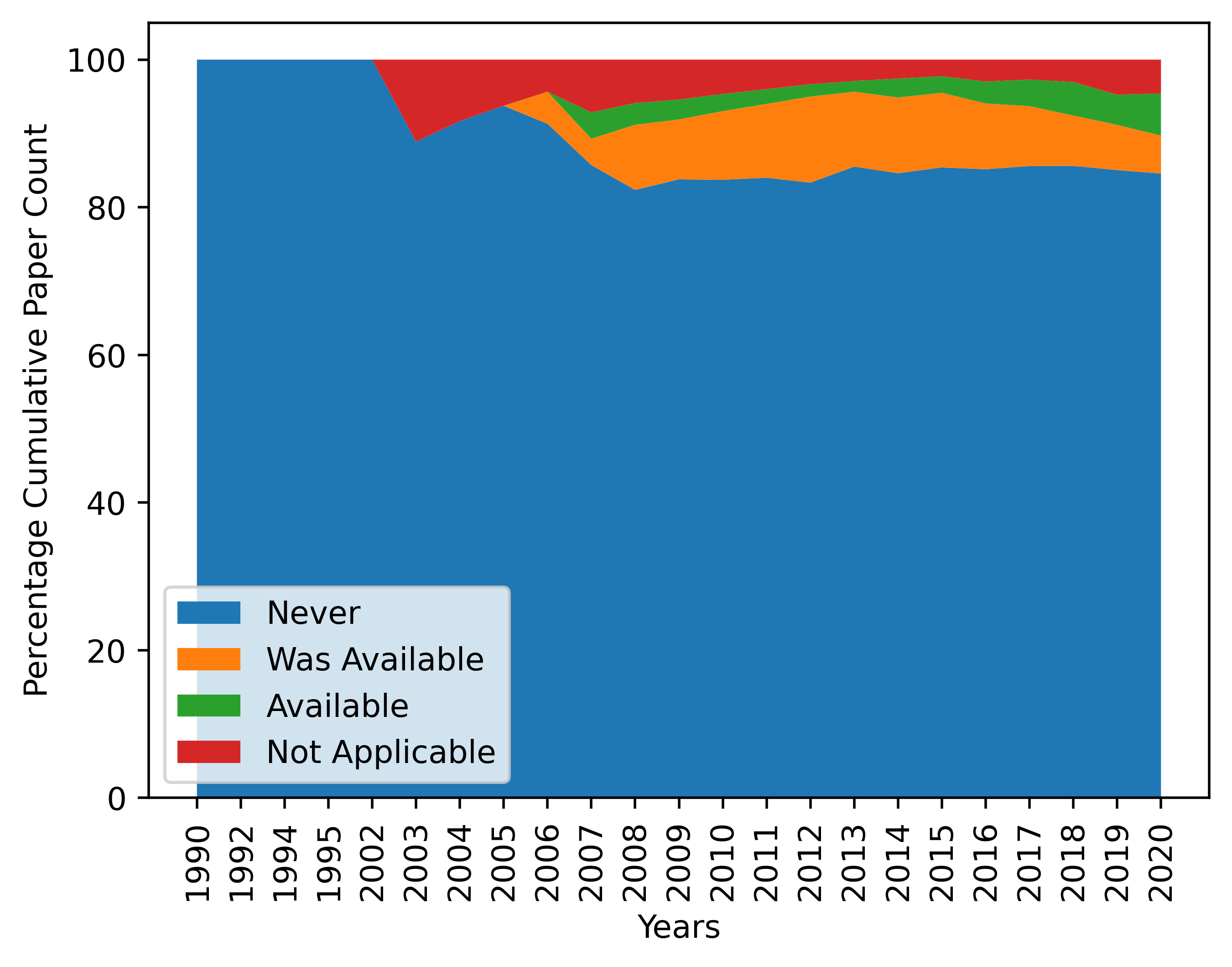}
         \caption{Tool Availability}
    \end{subfigure}
    \begin{subfigure}[!hbt]{\fourth\textwidth}
         \centering
         \includegraphics[width=\textwidth]{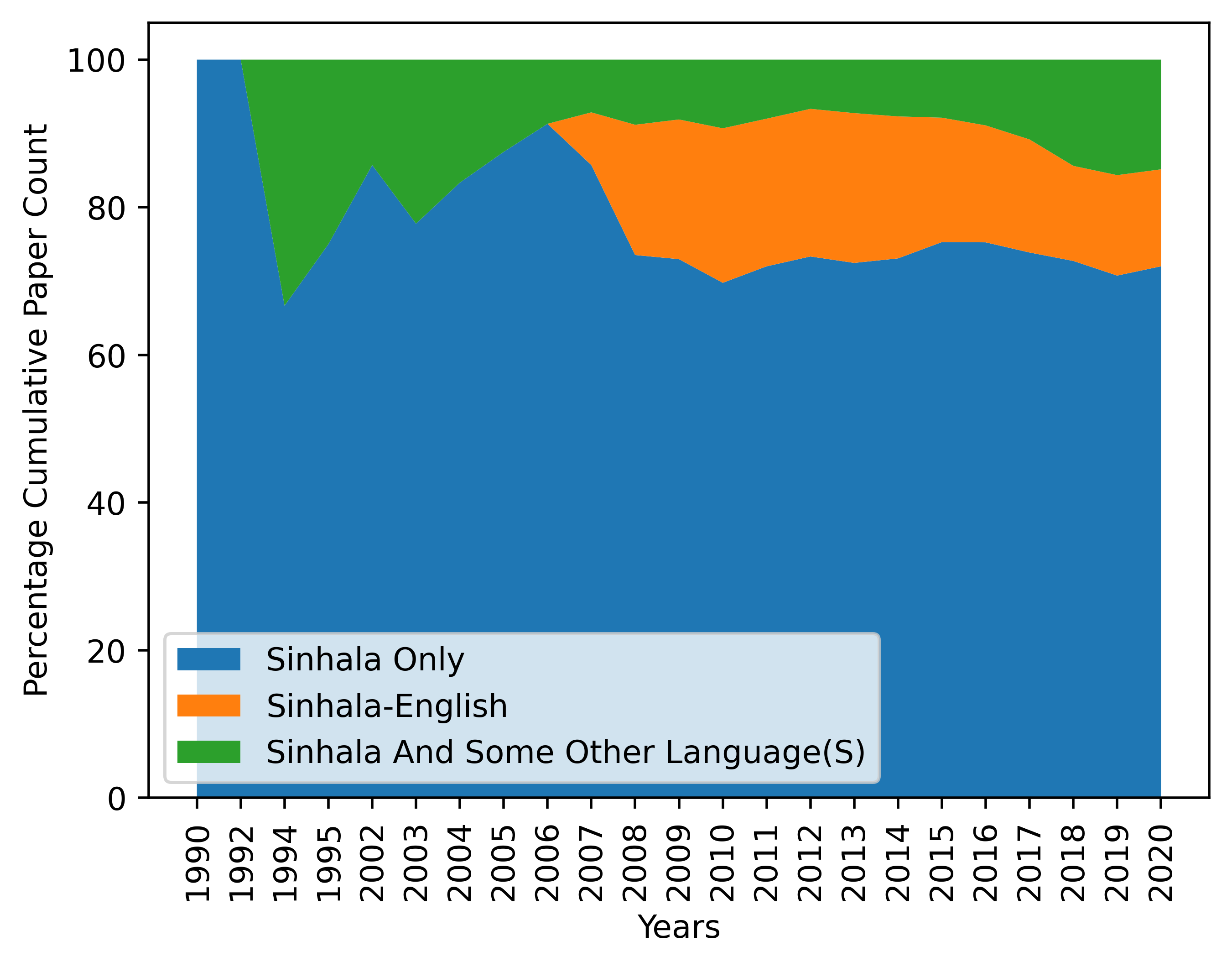}
         \caption{Used Language}
    \end{subfigure}
    \caption{Sinhala NLP Percentage Cumulative analysis from the papers listed by~\citet{de2021survey}}
    \label{fig:survey}
\end{figure*} 

We note a clear relationship between ~\citet{joshi-etal-2020-state} categories, and the Ethnologue classes. As shown in Tables~\ref{tab:confuseWitouthHugging} and~\ref{tab:confuseWithHugging} in Appendix~\ref{sec:HuggingfaceImpact}, all the \textit{Extinct} languages as well a vast majority of \textit{Endangered} languages are in \textit{class 0}  of \citet{joshi-etal-2020-state}'s categorization. On the other hand,~\textit{class 5} languages are all \textit{Large-Institutional}.\\ 
Although both graphs have the same trends, as shown in Figure~\ref{fig:joshiShift} and the discussion in Appendix~\ref{sec:HuggingfaceImpact}, 87 languages have changed their class (84 are promotions) when Huggingface is considered. Interestingly, class of Welsh changes from 1 to 3, and Azerbijanis changes from 1 to 4.
This cautions us not to rely on a hard categorisation based on a partial set of data repositories.

To further explain the limitations of a language categorisation that relies on annotated datasets derived only from a set of repositories, consider Gorontalo and Gujarati languages. Both belong to class 1 in~\citet{joshi-etal-2020-state}'s categorisation. Gorontalo is a mid-endangered language with 1 million L1 speakers. It is not in Google Translate or Facebook language list, nor is it included in pre-trained multilingual models. In contrast, Gujarati is a large institutional language with 56 million L1 speakers. It is included in all of the above three lists. In addition, \verb|gujarati| + \verb|"Natural Language Processing"| query returns 1960 results in Google scholar, and has 189 papers in ACL anthology corpus extracted by~\citet{Rohatgi2022ACL}. %\footnote{using the search option in Anthology website}
 The corresponding query for Gorontalo returns only 81 results, and 0 results in~\citet{Rohatgi2022ACL}'s corpus.~\citet{bird-2022-local} builds a similar argument by comparing Tamil (75 million speakers) and Cree (75,000 speakers).  %This agrees with~\citet{bird-2022-local}'s view - a language like Tamil, ``with 75 million speakers, most of them literate in the language, and a history of written texts that goes back thousands of years'' should not be given the label low-resource along with Cree, which is an Indigenous language in Canada, with 75,000 speakers and few written text. Thus, this highlights that the resourcefulness of a language cannot be condensed into the availability of annotated datasets.

\section{Amount of Research Conducted for Different Languages}
%Similar to~\citet{joshi-etal-2020-state}, we categorised the anthology publications as main conferences as mentioned in ACL Anthology, journals (TACL, CL), LREC, and others workshops and conferences that publish their proceedings in ACL Anthology. 
%As can be seen in the ACL visualization in Figure~\ref{fig:ethnologue_data}, ACL venues also have a less coverage for languages that are not in the group institutional-large. We show the cumulative counts of each category publishing in Main, LREC, journal, and Other ACL venues in Fig.~\ref{fig:ACLpubs}. 

%Now it is time we address the elephant in the room. %From Figure~\ref{fig:ethnologue_data} to~\ref{fig:boxplots} as well as Table~\ref{tab:datasets12} we have included data from the ACL Anthology but have not discussed them until this point. 
%What is the situation of ACL in the question we have discussed so far?
%
We use the research papers published in ACL Anthology curated in~\citet{Rohatgi2022ACL}'s corpus, which contains full papers and their metadata of all Anthology papers upto now\footnote{We extract the full text from the beginning of abstract to the beginning of references excluding acknowledgements.}. Figure~\ref{fig:ethnologue_data_ACL} shows that ACL Anthology, even when considering LREC and workshops associated with ACL, has less coverage for languages other than those belonging to the \textit{Large-Institutional} category. %\footnote{\sr{A full text analysis of the ACL publications following the method of~\citet{blasi-etal-2022-systematic} is also shown by the Fig~\ref{fig:ACLallBlasi} in Appendix~\ref{sec:ACL}}}. 
As further shown in Appendix~\ref{sec:ACL}, research papers in ACL anthology for categories other than \textit{Large-Institutional} category comes mainly from LREC and workshops.  This observation aligns with what~\citet{joshi-etal-2020-state} reported in their conference-language inclusion analysis. However, interestingly, our results show that ACL anthology covers more languages than what has been covered in data sources shown in Fig~\ref{fig:ethnologue_data}. This observation is affirmed by Fig~\ref{fig:boxACL}. While this could mean that datasets are re-used across research, it could mean the data used in these papers might be in personal/institutional repositories, or the data might have not been released at all. 

In order to further validate this hypothesis, we went through a random set of 50 papers extracted from ACL Anthology 2020. However, only 16 papers presented new datasets. Since the number is not enough to conduct a deeper analysis, we extracted the first 100 papers from LREC 2022 proceedings. Our assumption was LREC papers would be more focused on presenting new datasets.  Out of the 56 LREC papers that presented new datasets, only 5 (9\%) have published their data in public repositories. 80\% papers indicated that they have released the data in personal or public repositories. The process to collect this data, as well as the visualizations are given in Appendix~\ref{anthology_lrec-analysis}. 

We also conducted a mini survey (\url{https://forms.gle/FbWhChAeBE5KBvsQ8}) among NLP researchers\footnote{By sending the survey participation request via public mailing lists, private interest groups and personal contacts}.  The survey questions and the responses from 81 participants in 31 countries are given in Appendix~\ref{survey_results}. First and foremost, the results further confirm that categorising languages considering only a few data repositories is misleading, as there are many such repositories - the repository selection depends on personal, as well as institutional choices. It is also interesting to note that there is a noticeable number of respondents who are not aware of such data repositories. It also explains why the language count is higher in ACL Anthology compared to language counts in ELRA/LDC/HuggingFace - researchers mostly prefer to keep their data in their personal repositories.
%Conversely, this also hints that those published research \sr{either keeps the created datasets in personal/institutional repositories or some less known public repositories } , or has not bothered to submit the associated data to public repositories. 
% Just out of curiosity, we selected the languages that have been mentioned in at least one anthology paper, and carried out the Google scholar queries shown in Table~\ref{tab:scolar-anthology}. We wanted to identify the amount of research reported for each language, with respect to NLP in general, as well as for some low-level and high-level NLP tasks. The full results are reported in Appendix. While it is obvious that Google scholar results may have false positives, the difference between ACL numbers and scholar numbers is significant. 

\begin{figure}[!hbt]
     \centering
     \begin{subfigure}[!hbt]{0.49\textwidth}
         \centering
         \includegraphics[height=20pt,keepaspectratio]{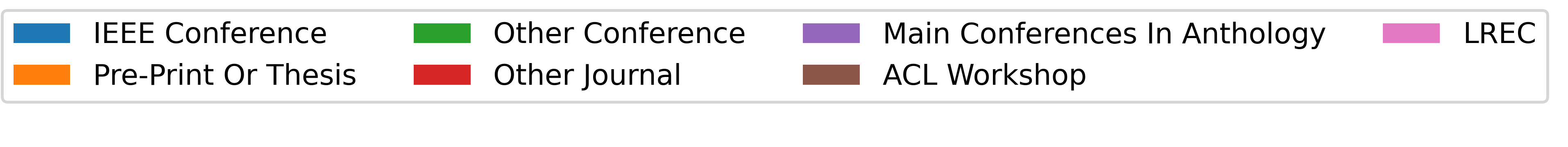}
     \end{subfigure}
     
 %    \begin{subfigure}[!hbt]{\fourthSingle\textwidth}
%         \centering         \includegraphics[width=\textwidth]{images/Sinhala_where_published_with_Axis.png}
%         \caption{Sinhala Count}
%    \end{subfigure}
    \begin{subfigure}[!hbt]{\sixth\textwidth}
         \centering         \includegraphics[width=\textwidth]{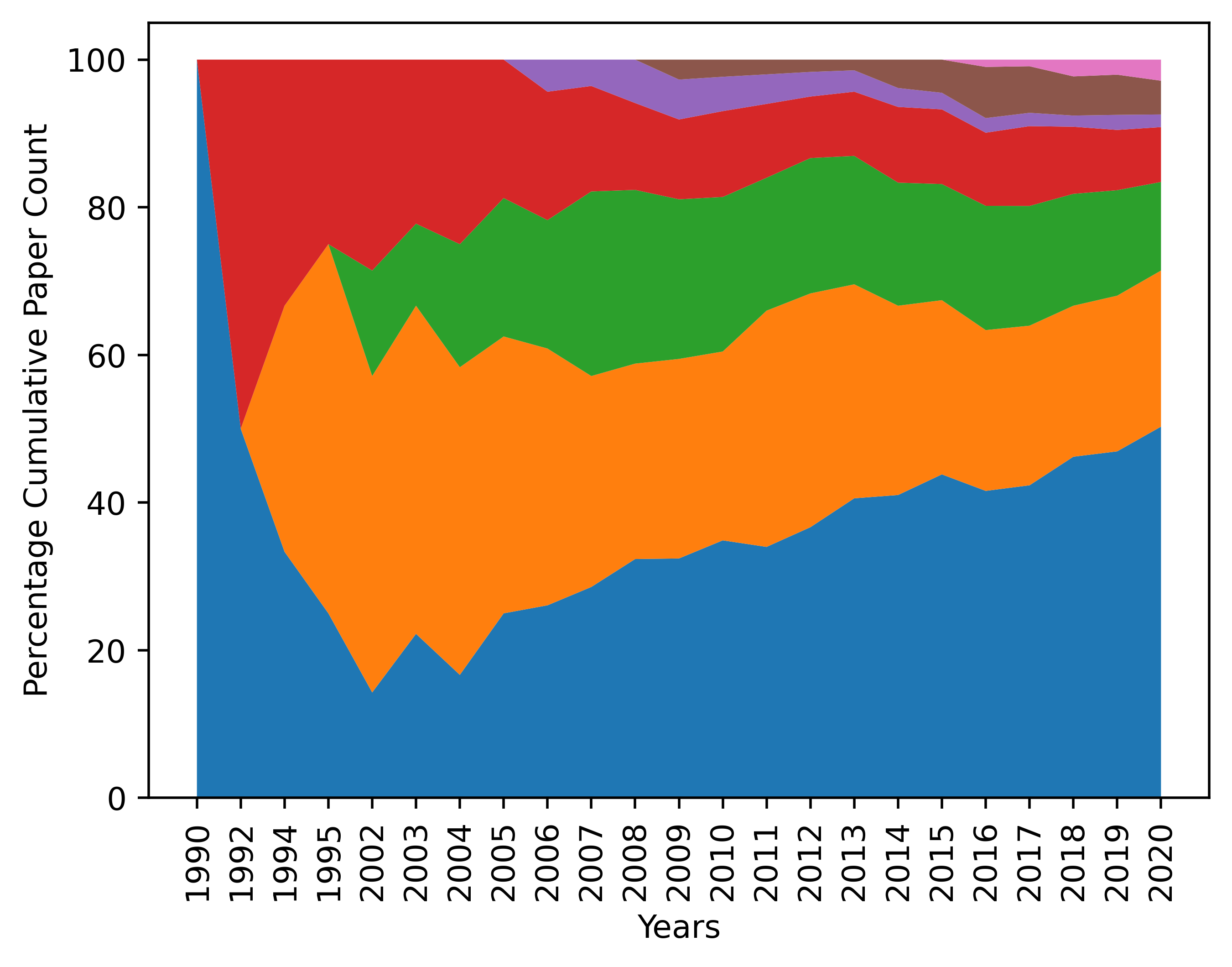}
         \caption{Sinhala}
    \end{subfigure}
%    \begin{subfigure}[!hbt]{\fourthSingle\textwidth}
%         \centering         \includegraphics[width=\textwidth]{images/Hausa_where_published_Axis.png}
 %        \caption{Hausa Count}
%    \end{subfigure}
    \begin{subfigure}[!hbt]{\sixth\textwidth}
         \centering         \includegraphics[width=\textwidth]{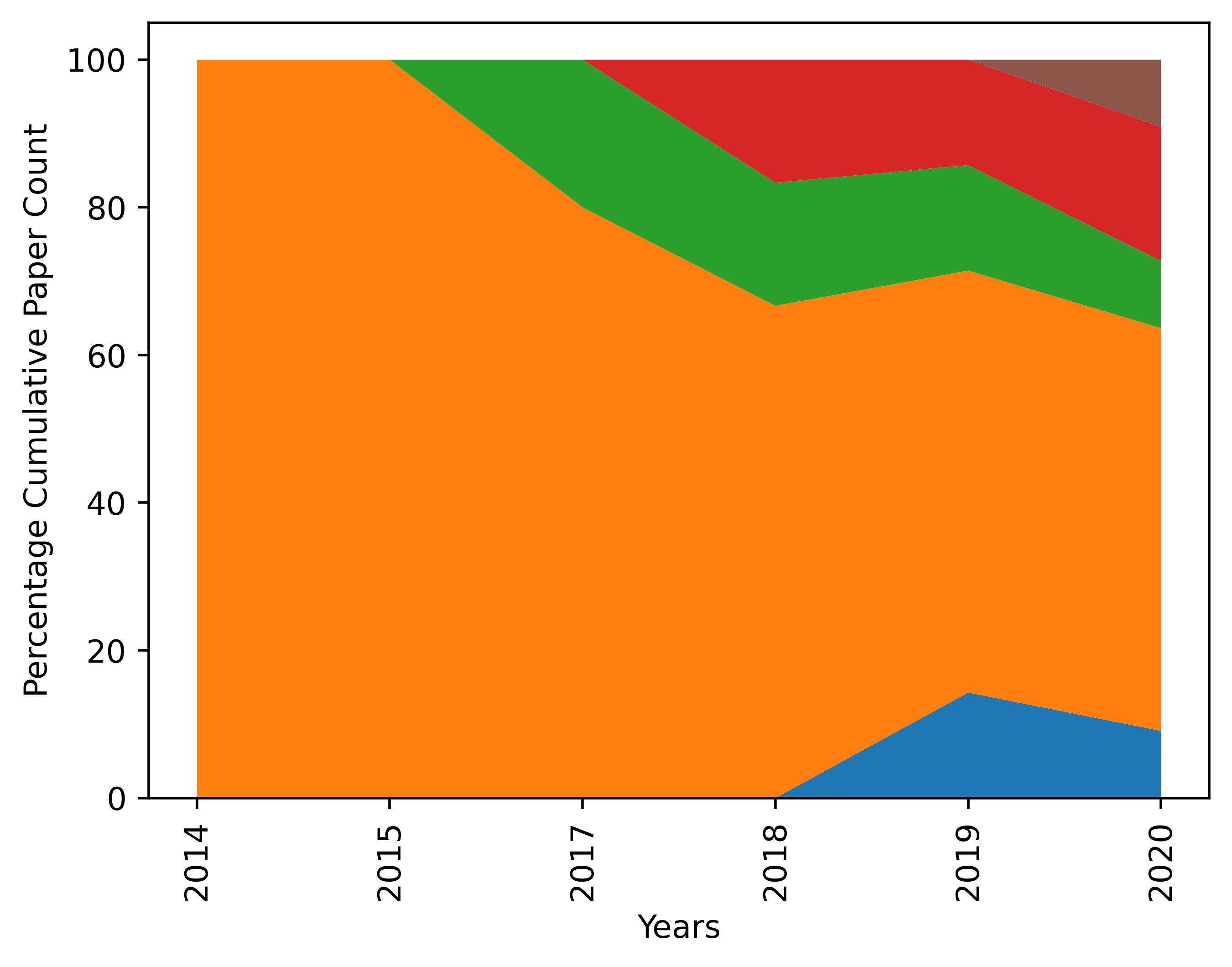}
         \caption{Hausa}
    \end{subfigure}
%    \begin{subfigure}[!hbt]{\fourthSingle\textwidth}
 %        \centering         \includegraphics[width=\textwidth]{images/Sindhi_where_published_Axis.png}
  %       \caption{Sindhi Count}
  %  \end{subfigure}
    \begin{subfigure}[!hbt]{\sixth\textwidth}
         \centering         \includegraphics[width=\textwidth]{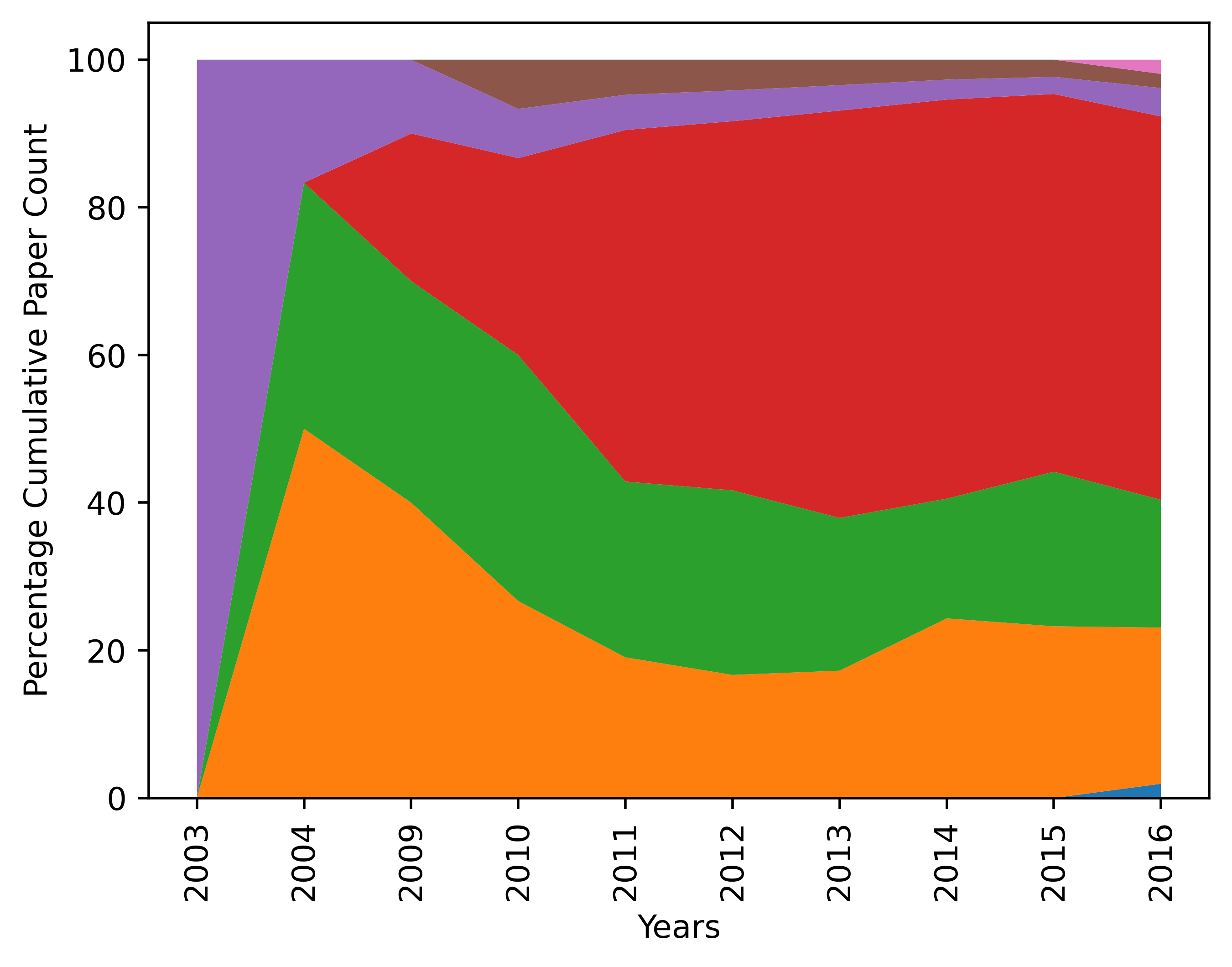}
         \caption{Sindhi}
    \end{subfigure}
    \caption{Cumulative percentage graphs - where the NLP research of each language has been published.}
    \label{fig:threeLang}
\end{figure}

In order to further understand where papers of languages traditionally known as low-resource languages are published, we carried out a language-specific analysis. We identified three survey papers: Sinhala~\cite{de2021survey}, Sindhi~\cite{jamro2017sindhi}, and Hausa~\cite{zakarisystematic} (all are large-institutional languages, with~\citet{joshi-etal-2020-state}'s category being 0, 1 and 2, respectively). 
%These languages have been traditionally known as low-resource languages. 
We noted down the publishing venues of the research papers cited in these surveys.  These results are plotted in Figure~\ref{fig:threeLang}. We see that
%Table~\ref{tab:scolar-anthology} shows the number of entries in ACL anthology and Google scholar, for these selected languages. 
%In addition to the ACL anthology categories identified in Section~\ref{sec:resource_dist_intro},
%we report ACL statistics under \textit{Main Conference in Anthology}, \textit{LREC}, and \textit{ACL workshop}. Apart from those, 
apart from the ACL venues, there are:  IEEE conferences, other conferences (not IEEE or ACL anthology), other journals (not in ACL anthology) and pre-prints/thesis/white papers/reports. While different languages show different patterns (e.g. Sinhala mostly gets published in regional IEEE conferences, while Sindhi gets published in other (regional) journals) there is one common observation - there is extremely low number of papers in anthology, even for LREC and workshops published in ACL Anthology. A further look confirms that most of the other conferences and journals are either local or regional.
Further, we carried out the Google scholar queries shown in Table~\ref{tab:scolar-anthology} in order to identify the amount of research reported for each language, with respect to NLP in general, as well as for some low-level and high-level NLP tasks. %The full results are reported in Appendix~\ref{sec:scholar}. 
While it has been shown that Google scholar results have false positives~\cite{ranathunga2021neural}, the difference between ACL numbers and scholar numbers is significant.

 \begin{table*}[!hbt]
\centering
\renewcommand*{\arraystretch}{\tabSmallComp}
%\begin{tabular}{|lllllll|}
\small
\begin{tabularx}{\textwidth}{XZZZZZZ}
\hline
Language & Anthology & Q1 & Q2 & Q3 & Q4 & Q5 \\
\hline
Hausa & 94 & 779	& 960 &	11 & 123 &	96\\
%\hline
Sindhi & 35 & 653 &	431 &	8 &	86 &	118\\
%\hline
Sinhala & 100 & 1130 &	644 &	14 &	146 &	187\\
\hline

\end{tabularx}
%\end{tabular}
\caption{Amount of research publications for the languages Hausa, Sindhi, and Sinhala. Anthology - number of Anthology papers that mentioned this paper. Q1:  “x”+ “natural language processing”, Q2: “x”+ “part of speech”, Q3: “x”+“grammar parsing”|“grammar parser”,Q4:  “x”+ “question answering”, Q5: “x”+ “text classification”, where Q1-Q5 are Google scholar queries, and x = name of the language.}

\label{tab:scolar-anthology}
%\end{minipage}
\end{table*}

%This observation could be due to several reasons (based on our experience and what other researchers discuss in public forums) : (1) the papers that are focusing on specific languages were not upto the standards of ACL main conferences or workshops\sr{ (or the reviewers do not care about those languages)}, (2) some authors did not know about the ACL venues, (3) some authors self-censored themselves thinking that they might not get in  or (4) some authors could not afford the registration and travel costs to ACL conferences.
%
%Considering the fact that most of the papers appeared in local/regional conferences and journals,
%and \citet{joshi-etal-2020-state}'s observation that recent ACL venues have been more tolerant to this work on low-resource languages,
%the most possible reason for lack of papers in anthology could be the third. %Whatever the reason, this is a worrying fact. This means the low-resource language researchers may be late to learn about the new developments in the field, and might miss the opportunity to interact with the like-minded researchers in the global north.

\section{Case Study: Sinhala}
In~\citet{joshi-etal-2020-state}'s language categorisation, the class of Sinhala is ambiguous - while Sinhala is categorised as class 0, its synonymous term `Sinhalese' is categorised into class 1. Despite its exact category,  Sinhala has been considered a low-resource language even in recent research~\cite{guzman-etal-2019-flores, sarioglu-kayi-etal-2020-detecting}. In contrary, Sinhala has its presence in Wikipedia, Huggingface, Google Translate, Facebook, as well as XLM-R. So why is Sinhala still considered low-resource?

We went through all the Sinhala NLP papers cited in~\citet{de2021survey}'s survey paper to get an idea about the datasets presented in each of the papers, whether the code and data are publicly available and whether any language tool has been released.
Figure~\ref{fig:survey} visualizes this information. Only 11.43\% of papers has data set publicly released (10.29\% in personal repositories, 1.14\% in public repositories) and only 9.71\% of papers have code publicly released. Only 5.71\% have released tools.

Working behind closed doors has shown its negative consequences - within a small time span, two research groups started working on Sinhala WordNet~\cite{welgama2011towards,wijesiri-etal-2014-building}, but none has been successfully completed. Interestingly, none is available to be accessed now. This is common with some other tools that are claimed to be publicly released - they are not accessible. This suggests the lack of infrastructure support to maintain such tools.~\citet{de2021survey}'s author graph highlights another problem - the researchers seem to be working in silos, with almost zero interaction between research groups. 
On the positive side, recently, the use of pre-trained multilingual models has shown its benefit~\cite{rathnayake2022adapter, thillainathan2021fine,dhananjaya-etal-2022-bertifying}. %Interestingly, for text generation, results for Sinhala only lags slightly against English [].

\section{Discussion}
We analysed the linguistic disparity in a global scale. Thus, inevitably, the analysis was limited to only a set of factors, which could be determined by the freely available data. In contrast, the  EU-funded European Language Equality (ELE) project~\cite{grutzner-zahn-rehm-2022-introducing}  categorised European languages with respect to language resources, tools, as well as contextual factors such as economic and financial factors. This analysis is very comprehensive, however, it does not shed any light on the vast majority of the languages in the world. An ambitious project would be to extend this effort in a global scale.  

In order to highlight the importance of carrying out frequent analysis of linguistic disparity, we recorded the number of Wikipedia articles and Huggingface dataset counts as of July 2022. As shown in Tables~\ref{tab:HuggingShift} and \ref{tab:WikiShift} in Appendix~\ref{sec:WikiShift}, 611 new datasets were added to \textit{Large-Institutional} category alone, within less than an year. However, for the small-extinct/endangered/stable/institutional classes altogether, only 9 datasets have been added.  This trend of rich getting richer is a concern as this shows that the average interest still lies with the few languages that are already enjoying an abundance of datasets. As for Wikipedia, an astounding number of articles have been added to \textit{Large-Institutional} category.  Many other language categories have also received articles, suggesting community involvement in content creation. It would be interesting to check whether this content addition impacts the Ethnologue categorisation, however, we lack historical Ethnologue data to conduct this analysis.

We highlighted that the inclusion of a language in a pre-trained multilingual model provides an added advantage for a language. However, not many languages are included in the available models. At least for the languages where text data is there, pre-trained multilingual models should be publicly released. While doing so, models including related languages would be more beneficial~\cite{khanuja2022evaluating, kakwani-etal-2020-indicnlpsuite}.

Many languages are missing in Wikipedia or CommonCrawl. Thus, community engagement should be promoted and funded to improve language-specific Wikipedias. Wikimedia grant scheme is one useful lifeline~\footURL{https://meta.wikimedia.org/wiki/Grants:Start}. ~\citet{bapna2022building} reported the possibility to web-mine data for 1500 languages. We hope this data will be publicly available. For spoken languages that do not have any text~\cite{bird-2022-local}, extra effort is needed to collect speech data. There should be initiatives (preferably funded, for languages in Global South) to create annotated data, even in small quantities, for languages that have monolingual data.% Thus, we are hopeful of the ACL 2022 D\&I Special Initiative\footURL{https://www.2022.aclweb.org/dispecialinitiative}.

Inuktitut, a mid-institutional language with about 40,000 speakers has been recently included in Facebook, with the support from a local learning center~\cite{CBC_News}. This is welcome news - collaborations between locals and tech giants can facilitate the inclusion of languages in the web platforms. However, Inuktitut is a North American language. Adding an African language to Facebook or Google language list may face more challenges.

Not all authors have added data to public repositories, which also have limitations. Particularly, many do not have language or task-wise categorisation of data, and meta data is not collected. We hope ACL can take the initiative to setup a repository that does not have the limitations identified in our survey. A similar initiative is preferable to create an infrastructure to host language tools.

As NLP researchers from Global South, we have our own interpretation of the reasons for many languages having research papers in non-ACL venues. Many reviewers in ACL conferences are sceptical of techniques tested only on a language not popularly known. With time, authors stay away from submitting to these venues, as they anticipate the possible outcome. While there are several workshops welcoming low-resource language research, most of them are non-indexed. This is a concern in institutions that take indexed publications as a measure of academic success. Travelling to ACL venues is  expensive for researchers from the Global South, and many conferences are held in countries with high visa restrictions. Thus, hybrid events with less expensive online versions are a blessing for such researchers.~\citet{blasi-etal-2022-systematic}  found no evidence that research papers dealing with more languages in their evaluation having any advantage over those that do not when considering the number of citations, which means researchers have no incentive to test their systems in many languages. Organising multilingual shared tasks and more recognition for papers presenting multilingual datasets might help alleviating this problem.

Finally, we showed the need to discuss the full situation of languages used in research with respect to the socio-economic status as well as resource availability, rather than saying the language is low-resource just by considering data availability. 

These are  the limitations of this study: The use of language names is not consistent across different data sources. We put every effort to use a uniform language list across data sources, however there can be a few languages that we missed. We used the logic by~\citet{blasi-etal-2022-systematic} to check the existence of a language name in a paper. Thus, the extracted data may have some noise, so does Google scholar search. As already mentioned, task-wise  dataset analysis is extremely noisy. We used information available in the Internet to extract population and GDP information. Therefore this data also could have some noise.

In order to carry out better analysis in the future, we recommend: (1) Creating a map of synonyms of languages, (2) a widely accepted list of NLP tasks, (3) NLP papers adhering to the Bender rule~\cite{bender2019rule} and (4) recording the meta data of the datasets reported in repositories and in research papers (Data statements~\cite{bender2018data} would be a good starting point).

\section{Conclusion}
The objective of this research was to provide a multi-facet analysis of the linguistic disparity in the world. We showed that such an analysis provides a more detailed view of the linguistic disparity, rather than depending on the dataset (particularly annotated) availability. %We showed that this problem is due to socio-economic-linguistic factors.
%, as well as the low-resource language researchers working behind close doors. 
We provided some preliminary recommendations to get these languages out of \textit{low-resourcefulness}, which we hope would be taken positively by the stakeholders. We hope there would be more frequent analysis of this sort. In support of such efforts, we release our code to generate the visualisations shown in this paper as well as the relevant data\footURL{https://bit.ly/AACL2022SomeLanguages}.

\section{Acknowledgement}
This publication was funded by the publication support scheme of the SRC of University of Moratuwa (UoM). We  thank Graham Neubig and Antonis Anastasopoulos for providing their code to process ACL anthology, Shravan Nayak for providing OPUS data counts, Undergraduates of the department of CSE, UoM and survey participants. 

\section{Ethical Impacts (Responsible NLP)}
We employed three workers to manually enter statistics into a spreadsheet. One was an undergraduate, the other two were graduates. One was a male, and the other two were females. However, this demographic information was not recorded, as it is not needed for the task.  We gave them initial instructions verbally over a meeting, and demonstrated the data extraction process. They worked remotely. They were compensated on an hourly rate. Payment rates were according to the approved rates of the university.

The survey was anonymous. We did not collect the email addresses of the participants. The only demographic information we collected was the country of residence. The individual responses have not been publicly released. Only the aggregated results are included in this paper. The participants have discussed limitations of individual data repositories. However, such specific comments are not included in this paper.

The language list we created is publicly available. We mentioned the sources we used to extract data. The limitations in data collection and processing were listed in the discussion. Our code to generate visualisations is publicly available, for the same visualisations to be developed in the future. 

We believe that our study provided valuable insights to the linguistic disparity in a global scale, which would be useful in formulating action plans to mitigate this disparity. 


\bibliographystyle{acl_natbib}
\bibliography{custom,anthology}

\begin{thebibliography}{49}
\expandafter\ifx\csname natexlab\endcsname\relax\def\natexlab#1{#1}\fi

\bibitem[{Anastasopoulos et~al.(2020)Anastasopoulos, Cox, Neubig, and
  Cruz}]{anastasopoulos-etal-2020-endangered}
Antonios Anastasopoulos, Christopher Cox, Graham Neubig, and Hilaria Cruz.
  2020.
\newblock \href {https://doi.org/10.18653/v1/2020.coling-tutorials.7}
  {Endangered languages meet {M}odern {NLP}}.
\newblock In \emph{Proceedings of the 28th International Conference on
  Computational Linguistics: Tutorial Abstracts}, pages 39--45, Barcelona,
  Spain (Online). International Committee for Computational Linguistics.

\bibitem[{Bapna et~al.(2022)Bapna, Caswell, Kreutzer, Firat, van Esch,
  Siddhant, Niu, Baljekar, Garcia, Macherey et~al.}]{bapna2022building}
Ankur Bapna, Isaac Caswell, Julia Kreutzer, Orhan Firat, Daan van Esch, Aditya
  Siddhant, Mengmeng Niu, Pallavi Baljekar, Xavier Garcia, Wolfgang Macherey,
  et~al. 2022.
\newblock Building machine translation systems for the next thousand languages.
\newblock \emph{arXiv preprint arXiv:2205.03983}.

\bibitem[{Bender(2019)}]{bender2019rule}
Emily Bender. 2019.
\newblock \href
  {https://thegradient.pub/the-benderrule-on-naming-the-languages-we-study-and-why-it-matters/}
  {The \#{B}enderrule: {O}n naming the languages we study and why it matters}.
\newblock \emph{The Gradient}, 14.

\bibitem[{Bender and Friedman(2018)}]{bender2018data}
Emily~M Bender and Batya Friedman. 2018.
\newblock Data statements for natural language processing: Toward mitigating
  system bias and enabling better science.
\newblock \emph{Transactions of the Association for Computational Linguistics},
  6:587--604.

\bibitem[{Besacier et~al.(2014)Besacier, Barnard, Karpov, and
  Schultz}]{besacier2014automatic}
Laurent Besacier, Etienne Barnard, Alexey Karpov, and Tanja Schultz. 2014.
\newblock Automatic speech recognition for under-resourced languages: {A}
  survey.
\newblock \emph{Speech Communication}, 56:85--100.

\bibitem[{Bird(2020)}]{bird-2020-decolonising}
Steven Bird. 2020.
\newblock \href {https://doi.org/10.18653/v1/2020.coling-main.313}
  {Decolonising speech and language technology}.
\newblock In \emph{Proceedings of the 28th International Conference on
  Computational Linguistics}, pages 3504--3519, Barcelona, Spain (Online).
  International Committee on Computational Linguistics.

\bibitem[{Bird(2022)}]{bird-2022-local}
Steven Bird. 2022.
\newblock \href {https://doi.org/10.18653/v1/2022.acl-long.539} {Local
  languages, third spaces, and other high-resource scenarios}.
\newblock In \emph{Proceedings of the 60th Annual Meeting of the Association
  for Computational Linguistics (Volume 1: Long Papers)}, pages 7817--7829,
  Dublin, Ireland. Association for Computational Linguistics.

\bibitem[{Blasi et~al.(2022)Blasi, Anastasopoulos, and
  Neubig}]{blasi-etal-2022-systematic}
Damian Blasi, Antonios Anastasopoulos, and Graham Neubig. 2022.
\newblock \href {https://doi.org/10.18653/v1/2022.acl-long.376} {Systematic
  inequalities in language technology performance across the world{'}s
  languages}.
\newblock In \emph{Proceedings of the 60th Annual Meeting of the Association
  for Computational Linguistics (Volume 1: Long Papers)}, pages 5486--5505,
  Dublin, Ireland. Association for Computational Linguistics.

\bibitem[{Cains(2019)}]{cains2019Geographic}
Andrew Cains. 2019.
\newblock The geographic diversity of {NLP} conferences.
\newblock \emph{The Gradient}.

\bibitem[{CBC(2022)}]{CBC_News}
News CBC. 2022.
\newblock `{Reinforces the legitimacy of our language': Inuktitut officially
  available on Facebook desktop}.
\newblock \url{https://bit.ly/3IJraWW}.

\bibitem[{Conneau et~al.(2020)Conneau, Khandelwal, Goyal, Chaudhary, Wenzek,
  Guzm{\'a}n, Grave, Ott, Zettlemoyer, and
  Stoyanov}]{conneau-etal-2020-unsupervised}
Alexis Conneau, Kartikay Khandelwal, Naman Goyal, Vishrav Chaudhary, Guillaume
  Wenzek, Francisco Guzm{\'a}n, Edouard Grave, Myle Ott, Luke Zettlemoyer, and
  Veselin Stoyanov. 2020.
\newblock \href {https://doi.org/10.18653/v1/2020.acl-main.747} {Unsupervised
  cross-lingual representation learning at scale}.
\newblock In \emph{Proceedings of the 58th Annual Meeting of the Association
  for Computational Linguistics}, pages 8440--8451, Online. Association for
  Computational Linguistics.

\bibitem[{Conneau et~al.(2018)Conneau, Rinott, Lample, Williams, Bowman,
  Schwenk, and Stoyanov}]{conneau-etal-2018-xnli}
Alexis Conneau, Ruty Rinott, Guillaume Lample, Adina Williams, Samuel Bowman,
  Holger Schwenk, and Veselin Stoyanov. 2018.
\newblock \href {https://doi.org/10.18653/v1/D18-1269} {{XNLI}: Evaluating
  cross-lingual sentence representations}.
\newblock In \emph{Proceedings of the 2018 Conference on Empirical Methods in
  Natural Language Processing}, pages 2475--2485, Brussels, Belgium.
  Association for Computational Linguistics.

\bibitem[{de~Silva(2021)}]{de2021survey}
Nisansa de~Silva. 2021.
\newblock Survey on publicly available {S}inhala {N}atural {L}anguage
  {P}rocessing tools and research.
\newblock \emph{arXiv preprint arXiv:1906.02358v10}.

\bibitem[{Devlin et~al.(2019)Devlin, Chang, Lee, and
  Toutanova}]{devlin-etal-2019-bert}
Jacob Devlin, Ming-Wei Chang, Kenton Lee, and Kristina Toutanova. 2019.
\newblock \href {https://doi.org/10.18653/v1/N19-1423} {{BERT}: Pre-training of
  deep bidirectional transformers for language understanding}.
\newblock In \emph{Proceedings of the 2019 Conference of the North {A}merican
  Chapter of the Association for Computational Linguistics: Human Language
  Technologies, Volume 1 (Long and Short Papers)}, pages 4171--4186,
  Minneapolis, Minnesota. Association for Computational Linguistics.

\bibitem[{Dhananjaya et~al.(2022)Dhananjaya, Demotte, Ranathunga, and
  Jayasena}]{dhananjaya-etal-2022-bertifying}
Vinura Dhananjaya, Piyumal Demotte, Surangika Ranathunga, and Sanath Jayasena.
  2022.
\newblock \href {https://aclanthology.org/2022.lrec-1.803} {{BERT}ifying
  {S}inhala - a comprehensive analysis of pre-trained language models for
  {S}inhala text classification}.
\newblock In \emph{Proceedings of the Thirteenth Language Resources and
  Evaluation Conference}, pages 7377--7385, Marseille, France. European
  Language Resources Association.

\bibitem[{Eberhard et~al.(2021)Eberhard, Simons, and Fennig}]{ethnologue}
David~M. Eberhard, Gary~F. Simons, and Charles~D. Fennig. 2021.
\newblock \href {https://www.ethnologue.com/} {\emph{Ethnologue: {L}anguages of
  the World}}. Dallas, Texas: SIL International.

\bibitem[{Ebrahimi et~al.(2022)Ebrahimi, Mager, Oncevay, Chaudhary, Chiruzzo,
  Fan, Ortega, Ramos, Rios, Meza~Ruiz, Gim{\'e}nez-Lugo, Mager, Neubig, Palmer,
  Coto-Solano, Vu, and Kann}]{ebrahimi-etal-2022-americasnli}
Abteen Ebrahimi, Manuel Mager, Arturo Oncevay, Vishrav Chaudhary, Luis
  Chiruzzo, Angela Fan, John Ortega, Ricardo Ramos, Annette Rios, Ivan~Vladimir
  Meza~Ruiz, Gustavo Gim{\'e}nez-Lugo, Elisabeth Mager, Graham Neubig, Alexis
  Palmer, Rolando Coto-Solano, Thang Vu, and Katharina Kann. 2022.
\newblock \href {https://doi.org/10.18653/v1/2022.acl-long.435}
  {{A}mericas{NLI}: Evaluating zero-shot natural language understanding of
  pretrained multilingual models in truly low-resource languages}.
\newblock In \emph{Proceedings of the 60th Annual Meeting of the Association
  for Computational Linguistics (Volume 1: Long Papers)}, pages 6279--6299,
  Dublin, Ireland. Association for Computational Linguistics.

\bibitem[{Faisal et~al.(2022)Faisal, Wang, and
  Anastasopoulos}]{faisal-etal-2022-dataset}
Fahim Faisal, Yinkai Wang, and Antonios Anastasopoulos. 2022.
\newblock \href {https://doi.org/10.18653/v1/2022.acl-long.239} {Dataset
  geography: Mapping language data to language users}.
\newblock In \emph{Proceedings of the 60th Annual Meeting of the Association
  for Computational Linguistics (Volume 1: Long Papers)}, pages 3381--3411,
  Dublin, Ireland. Association for Computational Linguistics.

\bibitem[{Gr{\"u}tzner-Zahn and
  Rehm(2022)}]{grutzner-zahn-rehm-2022-introducing}
Annika Gr{\"u}tzner-Zahn and Georg Rehm. 2022.
\newblock \href {https://aclanthology.org/2022.tdle-1.2} {Introducing the
  digital language equality metric: Contextual factors}.
\newblock In \emph{Proceedings of the Workshop Towards Digital Language
  Equality within the 13th Language Resources and Evaluation Conference}, pages
  13--26, Marseille, France. European Language Resources Association.

\bibitem[{Guzm{\'a}n et~al.(2019)Guzm{\'a}n, Chen, Ott, Pino, Lample, Koehn,
  Chaudhary, and Ranzato}]{guzman-etal-2019-flores}
Francisco Guzm{\'a}n, Peng-Jen Chen, Myle Ott, Juan Pino, Guillaume Lample,
  Philipp Koehn, Vishrav Chaudhary, and Marc{'}Aurelio Ranzato. 2019.
\newblock \href {https://doi.org/10.18653/v1/D19-1632} {The {FLORES} evaluation
  datasets for low-resource machine translation: {N}epali{--}{E}nglish and
  {S}inhala{--}{E}nglish}.
\newblock In \emph{Proceedings of the 2019 Conference on Empirical Methods in
  Natural Language Processing and the 9th International Joint Conference on
  Natural Language Processing (EMNLP-IJCNLP)}, pages 6098--6111, Hong Kong,
  China. Association for Computational Linguistics.

\bibitem[{Hedderich et~al.(2021)Hedderich, Lange, Adel, Str{\"o}tgen, and
  Klakow}]{hedderich-etal-2021-survey}
Michael~A. Hedderich, Lukas Lange, Heike Adel, Jannik Str{\"o}tgen, and
  Dietrich Klakow. 2021.
\newblock \href {https://doi.org/10.18653/v1/2021.naacl-main.201} {A survey on
  recent approaches for natural language processing in low-resource scenarios}.
\newblock In \emph{Proceedings of the 2021 Conference of the North American
  Chapter of the Association for Computational Linguistics: Human Language
  Technologies}, pages 2545--2568, Online. Association for Computational
  Linguistics.

\bibitem[{Hoenen and Rahn(2021)}]{hoenen2021migration}
Armin Hoenen and Marc~D Rahn. 2021.
\newblock Migration of small and endangered languages into the {W}ikipedia.
\newblock In \emph{Proceedings of the Workshop on Computational Methods for
  Endangered Languages}, volume~2, pages 41--47.

\bibitem[{Hu et~al.(2020)Hu, Ruder, Siddhant, Neubig, Firat, and
  Johnson}]{hu2020xtreme}
Junjie Hu, Sebastian Ruder, Aditya Siddhant, Graham Neubig, Orhan Firat, and
  Melvin Johnson. 2020.
\newblock Xtreme: {A} massively multilingual multi-task benchmark for
  evaluating cross-lingual generalisation.
\newblock In \emph{International Conference on Machine Learning}, pages
  4411--4421. PMLR.

\bibitem[{Jamro(2017)}]{jamro2017sindhi}
Wazir~Ali Jamro. 2017.
\newblock Sindhi language processing: {A} survey.
\newblock In \emph{2017 International Conference on Innovations in Electrical
  Engineering and Computational Technologies (ICIEECT)}, pages 1--8. IEEE.

\bibitem[{Joshi et~al.(2020)Joshi, Santy, Budhiraja, Bali, and
  Choudhury}]{joshi-etal-2020-state}
Pratik Joshi, Sebastin Santy, Amar Budhiraja, Kalika Bali, and Monojit
  Choudhury. 2020.
\newblock \href {https://doi.org/10.18653/v1/2020.acl-main.560} {The state and
  fate of linguistic diversity and inclusion in the {NLP} world}.
\newblock In \emph{Proceedings of the 58th Annual Meeting of the Association
  for Computational Linguistics}, pages 6282--6293, Online. Association for
  Computational Linguistics.

\bibitem[{Kakwani et~al.(2020)Kakwani, Kunchukuttan, Golla, N.C.,
  Bhattacharyya, Khapra, and Kumar}]{kakwani-etal-2020-indicnlpsuite}
Divyanshu Kakwani, Anoop Kunchukuttan, Satish Golla, Gokul N.C., Avik
  Bhattacharyya, Mitesh~M. Khapra, and Pratyush Kumar. 2020.
\newblock \href {https://doi.org/10.18653/v1/2020.findings-emnlp.445}
  {{I}ndic{NLPS}uite: Monolingual corpora, evaluation benchmarks and
  pre-trained multilingual language models for {I}ndian languages}.
\newblock In \emph{Findings of the Association for Computational Linguistics:
  EMNLP 2020}, pages 4948--4961, Online. Association for Computational
  Linguistics.

\bibitem[{Khanuja et~al.(2022)Khanuja, Ruder, and
  Talukdar}]{khanuja2022evaluating}
Simran Khanuja, Sebastian Ruder, and Partha Talukdar. 2022.
\newblock Evaluating inclusivity, equity, and accessibility of {NLP}
  technology: A case study for {I}ndian languages.
\newblock \emph{arXiv preprint arXiv:2205.12676}.

\bibitem[{Kreutzer et~al.(2022)Kreutzer, Caswell, Wang, Wahab, van Esch,
  Ulzii-Orshikh, Tapo, Subramani, Sokolov, Sikasote
  et~al.}]{caswell2021quality}
Julia Kreutzer, Isaac Caswell, Lisa Wang, Ahsan Wahab, Daan van Esch,
  Nasanbayar Ulzii-Orshikh, Allahsera Tapo, Nishant Subramani, Artem Sokolov,
  Claytone Sikasote, et~al. 2022.
\newblock \href {https://doi.org/10.1162/tacl_a_00447} {{Quality at a Glance:
  An Audit of Web-Crawled Multilingual Datasets}}.
\newblock \emph{Transactions of the Association for Computational Linguistics},
  10:50--72.

\bibitem[{Lauscher et~al.(2020)Lauscher, Ravishankar, Vuli{\'c}, and
  Glava{\v{s}}}]{lauscher-etal-2020-zero}
Anne Lauscher, Vinit Ravishankar, Ivan Vuli{\'c}, and Goran Glava{\v{s}}. 2020.
\newblock \href {https://doi.org/10.18653/v1/2020.emnlp-main.363} {From zero to
  hero: {O}n the limitations of zero-shot language transfer with multilingual
  {T}ransformers}.
\newblock In \emph{Proceedings of the 2020 Conference on Empirical Methods in
  Natural Language Processing (EMNLP)}, pages 4483--4499, Online. Association
  for Computational Linguistics.

\bibitem[{Lewis and Simons(2010)}]{lewis2010assessing}
M~Paul Lewis and Gary~F Simons. 2010.
\newblock Assessing endangerment: {E}xpanding {F}ishman’s {GIDS}.

\bibitem[{META-NET(2020)}]{meta2021white}
META-NET. 2020.
\newblock {META-NET} white paper series: {K}ey results and cross-language
  comparison.
\newblock \emph{META}.

\bibitem[{Muller et~al.(2021)Muller, Anastasopoulos, Sagot, and
  Seddah}]{muller-etal-2021-unseen}
Benjamin Muller, Antonios Anastasopoulos, Beno{\^\i}t Sagot, and Djam{\'e}
  Seddah. 2021.
\newblock \href {https://doi.org/10.18653/v1/2021.naacl-main.38} {When being
  unseen from m{BERT} is just the beginning: Handling new languages with
  multilingual language models}.
\newblock In \emph{Proceedings of the 2021 Conference of the North American
  Chapter of the Association for Computational Linguistics: Human Language
  Technologies}, pages 448--462, Online. Association for Computational
  Linguistics.

\bibitem[{Orwell(1945)}]{orwell1945animal}
George Orwell. 1945.
\newblock \emph{{Animal Farm: A Fairy Story}}.
\newblock Secker and Warburg, London, England.

\bibitem[{Paranjape et~al.(2016)Paranjape, West, Zia, and
  Leskovec}]{paranjape2016improving}
Ashwin Paranjape, Robert West, Leila Zia, and Jure Leskovec. 2016.
\newblock Improving website hyperlink structure using server logs.
\newblock In \emph{Proceedings of the Ninth ACM International Conference on Web
  Search and Data Mining}, pages 615--624.

\bibitem[{Perez-Rosas et~al.(2020)Perez-Rosas, Kuo, Herman, Mihalcea
  et~al.}]{perez2020upstage}
Veronica Perez-Rosas, Shihchen Kuo, William~H Herman, Rada Mihalcea, et~al.
  2020.
\newblock {UPSTAGE}: {U}nsupervised context augmentation for utterance
  classification in patient-provider communication.
\newblock In \emph{Machine Learning for Healthcare Conference}, pages 895--912.
  PMLR.

\bibitem[{Ranathunga et~al.(2021)Ranathunga, Lee, Skenduli, Shekhar, Alam, and
  Kaur}]{ranathunga2021neural}
Surangika Ranathunga, En-Shiun~Annie Lee, Marjana~Prifti Skenduli, Ravi
  Shekhar, Mehreen Alam, and Rishemjit Kaur. 2021.
\newblock {N}eural {M}achine {T}ranslation for low-resource languages: {A}
  survey.
\newblock \emph{arXiv preprint arXiv:2106.15115}.

\bibitem[{Rathnayake et~al.(2022)Rathnayake, Sumanapala, Rukshani, and
  Ranathunga}]{rathnayake2022adapter}
Himashi Rathnayake, Janani Sumanapala, Raveesha Rukshani, and Surangika
  Ranathunga. 2022.
\newblock Adapter based fine-tuning of pre-trained multilingual language models
  for code-mixed and code-switched text classification.

\bibitem[{Ratnayaka et~al.(2020)Ratnayaka, de~Silva, Perera, and
  Pathirana}]{ratnayaka-etal-2020-effective}
Gathika Ratnayaka, Nisansa de~Silva, Amal~Shehan Perera, and Ramesh Pathirana.
  2020.
\newblock \href {https://aclanthology.org/2020.paclic-1.29} {Effective approach
  to develop a sentiment annotator for legal domain in a low resource setting}.
\newblock In \emph{Proceedings of the 34th Pacific Asia Conference on Language,
  Information and Computation}, pages 252--260, Hanoi, Vietnam. Association for
  Computational Linguistics.

\bibitem[{Ray~Chowdhury et~al.(2019)Ray~Chowdhury, Caragea, and
  Caragea}]{ray2019keyphrase}
Jishnu Ray~Chowdhury, Cornelia Caragea, and Doina Caragea. 2019.
\newblock Keyphrase extraction from disaster-related tweets.
\newblock In \emph{The world wide web conference}, pages 1555--1566.

\bibitem[{Rohatgi(2022)}]{Rohatgi2022ACL}
Shaurya Rohatgi. 2022.
\newblock \href {https://github.com/shauryr/ACL-anthology-corpus} {{ACL
  Anthology Corpus with Full Text}}.
\newblock Github.

\bibitem[{Sarioglu~Kayi et~al.(2020)Sarioglu~Kayi, Nan, Qu, Diab, and
  McKeown}]{sarioglu-kayi-etal-2020-detecting}
Efsun Sarioglu~Kayi, Linyong Nan, Bohan Qu, Mona Diab, and Kathleen McKeown.
  2020.
\newblock \href {https://doi.org/10.18653/v1/2020.coling-main.414} {Detecting
  urgency status of crisis tweets: A transfer learning approach for low
  resource languages}.
\newblock In \emph{Proceedings of the 28th International Conference on
  Computational Linguistics}, pages 4693--4703, Barcelona, Spain (Online).
  International Committee on Computational Linguistics.

\bibitem[{Schwartz et~al.(2019)Schwartz, Chen, Hunt, and
  Schreiner}]{schwartz-etal-2019-bootstrapping}
Lane Schwartz, Emily Chen, Benjamin Hunt, and Sylvia~L.R. Schreiner. 2019.
\newblock \href {https://aclanthology.org/W19-6012} {Bootstrapping a neural
  morphological analyzer for {S}t. {L}awrence {I}sland {Y}upik from a
  finite-state transducer}.
\newblock In \emph{Proceedings of the 3rd Workshop on the Use of Computational
  Methods in the Study of Endangered Languages Volume 1 (Papers)}, pages
  87--96, Honolulu. Association for Computational Linguistics.

\bibitem[{Taghipour and Ng(2016)}]{taghipour-ng-2016-neural}
Kaveh Taghipour and Hwee~Tou Ng. 2016.
\newblock \href {https://doi.org/10.18653/v1/D16-1193} {A neural approach to
  automated essay scoring}.
\newblock In \emph{Proceedings of the 2016 Conference on Empirical Methods in
  Natural Language Processing}, pages 1882--1891, Austin, Texas. Association
  for Computational Linguistics.

\bibitem[{Thillainathan et~al.(2021)Thillainathan, Ranathunga, and
  Jayasena}]{thillainathan2021fine}
Sarubi Thillainathan, Surangika Ranathunga, and Sanath Jayasena. 2021.
\newblock Fine-tuning self-supervised multilingual sequence-to-sequence models
  for extremely low-resource {NMT}.
\newblock In \emph{2021 Moratuwa Engineering Research Conference (MERCon)},
  pages 432--437. IEEE.

\bibitem[{Tiedemann and Thottingal(2020)}]{tiedemann-thottingal-2020-opus}
J{\"o}rg Tiedemann and Santhosh Thottingal. 2020.
\newblock \href {https://aclanthology.org/2020.eamt-1.61} {{OPUS}-{MT} {--}
  building open translation services for the world}.
\newblock In \emph{Proceedings of the 22nd Annual Conference of the European
  Association for Machine Translation}, pages 479--480, Lisboa, Portugal.
  European Association for Machine Translation.

\bibitem[{Welgama et~al.(2011)Welgama, Herath, Liyanage, Udalamatta,
  Weerasinghe, and Jayawardana}]{welgama2011towards}
Viraj Welgama, Dulip~Lakmal Herath, Chamila Liyanage, Namal Udalamatta, Ruvan
  Weerasinghe, and Tissa Jayawardana. 2011.
\newblock Towards a {S}inhala {W}ordnet.
\newblock In \emph{Proceedings of the Conference on Human Language Technology
  for Development}.

\bibitem[{Wijesiri et~al.(2014)Wijesiri, Gallage, Gunathilaka, Lakjeewa,
  Wimalasuriya, Dias, Paranavithana, and
  de~Silva}]{wijesiri-etal-2014-building}
Indeewari Wijesiri, Malaka Gallage, Buddhika Gunathilaka, Madhuranga Lakjeewa,
  Daya Wimalasuriya, Gihan Dias, Rohini Paranavithana, and Nisansa de~Silva.
  2014.
\newblock \href {https://aclanthology.org/W14-0114} {Building a {W}ord{N}et for
  {S}inhala}.
\newblock In \emph{Proceedings of the Seventh Global {W}ordnet Conference},
  pages 100--108, Tartu, Estonia. University of Tartu Press.

\bibitem[{Wu and Dredze(2020)}]{wu-dredze-2020-languages}
Shijie Wu and Mark Dredze. 2020.
\newblock \href {https://doi.org/10.18653/v1/2020.repl4nlp-1.16} {Are all
  languages created equal in multilingual {BERT}?}
\newblock In \emph{Proceedings of the 5th Workshop on Representation Learning
  for NLP}, pages 120--130, Online. Association for Computational Linguistics.

\bibitem[{Zakari et~al.(2021)Zakari, Lawal, and Abdulmumin}]{zakarisystematic}
Rufai~Yusuf Zakari, Zaharaddeen~Karami Lawal, and Idris Abdulmumin. 2021.
\newblock A systematic literature review of {H}ausa {N}atural {L}anguage
  {P}rocessing.

\end{thebibliography}

\appendix

\clearpage
\section{Language List used in the Study}
\label{sec:ehtnologue}

When looking at the list of languages used by \citet{joshi-etal-2020-state}, we noticed that it was quite inconsistent. It had dialects and alternate names of languages as separate entities. For example, it contained \textit{Sinhala} as well as \textit{Sinhalese}. The former is the correct name of the language. The latter is the name of the ethnicity of the people who speak \textit{Sinhala}. While there are online sources that erroneously use \textit{Sinhalese} as the name of the language, it would not suit a research on language to use this term. In addition to that, this also meant that the resources listed for the \textit{Sinhala} are distributed among the two alternate names. This resulted in \citet{joshi-etal-2020-state} categorising \textit{Sinhala} as a class 0 language and \textit{Sinhalese} as a class 1 language. Moreover,~\citet{joshi-etal-2020-state}'s list covers less than half of the languages in the world. Shortfalls such as this motivated us to look elsewhere for a more reliable and consistent source for creating our language list.

We used Ethnologue as our primary source for creating the language list. They list information on 7139 living language entries\footURL{https://www.ethnologue.com/browse/names} in the world, including dialects. 
Ethnologue also lists some dialects and minor schisms within languages as separate entities. However, they are consistent in reporting them. For example, for \textit{German}, they cleanly list \textit{German, Pennsylvania}, \textit{German, Standard}, and \textit{German, Swiss}. Thus, when we were collecting language names from them, we could simply take the term that precedes the comma. 

While this was an efficient strategy to automatically reduce dependencies, when we proceeded to prepare data sets as explained in Appendix~\ref{sec:dataset} with the `list of Wikipedias'\footURL{https://bit.ly/Wikipedias_Details_table}, it was evident that some cases that are represented as a single language in Ethnologue has multiple entries in Wikipedia due to them being functionally distinct. An example of this is \textit{Norwegian}, which has only one entry in Ethnologue\footURL{https://www.ethnologue.com/language/nor} but separate Wikipedias for \textit{Norwegian (Bokmål)}\footURL{https://no.wikipedia.org/wiki/} and \textit{Norwegian (Nynorsk)}\footURL{https://nn.wikipedia.org/wiki/}. In these cases, we added distinct entries for the differing languages. When a singular language in Ethnologue was split this way, the resultant languages were given the class of the source language. Given that all such splits (rather predictably) happened with \textit{Large} languages, the margin of error is still within safe values given the vast difference between the threshold value for the \textit{Large} class and the \textit{Mid} class. Some languages have multiple names, and there were instances where different data sources were using different names. When a language in (say) Wikipedia was not is Ethnologue, we did a web search to check for the alternative names. We used the Ethnologue version of language names.

After these steps we compiled a list of 6420 unique languages to derive our language list, which we have made publicly available
%\footURL{https://anonymous.4open.science/r/AACL_2022_Language_List-000D/}
\footURL{https://bit.ly/AACL2022LangList}
for the benefit of future language researchers.

\section{Dataset Preparation}
\label{sec:dataset}
The `list of Wikipedias' page in Wikipedia  records the statistics of wiki pages in different languages\footURL{https://bit.ly/Wikipedias_Details_table}. We manually recorded the number of Wikipedia articles per language, according to this wiki page. CommonCrawl also has explicitly listed the number of HTML web pages per language\footURL{https://commoncrawl.github.io/cc-crawl-statistics/plots/language}, which we manually recorded. We manually recorded the dataset statistics from LDC, ELRA and Huggingface. In all these repositories, datasets are grouped by language. 

The L1 speakers for a language was extracted from the infobox\footURL{https://en.Wikipedia.org/wiki/Help:Infobox} of the corresponding Wikipedia page. There were few cases, where for some small languages, the number of L1 speakers were not mentioned in the infobox but were mentioned somewhere in the body text. This information was meticulously and manually gathered. 
The total speaker counts for the Language GDP in Billions of Dollars (log) vs Wikipedia Article Count (log) analysis shown in Figure~\ref{fig:GDPvsWiki}, as already mentioned in the main body text of this paper, were mainly collected from the publicly available website \textit{worlddata}\footURL{https://www.worlddata.info/} along with the corresponding information on GDP and percentage of language speakers of each country.
The Ethnologue size (\textit{Large}, \textit{Mid}, and \textit{Small}) as well as the Ethnologue Vitality (\textit{Institutional}, \textit{Stable}, \textit{Endangered}, and \textit{Extinct}) were of course, manually collected from Ethnologue. The language family information as well as the geographical origin of the languages were also collected from the Wikipedia infoboxes of the relevant languages. The count of ACL publications mentioning the relevant language was obtained executing the algorithm proposed by~\citet{blasi-etal-2022-systematic} on the full ACL text dataset published by~\citet{Rohatgi2022ACL}.
%processing the \verb|anthology.bib| file\footURL{http://aclweb.org/anthology/anthology.bib.gz} that is publicly available. 
The \textit{Huggingface} dataset counts for both November 2021 and July 2022 were manually collected from the \textit{Huggingface} dataset search web interface\footURL{https://huggingface.co/datasets}. 

Facebook language list was manually extracted according to the instructions in their Help Centre web page\footURL{https://www.facebook.com/help/327850733950290}. The language list supported by Google was manually extracted from the Google Translate web page~\footURL{https://translate.google.com/intl/en/about/languages/}. We selected the statistics in the `Type' column'.~\citet{conneau-etal-2020-unsupervised} has reported the list of languages covered in XLm-R. mBERT langauge list was manually extracted from its github repository\footURL{https://github.com/google-research/bert/blob/master/multilingual.md}. %This is affirmed by Fig~\ref{fig:cc_box}. Within class distribution looks similar to that of Wikipedia.

\section{CommonCrawl Analysis}
\label{sec:Cc}

\begin{figure}[!hbt]
\centering
\includegraphics[width=\halfSingle\textwidth]{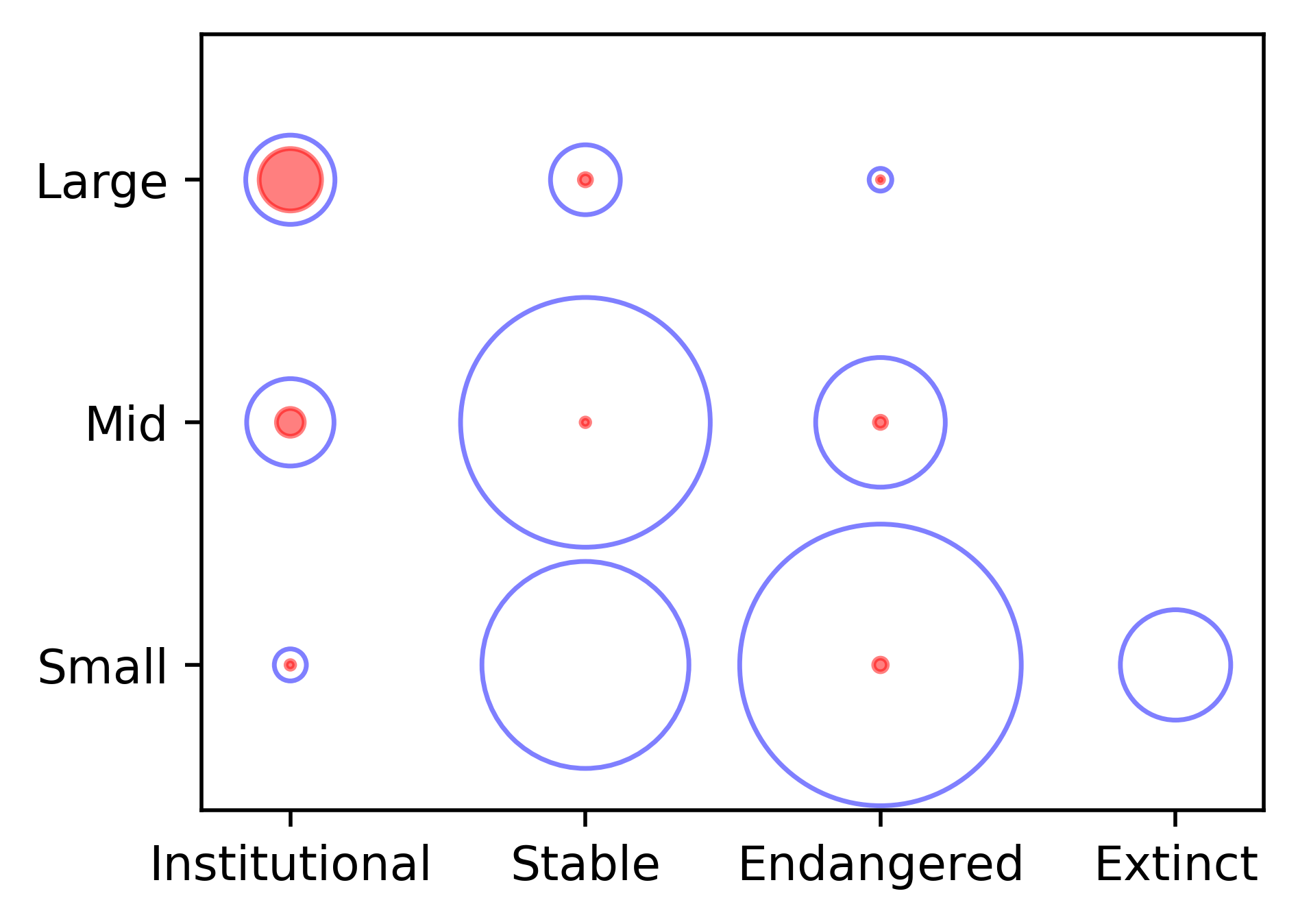}
         \caption{The 12 Ethnologue language classes where the size of each blue circle corresponds to the number of languages in that category and the size of each red circle corresponds to the coverage of that class in CommonCrawl.}
         \label{fig:cc_blob}
\end{figure}

\begin{figure}[!hbt]
\centering
\centering         \includegraphics[width=\halfSingle\textwidth]{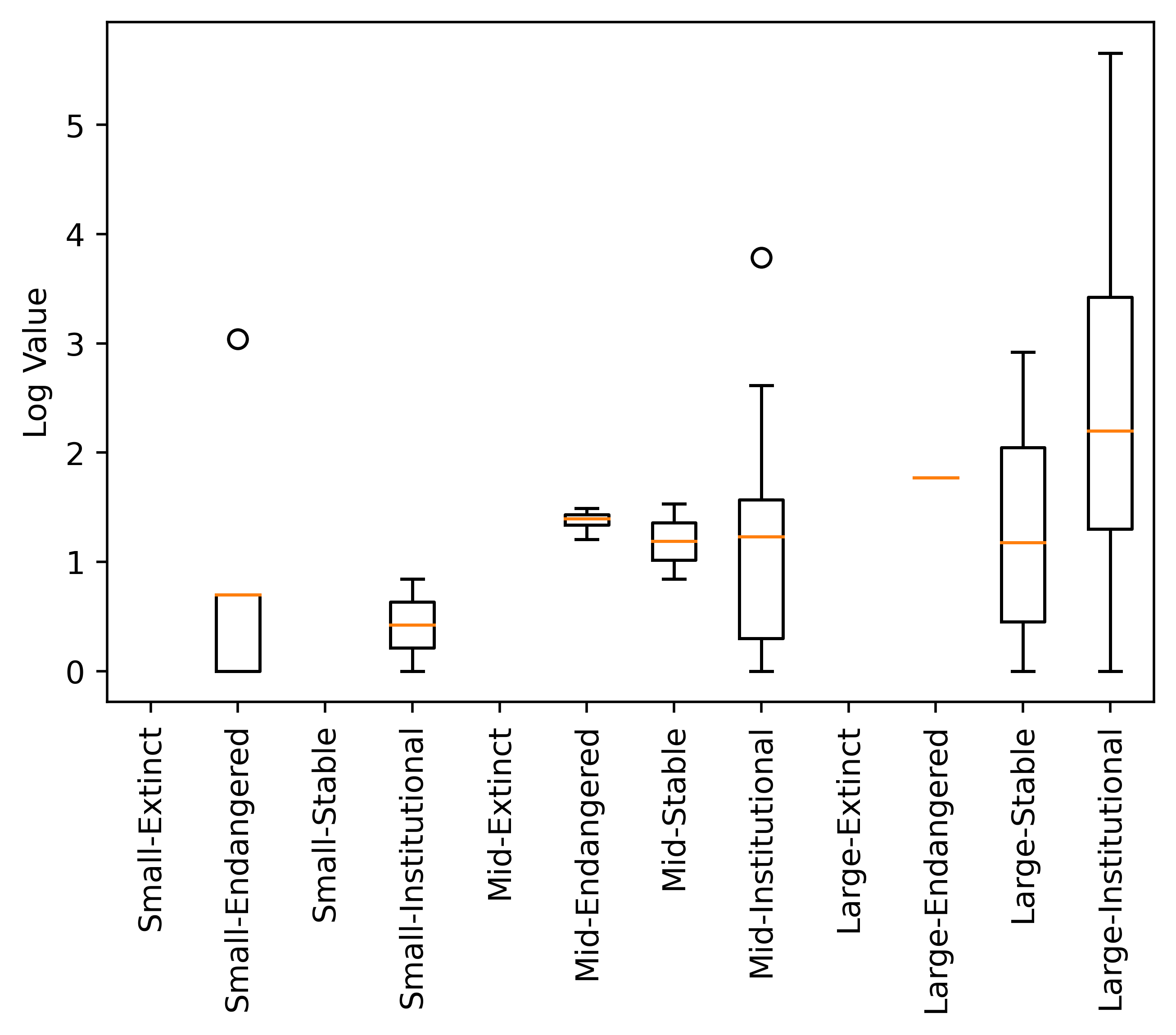}
         \caption{Boxplot showing CommonCrawl data with the amounts corresponding to the 12 Ethnologue language classes.}
         \label{fig:cc_box}
\end{figure}

As shown in Figure~\ref{fig:cc_blob}, CommonCrawl also covers mainly \textit{large-institutional} and \textit{mid-institutional} languages. Some language categories have no presence at all. Table~\ref{tab:cc12} shows the gravity of this problem - out of the 160 languages present in CommonCrawl, 100 come from \textit{large-institutional} category alone. Even \textit{large-endangered} and \textit{large-stable} categories do not have a significant presence in the web, despite a large population using those languages. This behaviour continues to Fig~\ref{fig:cc_box} where it can be observed that other than \textit{Large-Institutional}, all other classes display a disappointing spread.

%Figure~\ref{fig:cc_blob} continues the pattern that we observed in Fig~\ref{fig:ethnologue_data}. We can observe that the \textit{Large-Institutional} languages have near perfect coverage in CommonCrawl while the \textit{Small-Endangered} languages despite being relatively numerous, has extremely sparse coverage.

\begin{table}[!h]
\centering
%\begin{tabular}{lrrrrrrrr}
\begin{tabularx}{\halfSingle\textwidth}{lZZ}
\hline
Class & \multicolumn{2}{c}{CC}\\
\hhline{~--}
& Count & \%\\
\hline

       Small-Extinct & 0 & 0.00 \\
    Small-Endangered & 4 & 0.19 \\
        Small-Stable & 0 & 0.00 \\
 Small-Institutional & 1 & 3.57 \\
      Mid-Endangered & 4 & 0.87 \\
          Mid-Stable & 2 & 0.12 \\
   Mid-Institutional & 19 & 9.13 \\
    Large-Endangered & 1 & 7.14 \\
        Large-Stable & 4 & 3.01 \\
 Large-Institutional & 100 & 46.08 \\

\hline
\end{tabularx}
\caption{The Coverage of the 12 Ethnologue language classes in the CommonCrawl. The Count column shows the number of languages in the relevant class covered by the CommonCrawl and the \% column shows that number as a percentage of the total number of languages in the class.}
\label{tab:cc12}
%\end{tabular}
\end{table}

%\clearpage

\FloatBarrier
%\null\vfill\null\vfill\null
%\pagebreak

\section{\citet{joshi-etal-2020-state}'s Class Descriptions}
\label{sec:JoshiClass}

\begin{table}[!hbt]
\centering
{\small %
    %\begin{tabular}{p{.04\textwidth}p{.7\textwidth}p{.04\textwidth}p{.2\textwidth}}
\begin{tabularx}{\half\textwidth}{cXrl}
    \hline
    Class & Description & \multicolumn{2}{c}{Language} \\ 
    \hhline{~~--}
 & & Count & Examples \\
    \hline \hline
    0 & Have exceptionally limited resources, and have rarely been considered in language technologies. & 2191 & \makecell{Slovene\\Sinhala}  \\ \hline
    1 & Have some unlabelled data; however, collecting labelled data is challenging. & 222 & \makecell{Nepali\\Telugu} \\ \hline
    2 & A small set of labelled datasets has been collected, and language support communities are there to support the language. & 19 & \makecell{Zulu\\Irish} \\ \hline 
    3 & Has a strong web presence, and a cultural community that backs it. Have been highly benefited by unsupervised pre-training. & 28 & \makecell{Afrikaans\\Urdu} \\ \hline
    4 & Have a large amount of unlabelled data, and lesser, but still a significant amount of labelled data. have  dedicated NLP communities researching these languages. & 18 & \makecell{Russian\\Hindi}\\ \hline
    5 & Have a dominant online presence. There have been massive investments in the development of resources and technologies. & 7 & \makecell{English\\Japanese}\\ \hline
   \end{tabularx}
   %\end{tabular}
   }
  \caption{Language Categories identified by~\citet{joshi-etal-2020-state}}~\label{tab:language_categories}
  \label{joshi_classes}
\end{table}

This is the language categorisation originally proposed by~\citet{joshi-etal-2020-state}. Note that the number of languages reported here are the numbers originally reported by them.  This categorisation is done considering the number of Wikipedia pages and the total of ELRA and LDC datasets per language.

%\FloatBarrier
%\null\vfill\null\vfill\null
%\pagebreak
\section{Analysis of language Coverage in XLM-R and mBERT}
\label{sec:xlmr_coverage}
\begin{figure}[!hbt]
\centering
\centering         \includegraphics[width=\halfSingle\textwidth]{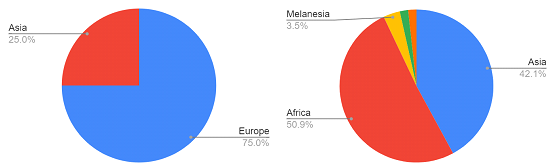}
         \caption{(a) Where the non-Large-Institutional languages included in XLM-R and mBERT models reside. (b) Where the Large-Institutional languages NOT included in XLM-R and mBERT reside.}
         \label{fig:xlmr_coverage}
\end{figure}

\section{Wikipedia 12 Class Analysis}
\label{sec:Wikipedia}

We conducted an analysis on the size of Wikipedias in each of the languages that have a Wikipedia in the relevant language. The first of the analysis, shown in Fig~\ref{fig:LangGeo}, shows the distribution of the languages belonging to the 12 Ethnologue language classes by the geographical origin of each of the languages. It is very important to note that, this means languages with colonial histories such as English, French, Spanish, Portuguese are counted for \textit{Western Europe} and not for locations that they have colonised and displaced the local languages. The reason for this is to show the disparity of prevalence of languages on Wikipedia where all things equal and free in the sense that, any person with knowledge in an under represented language or otherwise may go and write articles at no cost. But it seems, that is not happening. Consider specially the case of \textit{North America}, \textit{South America}, \textit{Australia and New Zealand}. When the colonial languages are taken off consideration from those areas and we look at the state of native languages, we see that they are being under utilised.

\begin{figure}[!hbt]
    \centering
    \includegraphics[width=\half\textwidth]{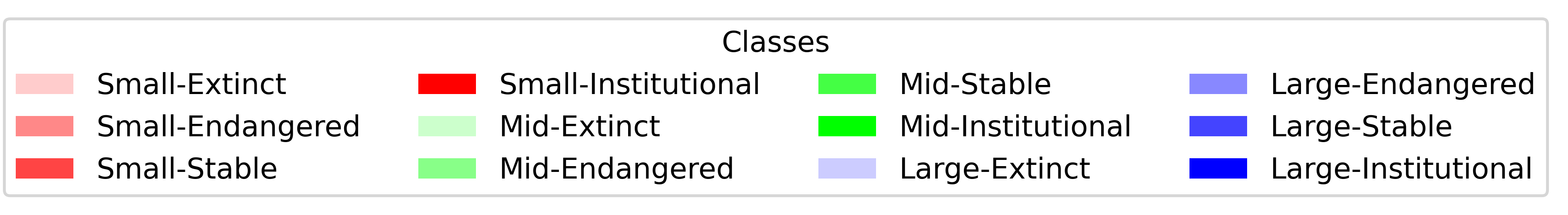}
    \includegraphics[width=\half\textwidth]{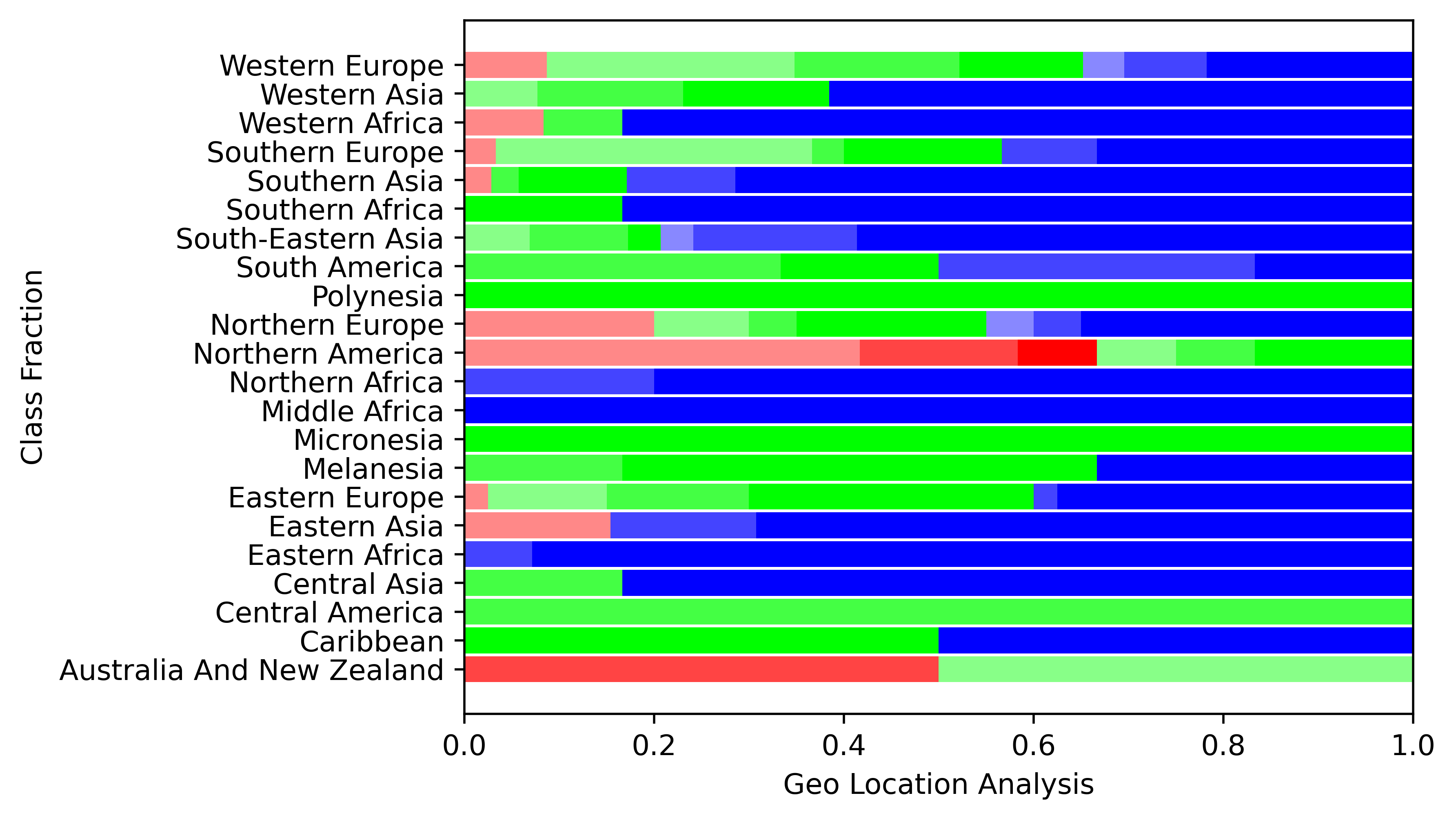}
    \caption{The distribution of languages that have wikis among the 12 Ethnologue Classes - By Geographical Location}
    \label{fig:LangGeo}
\end{figure}

The second analysis, shown in Figure~\ref{fig:LangGeo}, is similar to the first in set up but instead of geographical location, focuses on the language family. Most analysis done for language are commonly dominated by languages in the \textit{Indo-European} family given the wide global spread that family of languages enjoy. In our analysis, we have taken that pressure off the other language families and tried to look at them in an equal footing. By doing this we make a number of interesting observations. The \textit{Afro-Asiatic} group with contains \textit{Arabic} and \textit{Hebrew} seem to enjoy a spread skewed towards \textit{Institutionally} supported languages. The same pattern but with a slightly weaker bias can be observed from the \textit{Dravidian} family of languages native to the southern part of India. We also note that the language families such as \textit{Koreanic} and \textit{Japonic} which carry only the eponymous languages also enjoying complete \textit{Institutional} status.  

\begin{figure}[!hbt]
    \centering
    \includegraphics[width=\half\textwidth]{images/Axis_Dist.png}
    \includegraphics[width=\half\textwidth]{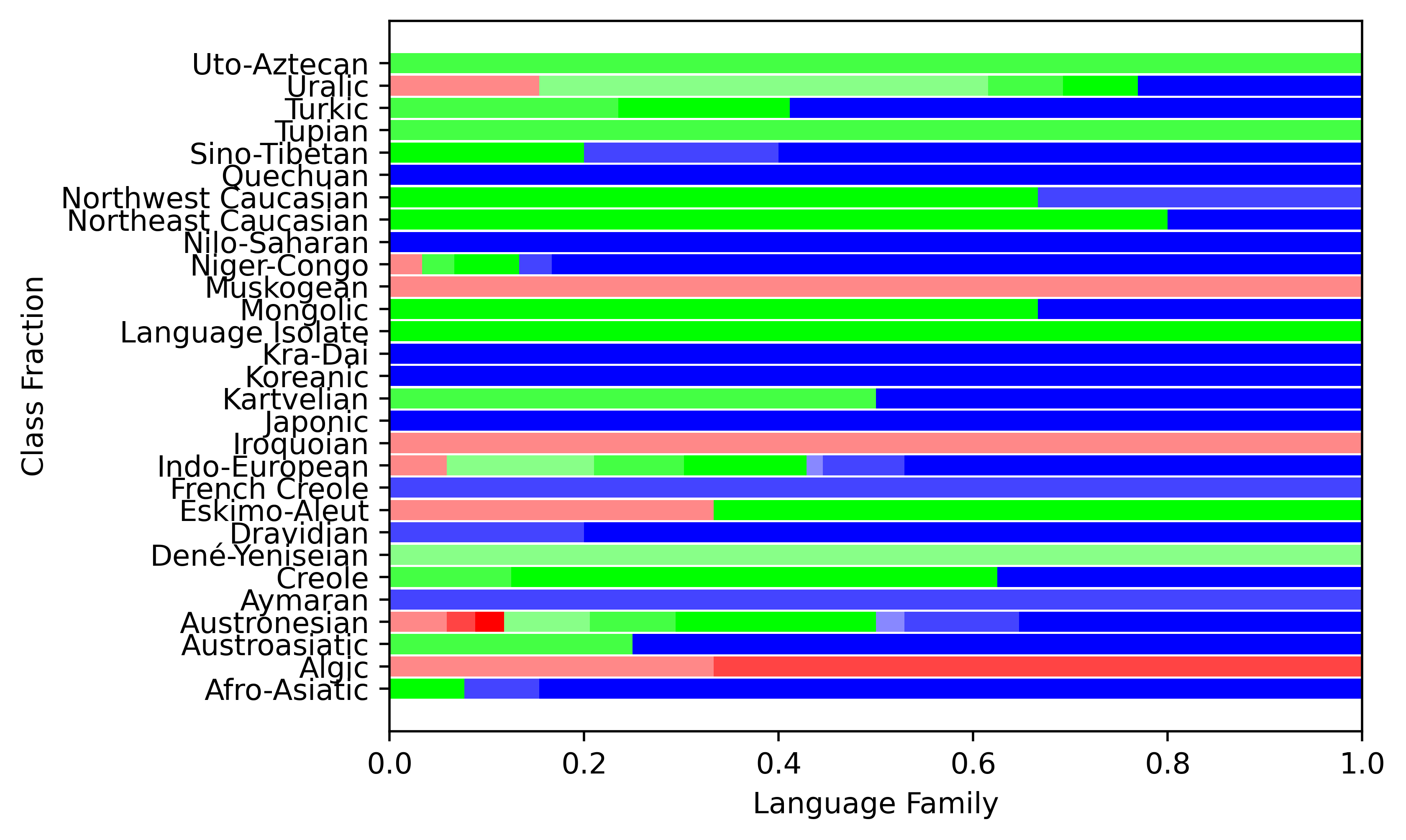}
    \caption{The distribution of languages that have wikis among the 12 Ethnologue Classes - By Language Families}
    \label{fig:LangGeo}
\end{figure}

These observations further re-enforce our earlier claims on the impact of resource distribution and support has on the ability of future research in a given language as Wikipedia is one of the most used language sources for NLP. Therefore, whose language has a seat at the Wikipedia table then partially influences, whose language gets a seat at the NLP research table. If we are to lift some of these languages out of resource and research poverty, starting it with building the relevant Wikipedia is a rational place to start given that it has a low barrier to entry and has an already established ecosystem with editor tools, translator tools, and most importantly collaborative community help.

\iffalse
\begin{figure}[!hbt]
    \centering
    \begin{subfigure}[!hbt]{0.48\textwidth}
         \centering
         \includegraphics[width=\textwidth]{images/Axis_Dist.png}
     \end{subfigure}
    
    \begin{subfigure}[!hbt]{\half\textwidth}
         \centering         \includegraphics[width=\textwidth]{images/geo_location_analysis_Dist.png}
         \caption{Geographical Location}
        \label{fig:LangGeo}
    \end{subfigure}
    
    \begin{subfigure}[!hbt]{\half\textwidth}
         \centering         \includegraphics[width=\textwidth]{images/language_family_Dist.png}
         \caption{Language Families}
        \label{fig:LangFam}
    \end{subfigure}
    \caption{The distribution of languages that have wikis among the 12 Ethnologue Classes}
\end{figure}
\fi

\FloatBarrier

\section{Impact of Population on the Wikipedia Article Count}
\label{sec:PopvsWiki}
\begin{figure}[!hbt]
     \centering
    \includegraphics[width=\half\textwidth]{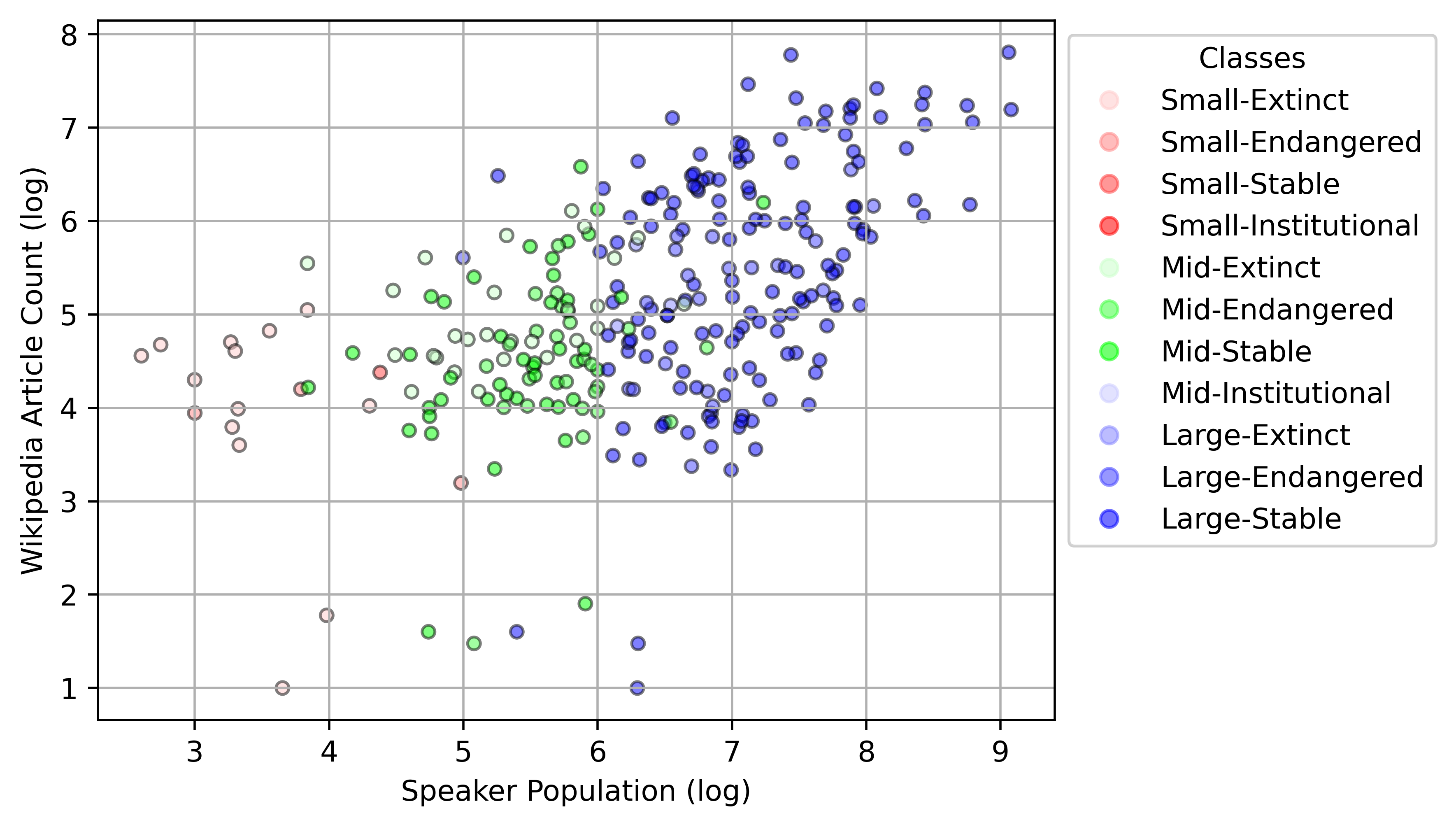}
    \caption{Speaker Population (log) vs Wikipedia Article Count (log).}
    \label{fig:PopvsWiki}
\end{figure}

%Similar to \cite{blasi-etal-2022-systematic}, we identified the GDP of a language by considering the proportionate GDP value from all the countries where the language is spoken. Then this was normalized by the speaker population.
%As can be seen, the number of Wikipedia articles available has a \textit{moderate correlation} ($\input{tables/CorrGDP}$) to the language GDP.
We plotted the graph shown in Figure~\ref{fig:PopvsWiki} and used Pearson correlation. As can be seen, the number of Wikipedia articles available has a \textit{moderate correlation} ($0.518789$) to the population that speaks the language. The coordinates are derived from the L1 and L2 speaker population reported in Wikipedia and the colour of each data point is taken according to the class in Ethnologue. Therefore, data points that violate the colour boundaries along the X-axis are instances where Wikipedia and Ethnologue do not agree. When a language is spoken as L1 in more than one geographical area, the numbers reported in Wikipadia are summed. 

%It is observed that even though the \textit{Language GDP in Billions of Dollars (log)} vs \textit{Wikipedia Article Count (log)} presented in Figure~\ref{fig:GDPvsWiki} also report as a \textit{moderate correlation}, the GDP vs  Wikipedia Article Count reported a Pearson correlation of $\input{tables/CorrGDP}$ as opposed to $\input{tables/Corr}$ reported between \textit{Speaker Population (log)} and \textit{Wikipedia Article Count (log)}. Therefore, it can be claimed that the Wikipedia Article Count is more dependant on the Language GDP rather than the Speaker Population.

\section{HuggingFace Datasets Task and Language Analysis}
\label{huggingface_tasks}

\begin{table*}[!hbt]
\tiny
\begin{tabularx}{\textwidth}{|l|ZZZZZZZZ|}
\hline
Task & Large-Institutional & Large-Stable & Large-Endangered & Mid-Institutional & Mid-Stable & Mid-Endangered & Small-Stable & Small-Endangered \\
\hline
translation & 1579 & 12 & 1 & 123 & 17 & 39 & 2 & 20 \\
text-classification & 896 & 6 & 0 & 35 & 6 & 14 & 0 & 10 \\
text-generation & 687 & 6 & 0 & 52 & 12 & 18 & 1 & 8 \\
fill-mask & 597 & 6 & 0 & 50 & 12 & 18 & 1 & 8 \\
token-classification & 469 & 5 & 0 & 24 & 5 & 9 & 0 & 6 \\
question-answering & 487 & 3 & 0 & 5 & 0 & 0 & 0 & 1 \\
conditional-text-generation & 387 & 3 & 0 & 32 & 6 & 7 & 0 & 3 \\
text-retrieval & 179 & 3 & 0 & 7 & 0 & 5 & 0 & 2 \\
text2text-generation & 183 & 0 & 0 & 2 & 0 & 1 & 0 & 2 \\
other & 137 & 2 & 0 & 7 & 3 & 2 & 0 & 1 \\
image-to-text & 125 & 2 & 0 & 6 & 0 & 5 & 0 & 2 \\
summarization & 118 & 0 & 0 & 0 & 0 & 0 & 0 & 0 \\
automatic-speech-recognition & 101 & 0 & 0 & 7 & 1 & 1 & 0 & 0 \\
multiple-choice & 104 & 1 & 0 & 1 & 0 & 0 & 0 & 0 \\
speech-processing & 74 & 0 & 0 & 6 & 3 & 2 & 0 & 0 \\
zero-shot-classification & 59 & 0 & 0 & 0 & 0 & 0 & 0 & 0 \\
table-question-answering & 58 & 0 & 0 & 0 & 0 & 0 & 0 & 0 \\
tabular-classification & 57 & 0 & 0 & 0 & 0 & 0 & 0 & 0 \\
audio-classification & 45 & 0 & 0 & 4 & 0 & 1 & 0 & 0 \\
sequence-modeling & 36 & 0 & 0 & 2 & 0 & 0 & 0 & 0 \\
structure-prediction & 35 & 0 & 0 & 0 & 0 & 0 & 0 & 0 \\
image-classification & 25 & 0 & 0 & 0 & 0 & 0 & 0 & 0 \\
conversational & 16 & 0 & 0 & 0 & 0 & 0 & 0 & 0 \\
sentence-similarity & 13 & 0 & 0 & 0 & 0 & 0 & 0 & 0 \\
tabular-to-text & 12 & 0 & 0 & 0 & 0 & 0 & 0 & 0 \\
table-to-text & 10 & 0 & 0 & 0 & 0 & 0 & 0 & 0 \\
paraphrase-mining & 8 & 0 & 0 & 0 & 0 & 0 & 0 & 0 \\
object-detection & 7 & 0 & 0 & 0 & 0 & 0 & 0 & 0 \\
text-scoring & 7 & 0 & 0 & 0 & 0 & 0 & 0 & 0 \\
commonsense reasoning & 4 & 0 & 0 & 0 & 0 & 0 & 0 & 0 \\
coreference resolution & 4 & 0 & 0 & 0 & 0 & 0 & 0 & 0 \\
sentiment-analysis & 4 & 0 & 0 & 0 & 0 & 0 & 0 & 0 \\
question-generation & 4 & 0 & 0 & 0 & 0 & 0 & 0 & 0 \\
image-to-image & 3 & 0 & 0 & 0 & 0 & 0 & 0 & 0 \\
text-to-image & 3 & 0 & 0 & 0 & 0 & 0 & 0 & 0 \\
email subject & 3 & 0 & 0 & 0 & 0 & 0 & 0 & 0 \\
one liner summary & 3 & 0 & 0 & 0 & 0 & 0 & 0 & 0 \\
topic modeling & 3 & 0 & 0 & 0 & 0 & 0 & 0 & 0 \\
symbolic-regression & 3 & 0 & 0 & 0 & 0 & 0 & 0 & 0 \\
text\_classification & 3 & 0 & 0 & 0 & 0 & 0 & 0 & 0 \\
meeting title & 3 & 0 & 0 & 0 & 0 & 0 & 0 & 0 \\
visual-question-answering & 3 & 0 & 0 & 0 & 0 & 0 & 0 & 0 \\
machine-translation & 3 & 0 & 0 & 0 & 0 & 0 & 0 & 0 \\
text-mining & 3 & 0 & 0 & 0 & 0 & 0 & 0 & 0 \\
image-segmentation & 3 & 0 & 0 & 0 & 0 & 0 & 0 & 0 \\
classification & 3 & 0 & 0 & 0 & 0 & 0 & 0 & 0 \\
masked-auto-encoding & 2 & 0 & 0 & 0 & 0 & 0 & 0 & 0 \\
closed-domain-abstrative-qa & 2 & 0 & 0 & 0 & 0 & 0 & 0 & 0 \\
dialog-response-generation & 2 & 0 & 0 & 0 & 0 & 0 & 0 & 0 \\
extractive-qa & 2 & 0 & 0 & 0 & 0 & 0 & 0 & 0 \\
neural-machine-translation & 2 & 0 & 0 & 0 & 0 & 0 & 0 & 0 \\
rendered-language-modelling & 2 & 0 & 0 & 0 & 0 & 0 & 0 & 0 \\
abstractive-qa & 2 & 0 & 0 & 0 & 0 & 0 & 0 & 0 \\
language-modelling & 2 & 0 & 0 & 0 & 0 & 0 & 0 & 0 \\
long-texts & 2 & 0 & 0 & 0 & 0 & 0 & 0 & 0 \\
other-test & 1 & 0 & 0 & 1 & 0 & 0 & 0 & 0 \\
feature-extraction & 2 & 0 & 0 & 0 & 0 & 0 & 0 & 0 \\
other-text-to-structured & 2 & 0 & 0 & 0 & 0 & 0 & 0 & 0 \\
text-understanding & 1 & 0 & 0 & 0 & 0 & 0 & 0 & 0 \\
commonsense-reasoning & 1 & 0 & 0 & 0 & 0 & 0 & 0 & 0 \\
moral-reasoning & 1 & 0 & 0 & 0 & 0 & 0 & 0 & 0 \\
social-reasoning & 1 & 0 & 0 & 0 & 0 & 0 & 0 & 0 \\
style-transfer & 1 & 0 & 0 & 0 & 0 & 0 & 0 & 0 \\
task-dialogue & 1 & 0 & 0 & 0 & 0 & 0 & 0 & 0 \\
natural-language-understanding & 1 & 0 & 0 & 0 & 0 & 0 & 0 & 0 \\
text-comprehension & 1 & 0 & 0 & 0 & 0 & 0 & 0 & 0 \\
story-generation & 1 & 0 & 0 & 0 & 0 & 0 & 0 & 0 \\
natural-language-generation & 1 & 0 & 0 & 0 & 0 & 0 & 0 & 0 \\
data-to-text & 1 & 0 & 0 & 0 & 0 & 0 & 0 & 0 \\
MultiLabel Text Classification & 1 & 0 & 0 & 0 & 0 & 0 & 0 & 0 \\
commonsense-generation & 1 & 0 & 0 & 0 & 0 & 0 & 0 & 0 \\
sequence-modelling & 1 & 0 & 0 & 0 & 0 & 0 & 0 & 0 \\
open-dialogue & 1 & 0 & 0 & 0 & 0 & 0 & 0 & 0 \\
patents & 1 & 0 & 0 & 0 & 0 & 0 & 0 & 0 \\
deduplication & 1 & 0 & 0 & 0 & 0 & 0 & 0 & 0 \\
Information Retrieval & 1 & 0 & 0 & 0 & 0 & 0 & 0 & 0 \\
named-entity-recognition & 1 & 0 & 0 & 0 & 0 & 0 & 0 & 0 \\
simplification & 1 & 0 & 0 & 0 & 0 & 0 & 0 & 0 \\
video-captionning & 1 & 0 & 0 & 0 & 0 & 0 & 0 & 0 \\
text-generation-other-common-sense-inference & 1 & 0 & 0 & 0 & 0 & 0 & 0 & 0 \\
text-generation-other-discourse-analysis & 1 & 0 & 0 & 0 & 0 & 0 & 0 & 0 \\
other-text-to-tabular & 1 & 0 & 0 & 0 & 0 & 0 & 0 & 0 \\
other-text-search & 1 & 0 & 0 & 0 & 0 & 0 & 0 & 0 \\
question-pairing & 1 & 0 & 0 & 0 & 0 & 0 & 0 & 0 \\
Semantic Search & 1 & 0 & 0 & 0 & 0 & 0 & 0 & 0 \\
question\_answering & 1 & 0 & 0 & 0 & 0 & 0 & 0 & 0 \\
Evaluation of language models & 1 & 0 & 0 & 0 & 0 & 0 & 0 & 0 \\
masked-language-modeling & 1 & 0 & 0 & 0 & 0 & 0 & 0 & 0 \\
multi-class classification & 1 & 0 & 0 & 0 & 0 & 0 & 0 & 0 \\
topic-classification & 1 & 0 & 0 & 0 & 0 & 0 & 0 & 0 \\
paraphrase & 1 & 0 & 0 & 0 & 0 & 0 & 0 & 0 \\
language-modeling & 1 & 0 & 0 & 0 & 0 & 0 & 0 & 0 \\
machine translation & 1 & 0 & 0 & 0 & 0 & 0 & 0 & 0 \\
text-to-speech & 1 & 0 & 0 & 0 & 0 & 0 & 0 & 0 \\
image-generation & 1 & 0 & 0 & 0 & 0 & 0 & 0 & 0 \\
\hline
\end{tabularx}
\caption{Datasets for different task-language category combinations (Excluding the 50 tasks that are not tagged with any language).}
\label{tab:HugginTask}
\end{table*}

In Table~\ref{tab:HugginTask} we show the datasets that are tagged with  languages and tasks on HuggingFace classified to the Ethnologue language classes. From the get go, it is evident that all the languages are not represented. 
We observe that only 8 Ethnologue classes: \textit{Large-Institutional},\textit{Large-Stable}, \textit{Large-Endangered} \textit{Mid-Institutional}, \textit{Mid-Stable}, \textit{Mid-Endangered}, \textit{Small-Stable},\textit{Small-Endangered} have any data sets tagged with their member languages. 

Even if we disregard \textit{Large-Extinct} and \textit{Mid-Extinct} which are missing in all other analyses, this still comes short for \textit{Small-Institutional} and and \textit{Small-Extinct}. 
On the other end, we note that the following 50 tasks has zero languages tagged on their data sets: \textit{information-retrieval}, \textit{zero-shot-retrieval}, \textit{zero-shot-information-retrieval}, \textit{time-series-forecasting}, \textit{computer-vision}, \textit{reasoning}, \textit{paraphrasing}, \textit{code-generation}, \textit{tts}, \textit{image}, \textit{image-retrieval}, \textit{image-captioning}, \textit{text-generation-other-code-modeling}, \textit{Code Generation}, \textit{Translation}, \textit{Text2Text generation}, \textit{text-to-slide}, \textit{paraphrase detection}, \textit{Summarization}, \textit{cross-language-transcription}, \textit{grammatical error correction}, \textit{named-entity-disambiguation}, \textit{textual-entailment}, \textit{natural-language-inference}, \textit{query-paraphrasing}, \textit{text-regression}, \textit{entity-extraction}, \textit{unpaired-image-to-image-translation}, \textit{generative-modelling}, \textit{Token Classification}, \textit{caption-retrieval}, \textit{gpt-3}, \textit{crowdsourced}, \textit{sequence2sequence}, \textit{Inclusive Language}, \textit{Text Neutralization}, \textit{super-resolution}, \textit{image-enhancement}, \textit{speech-synthesis}, \textit{data-integration}, \textit{Language-model}, \textit{Automatic-Speech-Recognition}, 

\FloatBarrier

\textit{influence-attribution}, \textit{question-answering-retrieval}, \textit{text}, \textit{linear-regression}, \textit{syntactic-evaluation}, \textit{text classification}, \textit{text tagging}, \textit{named entity recognition}.

Now this does not imply that all of these are not text based tasks. Some of them, (e.g., \textit{image}) may fall into that category. But some, (e.g., \textit{Text Neutralization}, \textit{Text2Text generation}) are ostensibly text based tasks. So is \textit{Translation} which a variant capitalisation of \textit{translation} which is the highest language tagged task.
What we can say here, given how HuggingFace search gives the intersection of the labels, is that, this must be an artefact of how users tag their data sets on HuggingFace. It seems some users tag their task, but have not taken steps to tag the languages in their data set.

Therefore, it is vital that before using the HuggingFace tags for any meta-analysis on the NLP domain datasets, a large-scale data-clean up task be done on them. While the task still seem to be manually tractable, with the speed of growth shown by HuggingFace datasets, it is conceivable that it would soon cease to be so. Alternatively, it can be suggested to introduce a levelled tag system to HuggingFace where the top level tag is selected from a pre-set collection of tags set by HuggingFace while the lower level tag can be typed-in by the person who upload the data set. %This may make future meta-analysis of language datasets on HuggingFace more efficient.

\section{OPUS Data}
\label{sec:OPUS}
We extracted the number of sentences available for each language listed in OPUS as shown in Table~\ref{tab:OPUS}.

\begin{table}[!htb]
    \centering
    \begin{tabularx}{0.5\textwidth}{|X|Z|}
\hline
Language Class &     Data Set Count \\
\hline
Large-Institutional &  1.556114e+10 \\
Large-Stable        &  3.216824e+07 \\
Mid-Institutional   &  6.123440e+07 \\
Mid-Stable          &  4.243600e+04 \\
Mid-Endangered      &  7.833096e+06 \\
Small-Institutional &  1.104000e+03 \\
Small-Stable        &  1.200500e+04 \\
Small-Endangered    &  1.278468e+06 \\
Small-Extinct       &  8.000000e+00 \\
\hline
        \end{tabularx}
\caption{OPUS Data Set Counts}
\label{tab:OPUS}
\end{table}

%\onecolumn

\section{The Distribution of Resources}
\label{sec:reso}

\begin{figure*}[!hbt]
    \centering
    \includegraphics[width=\textwidth]{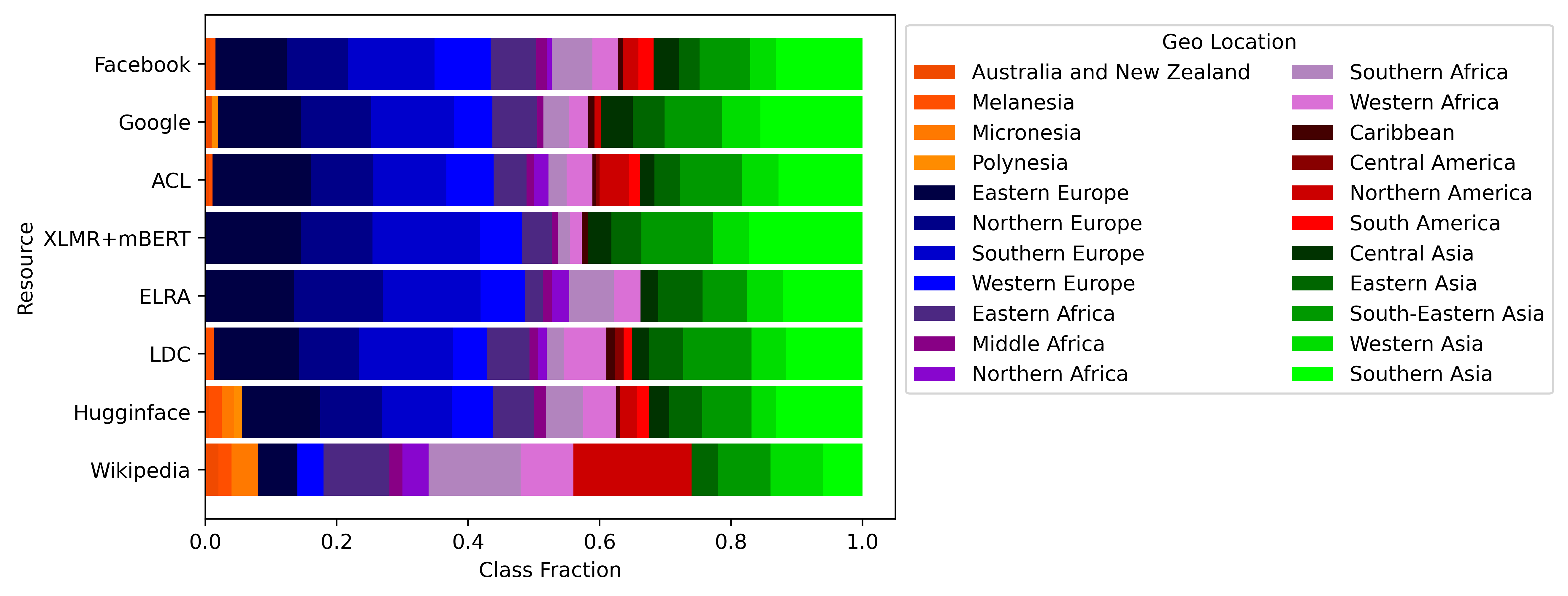}
    \caption{By Geographical Location of the Language Origin}
    \label{fig:LangGeoWMAppen}
\end{figure*}

\begin{figure*}[!hbt]
    \centering
    \includegraphics[width=\textwidth]{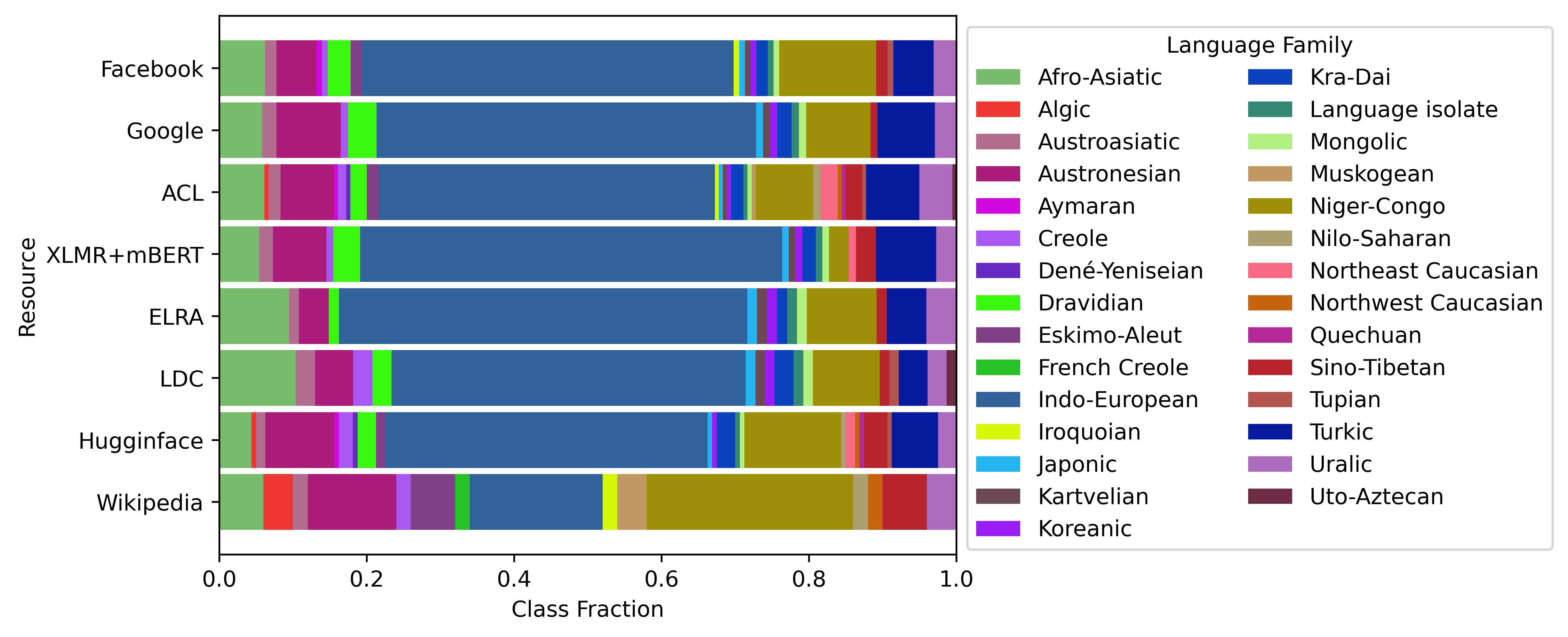}
    \caption{By Language Families}
    \label{fig:LangFamWMAppn}
\end{figure*}

We have added larger versions of Fig~\ref{fig:LangGeoWM} and Fig~\ref{fig:LangFamWM} at Fig~\ref{fig:LangGeoWMAppen} and Fig~\ref{fig:LangFamWMAppn} respectively.   

\section{Impact of using Huggingface as a Data Source}
\label{sec:HuggingfaceImpact}
When \textit{Huggingface} data sets were introduced, 87 languages changed their class. Out of this, 84 were promotions. The three demotions are Afrikaans, Bosnian, and Croatian. The full list of class changes are given below. The list header gives the \textit{Ethnologue} language class followed by the \citet{joshi-etal-2020-state} class shift in parenthesis. The cases where language classes are demoted are indicated by an ``*'' at the end of the list header.

\begin{table*}[!hbt]
\centering
\renewcommand*{\arraystretch}{\tabComp}
\small
\begin{tabularx}{\textwidth}{|Y|YYYY|YYYY|YYYY|Y|}
\hline
Joshi & \multicolumn{4}{c|}{Small} & \multicolumn{4}{c|}{Mid} & \multicolumn{4}{c|}{Large}& \multirow{2}{*}{Total}\\
\hhline{~------------}
& Ex & En & St & In & Ex & En & St & In & Ex & En & St & In & \\
\hline
\hline

0 & 331 & 2146 & 1165 & 27 & 0 & 430 & 1676 & 164 & 0 & 11 & 109 & 75 & 6134 \\
1 & 1 & 15 & 3 & 1 & 0 & 28 & 24 & 41 & 0 & 2 & 22 & 73 & 210 \\
2 & 0 & 0 & 0 & 0 & 0 & 0 & 0 & 2 & 0 & 1 & 0 & 19 & 22 \\
3 & 0 & 1 & 0 & 0 & 0 & 0 & 0 & 0 & 0 & 0 & 2 & 26 & 29 \\
4 & 0 & 0 & 0 & 0 & 0 & 0 & 0 & 1 & 0 & 0 & 0 & 17 & 18 \\
5 & 0 & 0 & 0 & 0 & 0 & 0 & 0 & 0 & 0 & 0 & 0 & 7 & 7 \\
\hline
Total & 332 & 2162 & 1168 & 28 & 0 & 458 & 1700 & 208 & 0 & 14 & 133 & 217 & 6420 \\

\hline
\end{tabularx}
\caption{Confusion Matrix of \citet{joshi-etal-2020-state} classes and Ethnologue language classes considering only \textit{LDC} and \textit{ELRA} as the annotated sources, where Ex=\textit{Extinct}, En=\textit{Endangered}, St=\textit{Stable}, and In=\textit{Institutional}.}
\label{tab:confuseWitouthHugging}
\end{table*}

\begin{table*}[!hbt]
\centering
\renewcommand*{\arraystretch}{\tabComp}
\small
\begin{tabularx}{\textwidth}{|Y|YYYY|YYYY|YYYY|Y|}
\hline
Joshi & \multicolumn{4}{c|}{Small} & \multicolumn{4}{c|}{Mid} & \multicolumn{4}{c|}{Large}& \multirow{2}{*}{Total}\\
\hhline{~------------}
& Ex & En & St & In & Ex & En & St & In & Ex & En & St & In & \\
\hline
\hline

0 & 331 & 2146 & 1165 & 27 & 0 & 430 & 1676 & 164 & 0 & 11 & 109 & 75 & 6134 \\
1 & 1 & 12 & 3 & 1 & 0 & 19 & 23 & 24 & 0 & 2 & 18 & 27 & 130 \\
2 & 0 & 3 & 0 & 0 & 0 & 9 & 1 & 18 & 0 & 1 & 4 & 61 & 97 \\
3 & 0 & 1 & 0 & 0 & 0 & 0 & 0 & 1 & 0 & 0 & 2 & 26 & 30 \\
4 & 0 & 0 & 0 & 0 & 0 & 0 & 0 & 1 & 0 & 0 & 0 & 21 & 22 \\
5 & 0 & 0 & 0 & 0 & 0 & 0 & 0 & 0 & 0 & 0 & 0 & 7 & 7 \\
\hline
Total & 332 & 2162 & 1168 & 28 & 0 & 458 & 1700 & 208 & 0 & 14 & 133 & 217 & 6420 \\

\hline
\end{tabularx}
\caption{Confusion Matrix of \citet{joshi-etal-2020-state} classes and Ethnologue language classes considering \textit{Huggingface}, \textit{LDC}, and \textit{ELRA} as the annotated sources, where Ex=\textit{Extinct}, En=\textit{Endangered}, St=\textit{Stable},and In=\textit{Institutional}.}
\label{tab:confuseWithHugging}
\end{table*}

%\FloatBarrier

\begin{itemize}
    \item \textbf{\textit{Large-Institutional} (1 → 2):} 
    Akan, 
    Albanian,
Assamese,
Bamanankan,
Bikol,
Burmese,
Chichewa,
Chuvash,
Fulah,
Ganda,
Gujarati,
Igbo,
Javanese,
Kannada,
Kashmiri,
Kinyarwanda,
kurdish (kurmanji),
Kyrgyz,
Limburgish,
Lingala,
Maithili,
Malagasy,
Malayalam,
Nepali,
Quechua,
Rundi,
Sango,
Shan,
Shona,
Sindhi,
Sinhala,
Somali,
Southern Sotho,
Swati,
Tajik,
Telugu,
Tibetan,
Tsonga,
Turkmen, and
Venda.
\item \textbf{\textit{Large-Stable} (1 → 2):}
Aymara,
Scots,
Sicilian, and
Sunda.
\item \textbf{\textit{Mid-Institutional} (1 → 2):}
Abkhaz,
Avar,
Bislama,
Chamorro,
Dzongkha,
Faroese,
Fijian,
Inuktitut,
Luxembourgish,
Ossetic,
Romansh,
Samoan,
Scottish Gaelic,
Tahitian,
Yakut, and
Yiddish.
\item \textbf{\textit{Mid-Stable} (1 → 2):}
GuaranÃ­.
\item \textbf{\textit{Mid-Endangered} (1 → 2):}
Aragonese,
Breton,
Corsican,
Maori,
Navajo,
Occitan,
Sardinian,
Udmurt, and
Walloon.
\item \textbf{\textit{Small-Endangered} (1 → 2):}
Cornish,
Manx, and
Pali.
\item \textbf{\textit{Large-Institutional} (1 → 3):}
Armenian,
Chechen,
Esperanto,
Macedonian, and
Tatar.
\item \textbf{\textit{Mid-Institutional} (1 → 3):}
Welsh.
\item \textbf{\textit{Large-Institutional} (1 → 4):}
Azerbaijani.
\item \textbf{\textit{Large-Institutional} (3 → 2)*:}
Afrikaans and
Bosnian. 
\item \textbf{\textit{Large-Institutional} (3 → 4):}
Indonesian,
Norwegian,
Romanian and
Ukrainian.
\item \textbf{\textit{Large-Institutional} (4 → 3)*:}
Croatian.
\end{itemize}

%\twocolumn
%\FloatBarrier 

% add paragraph on  and 
We show the confusion Matrix of \citet{joshi-etal-2020-state} classes and the 12 Ethnologue language classes resluting when the \citet{joshi-etal-2020-state} classes are derived only considering \textit{LDC} and \textit{ELRA} as the annotated sources in Table~\ref{tab:confuseWitouthHugging}.

Then we show the same confusion Matrix but considering \textit{Huggingface} in addition to \textit{LDC} and \textit{ELRA} as the annotated sources in Table~\ref{tab:confuseWithHugging}. The information in Table~\ref{tab:confuseWitouthHugging} corresponds to Fig~\ref{fig:joshiWithoutHugging} while the information in Table~\ref{tab:confuseWithHugging} corresponds to Fig~\ref{fig:joshiWithHugging}. We can clearly see some of the promotions and demotions that we discussed above. One very easy to spot transition is the promotion of the three \textit{Small-Endangered} languages: \textit{Cornish}, \textit{Manx}, and \textit{Pali} from class 1 to class 2. Note how in the \textit{Small-Endangered} column of Table~\ref{tab:confuseWitouthHugging}, there are 15 languages in class 1 and 0 languages in class 2. Then in the \textit{Small-Endangered} column of Table~\ref{tab:confuseWithHugging}, there are 12 languages in class 1 and 3 languages in class 2 attesting to the promotion of the aforementioned languages. %

%In Table~\ref{tab:HuggingShift} we show the number of datasets available in \textit{Huggingface} for the 12 Ethnologue language classes  in November 2021 compared to the number of datasets available in \textit{Huggingface} for the 12 Ethnologue language classes in July 2022. The \textit{Difference} column shows the growth in number and each of the normalised columns carries the value obtained by dividing the values in adjoining \textit{count} column by the the number in the \textit{count} column for the relevant class. It can be observed that even after normalising to the class size, the \textit{Large-Institutional} class has the largest representation as well as the steepest growth. This trend of rich getting richer is a cause for concern for those who are interested in developing and using data sets to and from low-resourced languages as this shows that the average interest still lies with the few languages that are already enjoying an abundance of datasets. 

\section{ACL Publication History and Performance}
\label{sec:ACL}
\begin{figure}[!hbt]
     \centering
    \includegraphics[width=\half\textwidth]{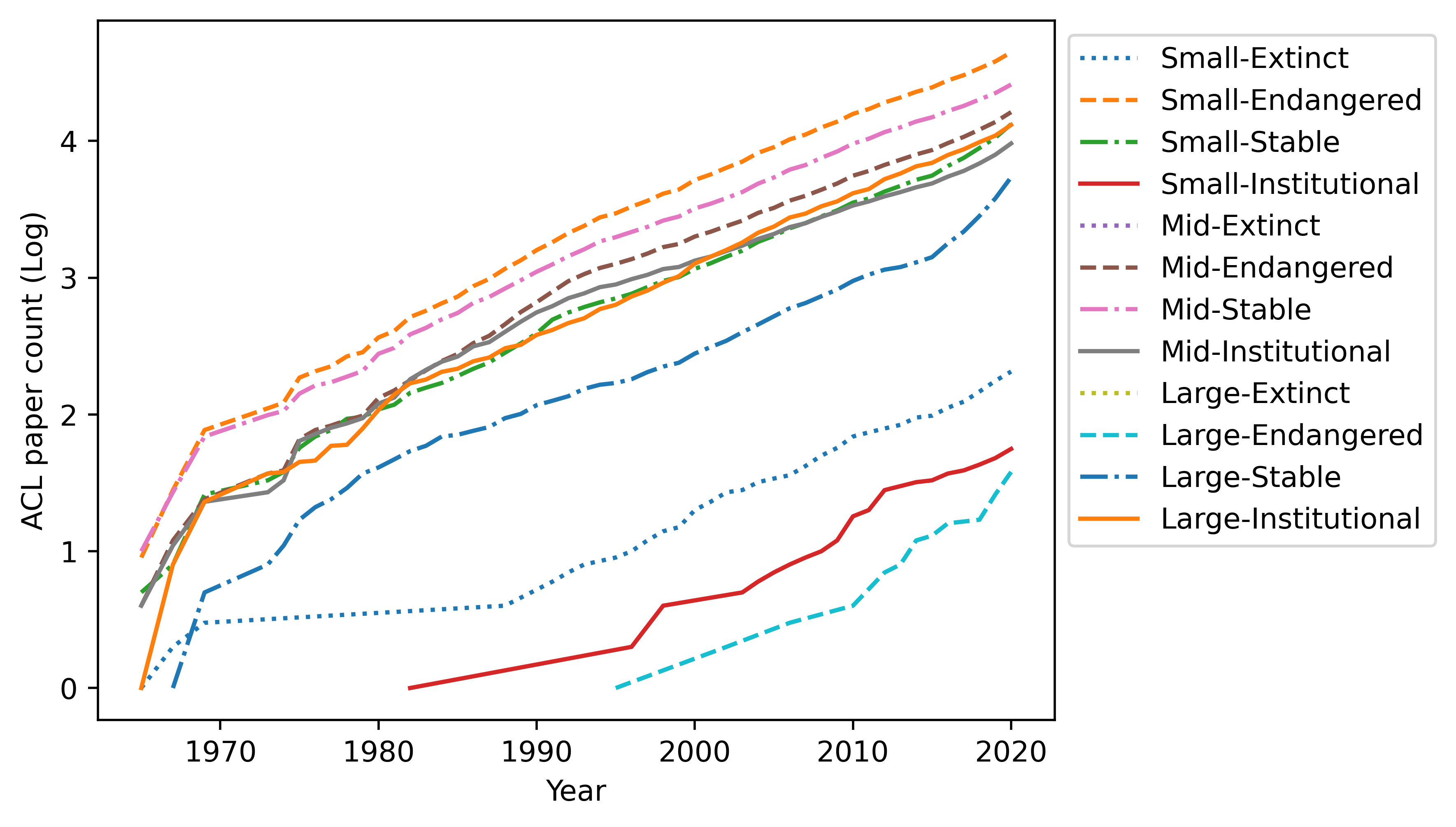}
    \caption{ACL publication count for the 12 Ethnologue language classes (cumulative log)}
    \label{fig:ACLall}
\end{figure}

As shown in Figure\ref{fig:ACLall} (considering all the publications in ACL Anthology), there is a continuous increase of publications for all categories. There are some interesting observations here - (1) research on some language categories started much later than categories such as large-institutional and (2) the number of papers for large-institutional is less than some other categories. We believe this is the impact of workshops. As mentioned by~\citet{bender2019rule}, many research that focused on English did not bother to mention the language in the paper as it is assumed \textit{de facto}.

\begin{figure*}[!hbt]
     \centering   \begin{subfigure}[!hbt]{0.49\textwidth}
         \centering
         \includegraphics[height=15pt,keepaspectratio]{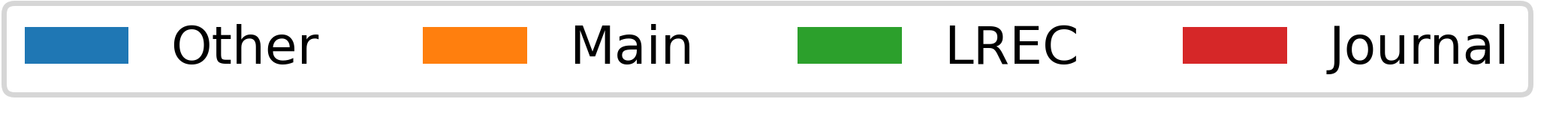}
     \end{subfigure}
     
     \begin{subfigure}[!hbt]{\fourthSingle\textwidth}
         \centering
         \includegraphics[width=\textwidth]{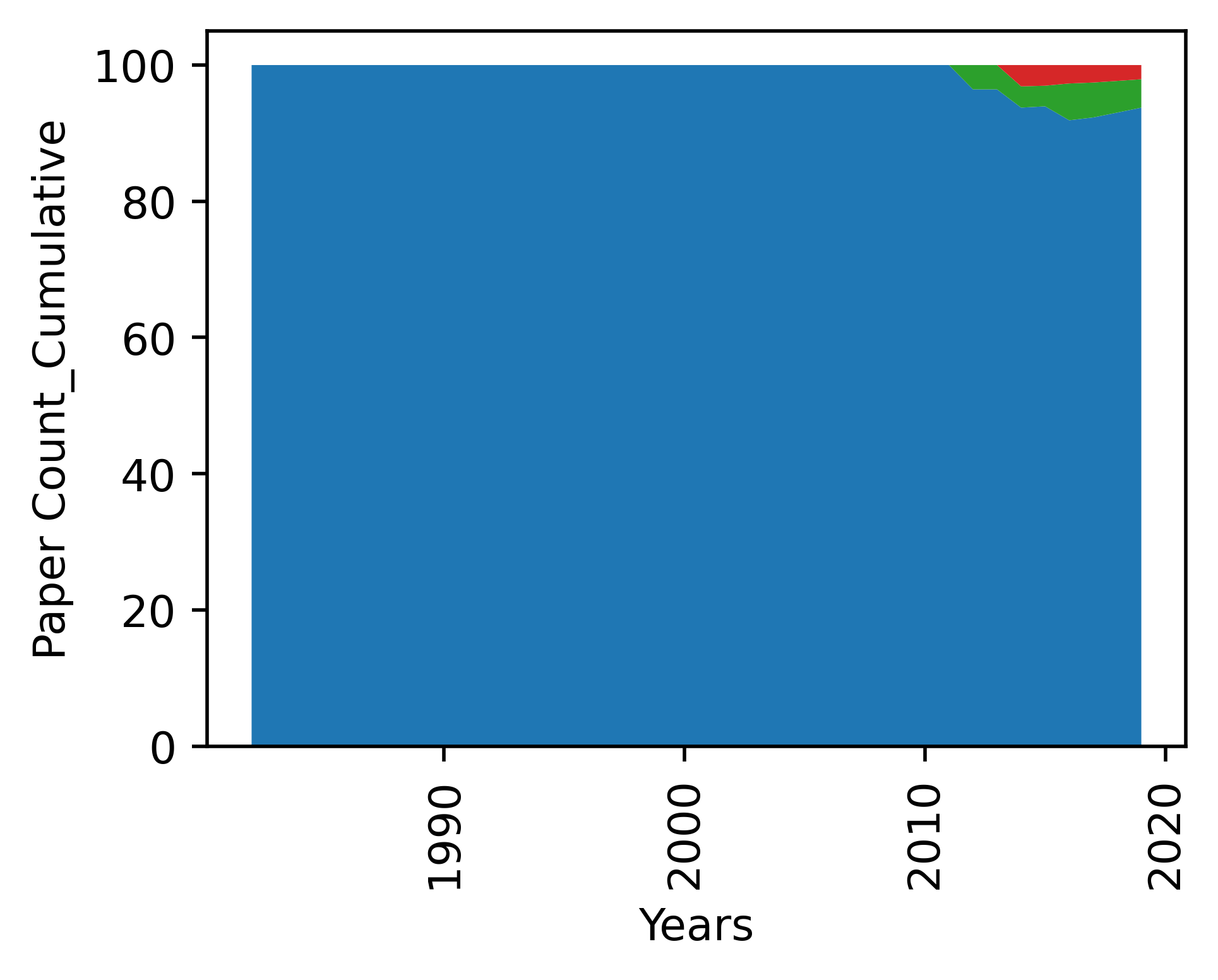}
         \caption{Small-Institutional}
    \end{subfigure}
    \begin{subfigure}[!hbt]{\fourthSingle\textwidth}
         \centering
         \includegraphics[width=\textwidth]{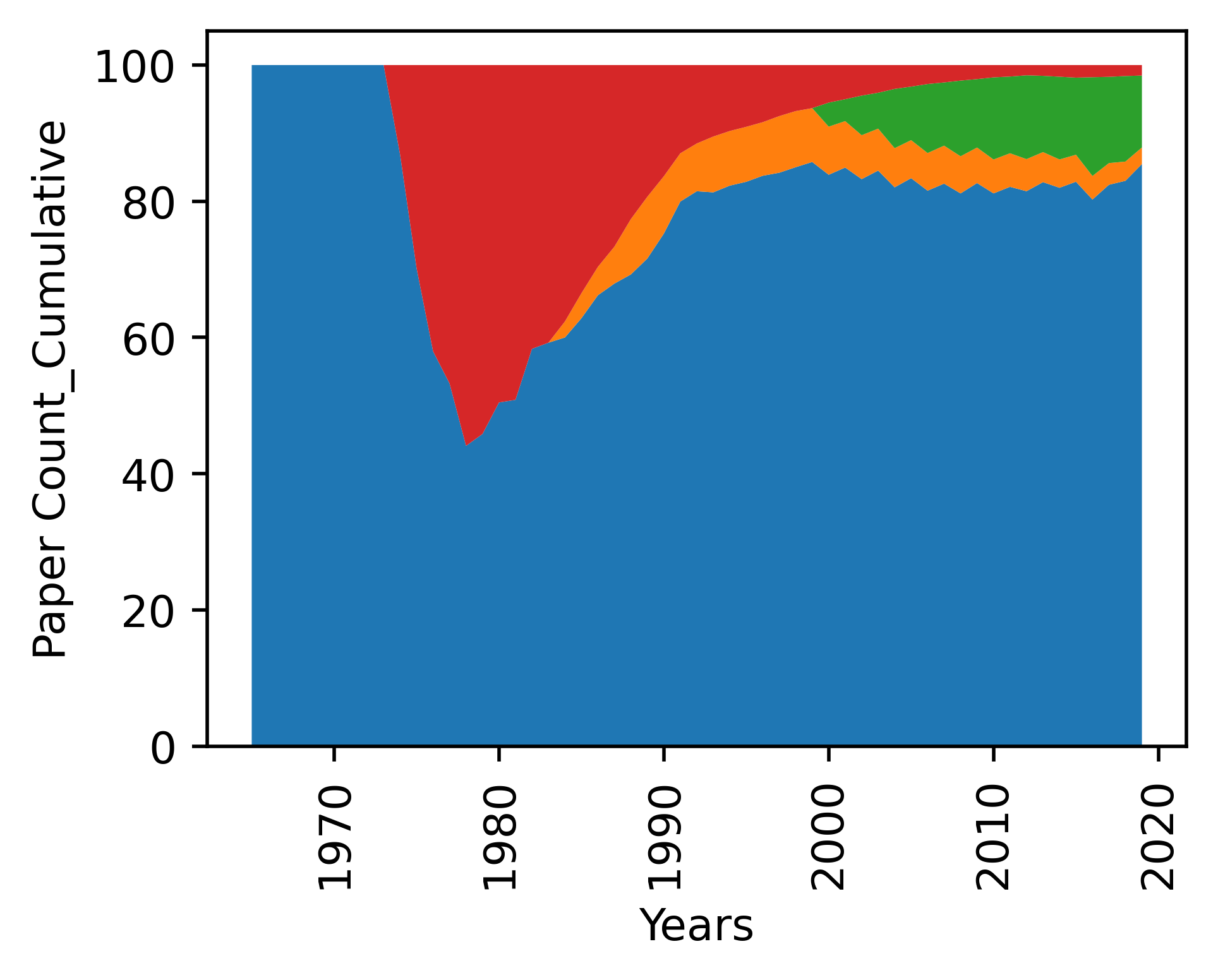}
         \caption{Small-Stable}
    \end{subfigure}
    \begin{subfigure}[!hbt]{\fourthSingle\textwidth}
         \centering         \includegraphics[width=\textwidth]{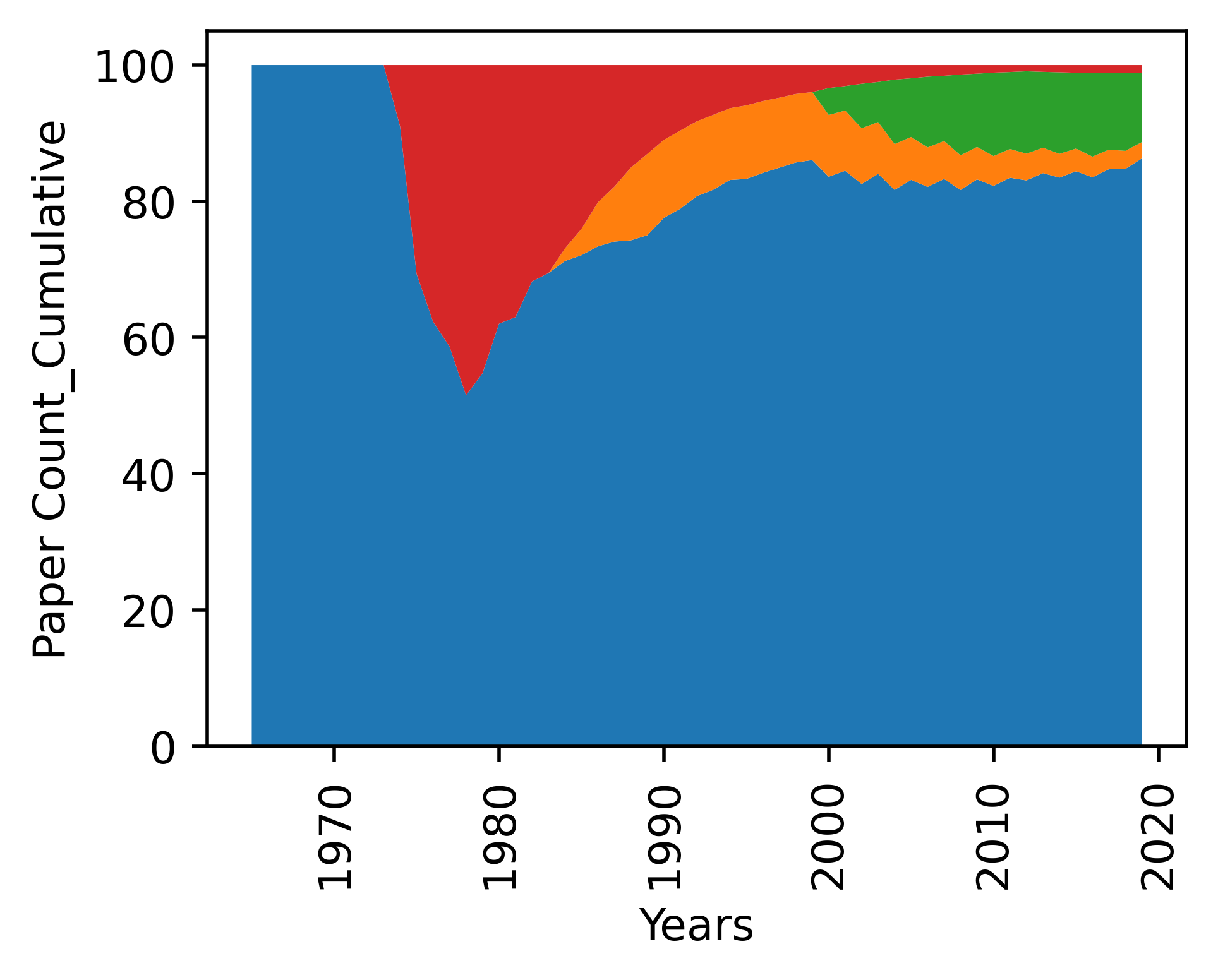}
         \caption{Small-Endangered}
    \end{subfigure}
    \begin{subfigure}[!hbt]{\fourthSingle\textwidth}
         \centering
         \includegraphics[width=\textwidth]{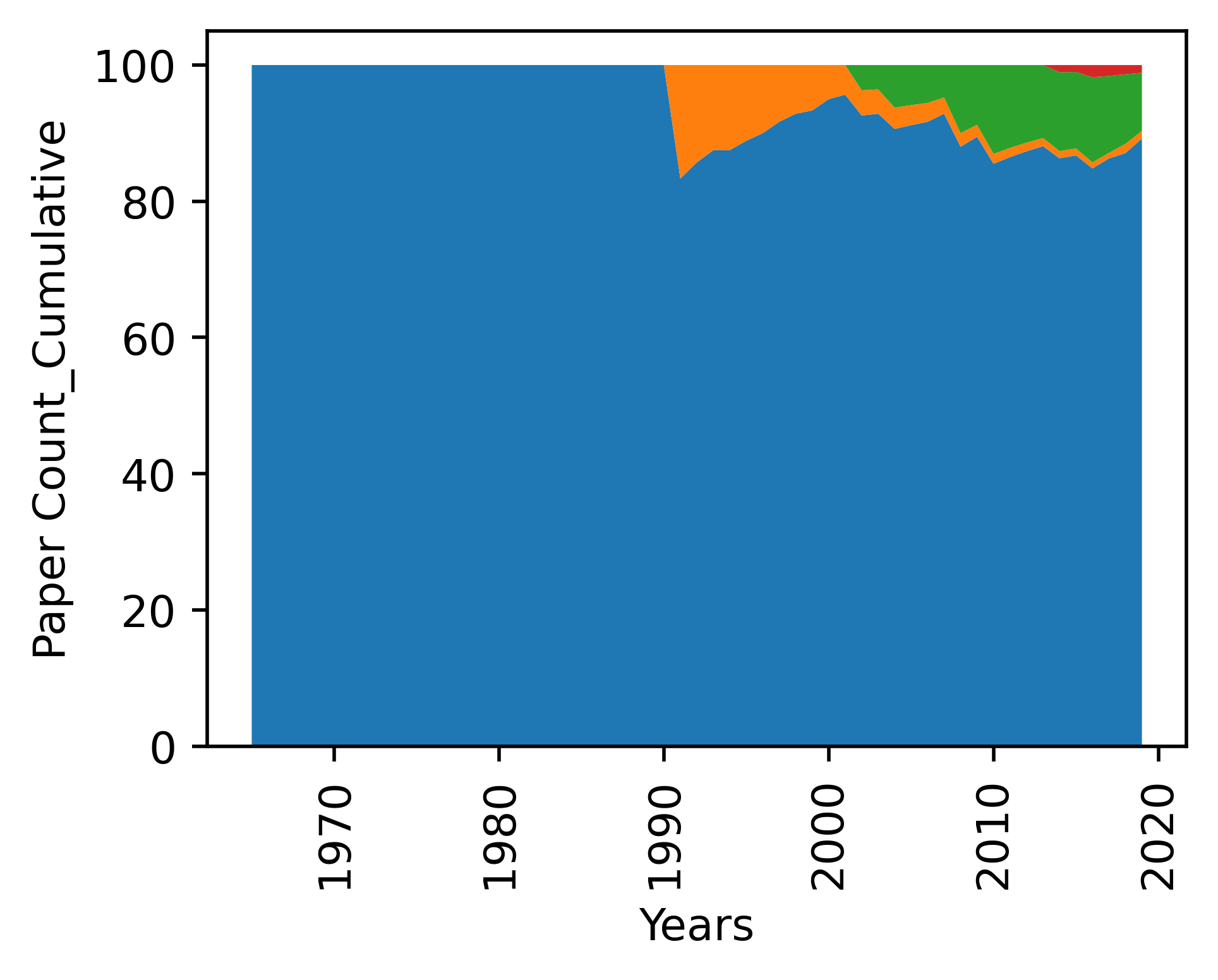}
         \caption{Small-Extinct}
    \end{subfigure}
    
    \begin{subfigure}[!hbt]{\fourthSingle\textwidth}
         \centering         \includegraphics[width=\textwidth]{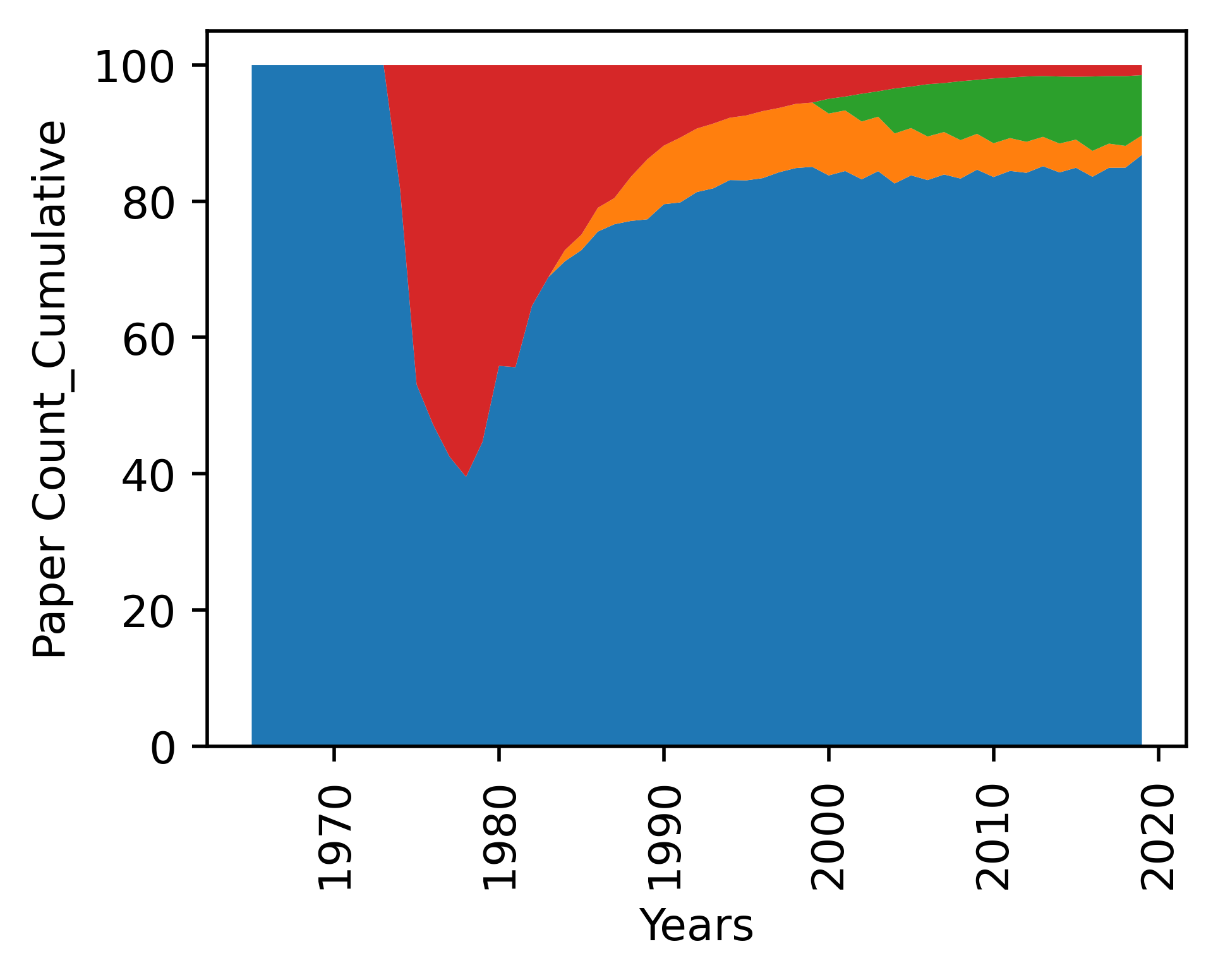}
         \caption{Mid-Institutional}
    \end{subfigure}
    \begin{subfigure}[!hbt]{\fourthSingle\textwidth}
         \centering
         \includegraphics[width=\textwidth]{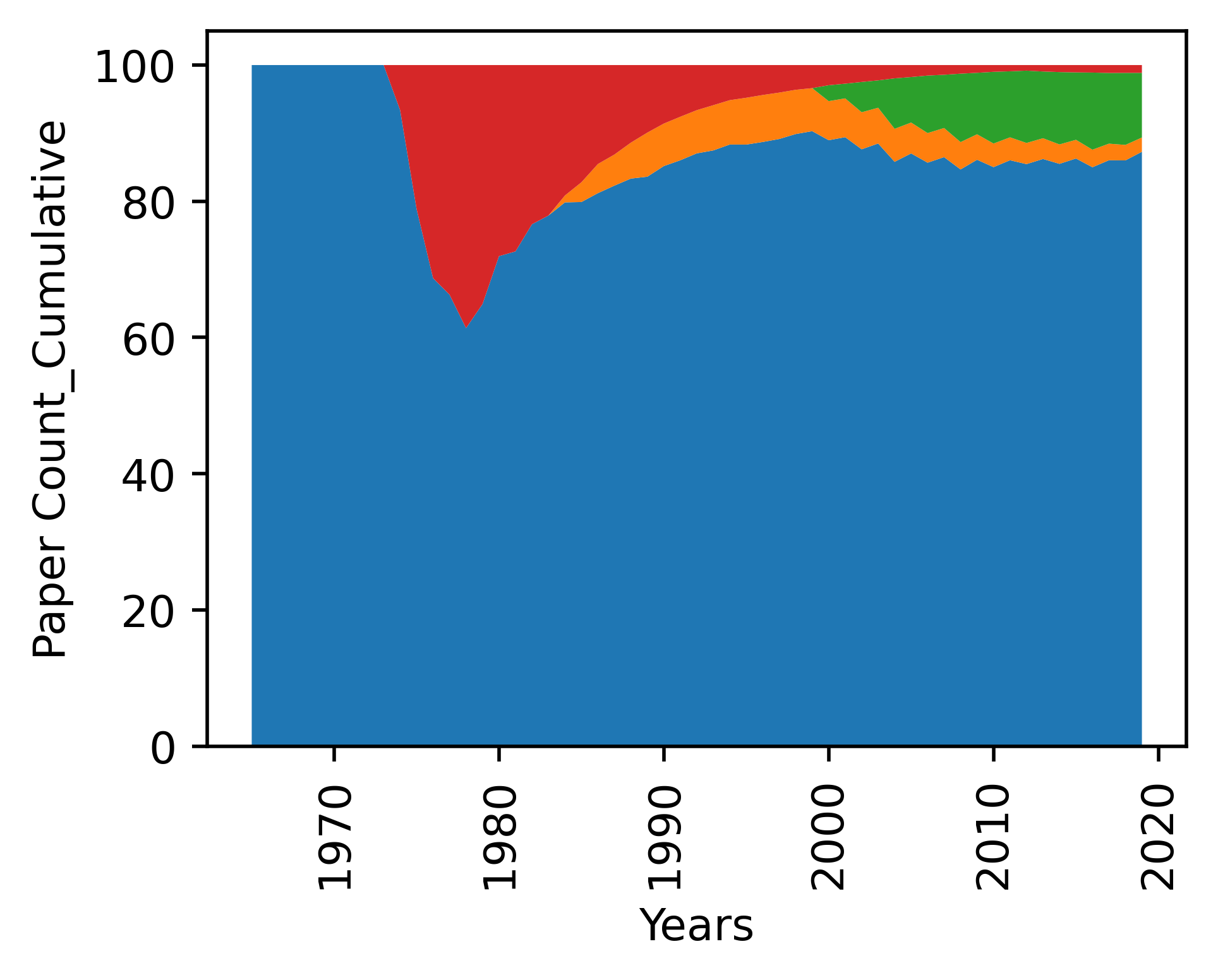}
         \caption{Mid-Stable}
    \end{subfigure}
    \begin{subfigure}[!hbt]{\fourthSingle\textwidth}
         \centering
         \includegraphics[width=\textwidth]{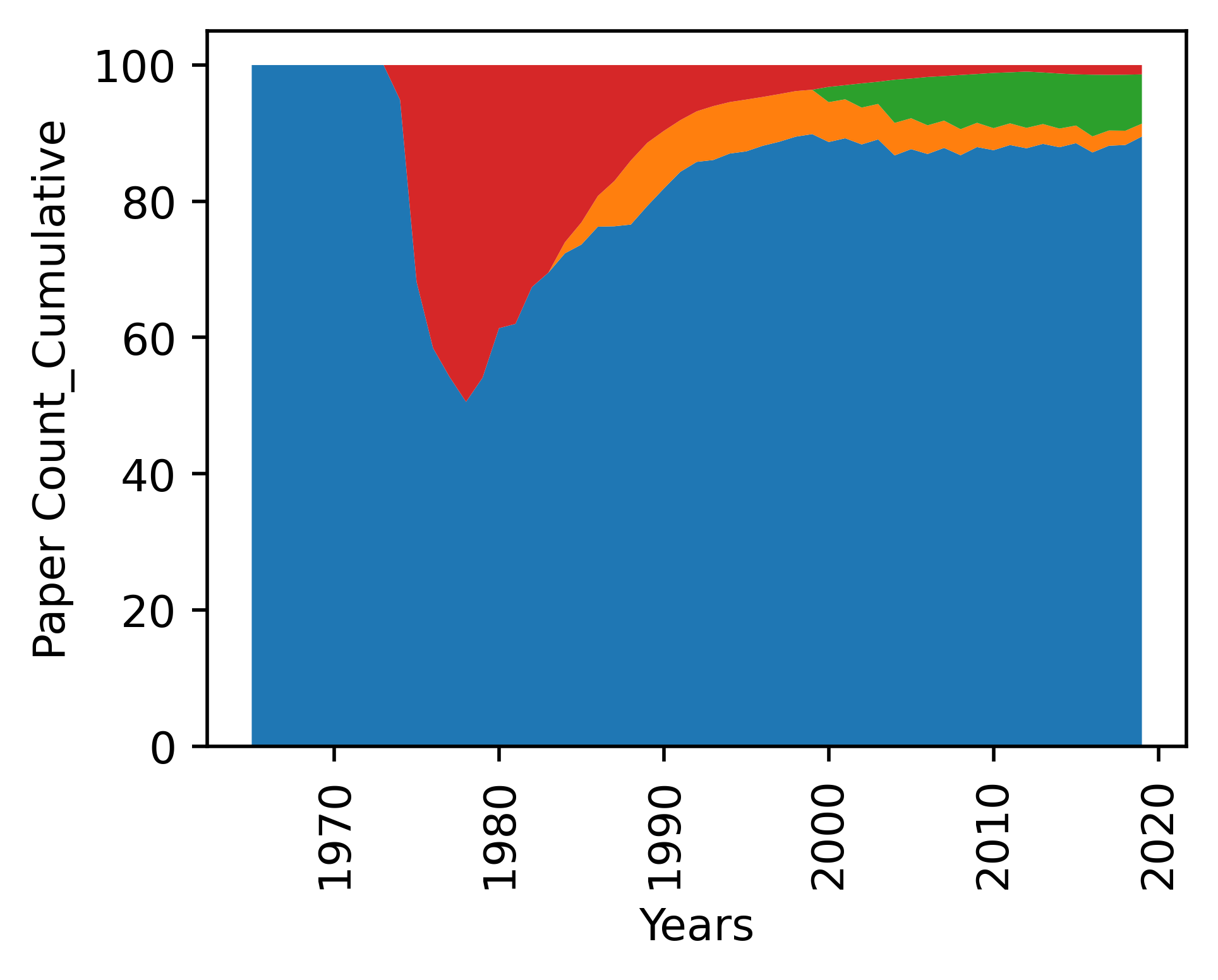}
         \caption{Mid-Endangered}
    \end{subfigure}
    
    \begin{subfigure}[!hbt]{\fourthSingle\textwidth}
         \centering
         \includegraphics[width=\textwidth]{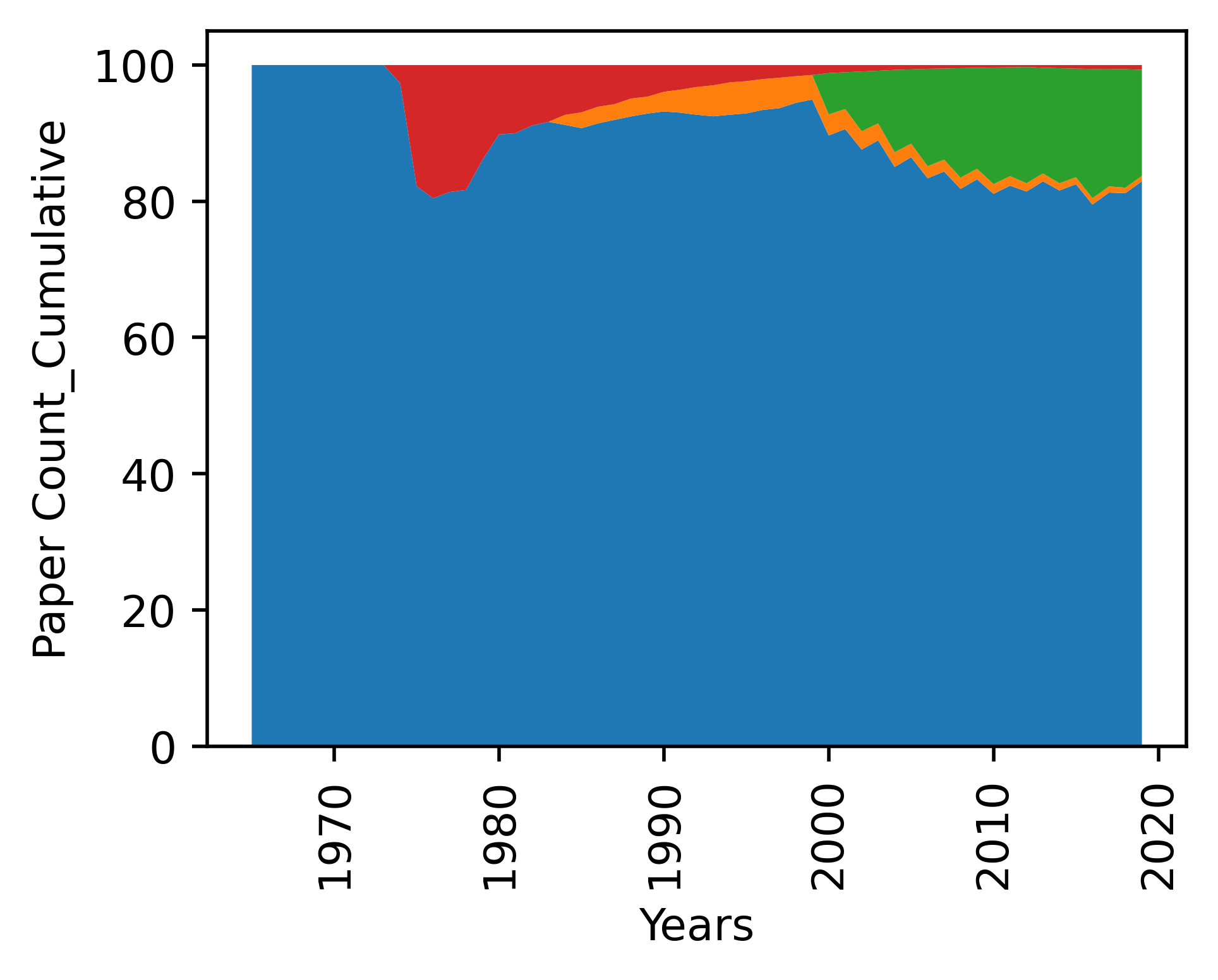}
         \caption{Large-Institutional}
    \end{subfigure}
    \begin{subfigure}[!hbt]{\fourthSingle\textwidth}
         \centering
         \includegraphics[width=\textwidth]{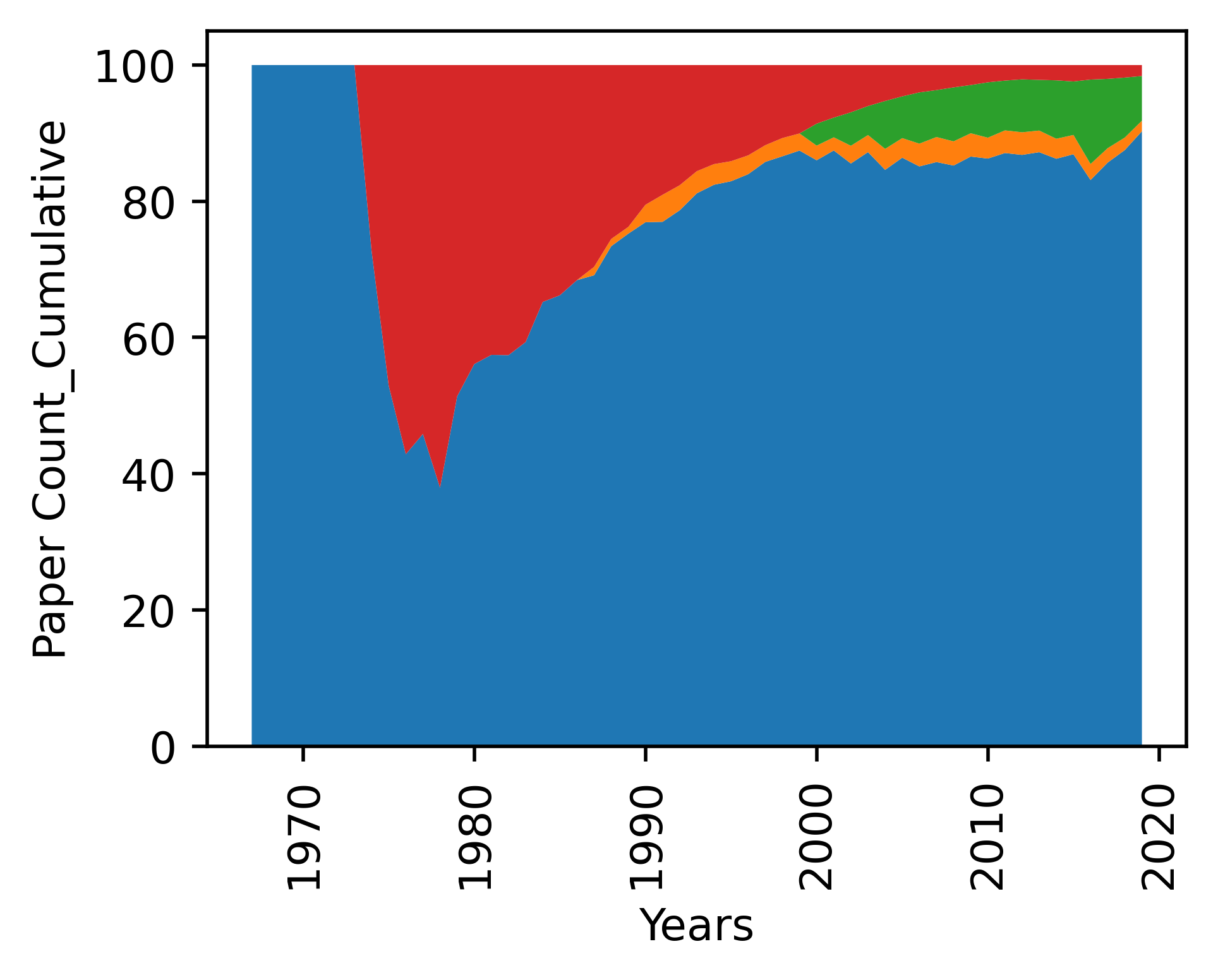}
         \caption{Large-Stable}
    \end{subfigure}
    \begin{subfigure}[!hbt]{\fourthSingle\textwidth}
         \centering
         \includegraphics[width=\textwidth]{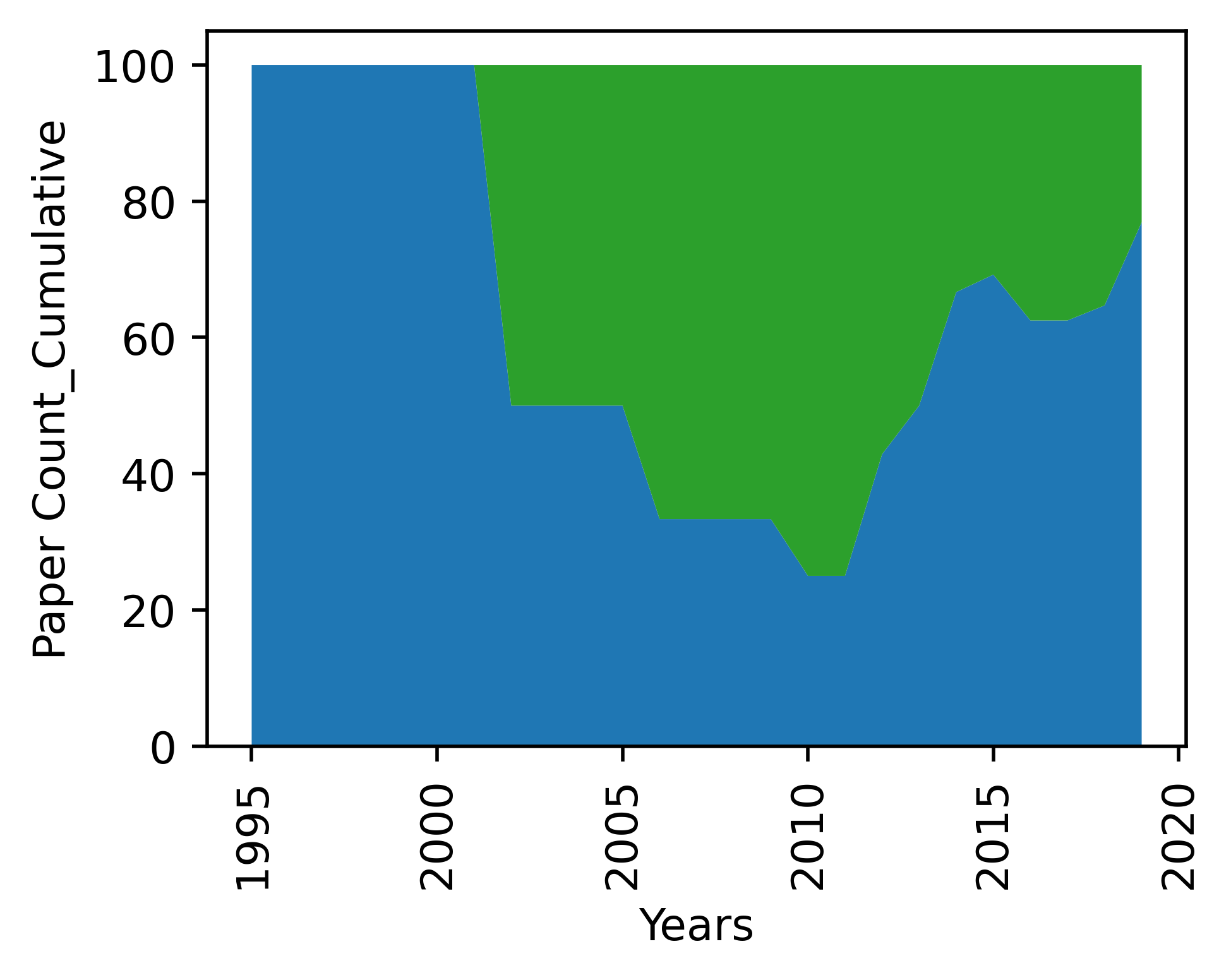}
         \caption{Large-Endangered}
         \label{fig:ACLpubs:LaEn}
    \end{subfigure}
    \caption{ACL Abstract Participation of the languages belonging to the 12 Ethnologue language classes (Only the existing 10 classes shown here.)}
    \label{fig:ACLpubs}
\end{figure*}

Figure~\ref{fig:ACLpubs} shows a breakdown of mentions in the abstracts of ACL Anthology publications. Here, \textit{Main} venues include (1) Annual Meeting of the Association for Computational Linguistics, (2) North American Chapter of the Association for Computational Linguistics, (3) European Chapter of the Association for Computational Linguistics, (4) Empirical Methods in Natural Language Processing, (5) International Conference on Computational Linguistics, (6) Conference on Computational Natural Language Learning (7)       International Workshop on Semantic Evaluation, (8) Conference of the Asia-Pacific Chapter of the Association for Computational Linguistics, and (9) Conference on Computational Natural Language Learning.
\textit{Journals} include (1) Transactions of the Association for Computational Linguistics and (2) Computational Linguistics. \textit{Other} category means everything except the aforementioned conferences/journals and \textit{LREC}. We have given \textit{LREC} a separate category as it is a venue where a considerable amount of researchers in under-resourced languages target. This decision is especially justified by the observations in Fig~\ref{fig:ACLpubs:LaEn}.
%\clearpage
It can be seen that despite the language category, most of the papers that mention a language name are in workshops. Interestingly, only \textit{LREC} and \textit{other} category has coverage for large-endangered languages.

%\begin{figure}[!hbt]
%     \centering
%    \includegraphics[width=\half\textwidth]{images/bubbleplot_BlasiCounts.png}
%    \caption{The 12 Ethnologue language classes where the size of each outer circle corresponds to the number of languages in that category and the size of each red circle corresponds to the coverage of ACL in the manner of~\citet{blasi-etal-2022-systematic}.}
 %   \label{fig:ACLallBlasi}
%\end{figure}

\section{Analysis on Where NLP Researchers Publish their Datasets }
\label{anthology_lrec-analysis}
\subsection{How the Analysis was Carried out}
We first checked the Dataset section of each paper. If the paper has used a dataset, we recorded whether it is a new dataset presented in the paper. If so, we check whether the dataset has been published. We mainly checked the Abstract, Introduction Dataset and Conclusion sections to see if information related to dataset publishing has been given. If not, we do a search using keywords such as data, corpus, publicly, share, release, free and available. This analysis was manually carried out.

\subsection{Dataset Publication Details}
As mentioned above, we first identify whether a paper has created a new dataset. Then we note down whether the dataset has been released in any of the following forms:
\begin{itemize}
    \item Via personal repository (github, personal web page, Google drive, etc)
    \item Via institutional repository (github, institutional website, etc). We also note whether the dataset is available freely or based on request. In some papers, this is clearly mentioned. For others, we visited the corresponding website and checked.
    \item via a public repository (ELRA, LDC, HuggingFace, CLARIN, etc)
\end{itemize}
If a link to any of the above has not been given, or if the paper explicitly mentions that the dataset cannot be released, we consider the dataset not released. Results are shown in Figure~\ref{fig:data_release}.

\iffalse
\begin{figure}[!hbt]
     \centering
      
     \begin{subfigure}[!hbt]{\half\textwidth}
         \centering      \includegraphics[width=\textwidth]{Figures/lrec_papers.png}
         \caption{Analysis based on LREC papers}       \label{fig:lrec_papers}
    \end{subfigure}
    
    \begin{subfigure}[!hbt]{\half\textwidth}
         \centering
         \includegraphics[width=\textwidth]{Figures/anthology_papers.png}
         \caption{Analysis based on ACL Anthology papers}
         \label{fig:anthology_papers} 
    \end{subfigure}
    \caption{Information of the use and release of data used in NLP research papers}
    \label{fig:data_release}
\end{figure}
\fi

\begin{figure*}[!htb]
     \centering
      % https://sankeymatic.com/build/
     \begin{subfigure}[!hbt]{\textwidth}
         \centering      \includegraphics[width=\textwidth]{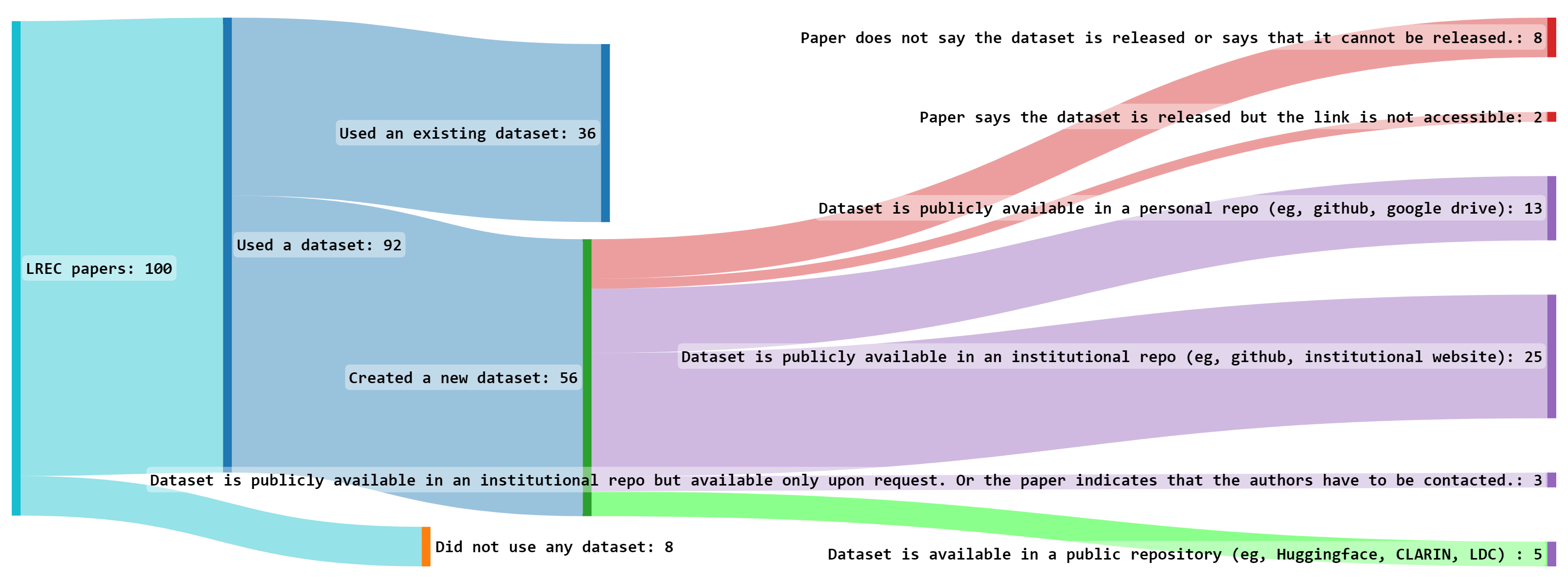}
         \caption{Analysis based on LREC papers}       \label{fig:lrec_papers}
    \end{subfigure}
    
    \begin{subfigure}[!hbt]{\textwidth}
         \centering
         \includegraphics[width=\textwidth]{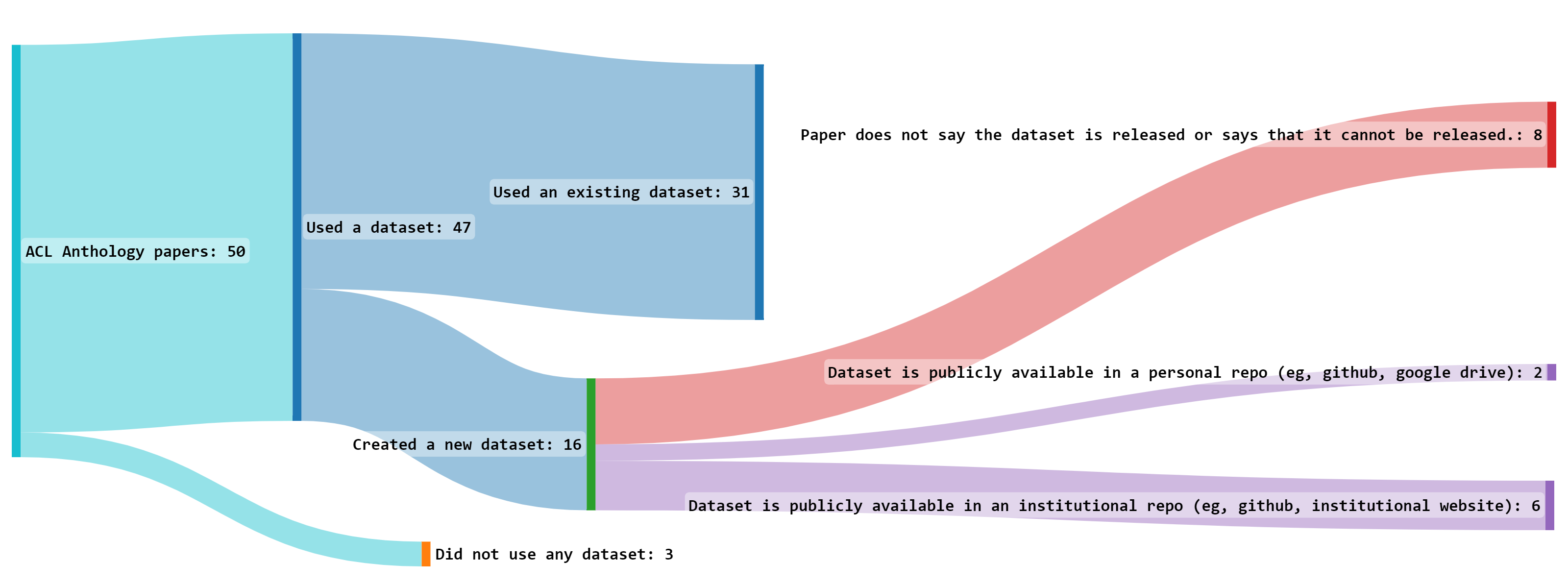}
         \caption{Analysis based on ACL Anthology papers}
         \label{fig:anthology_papers} 
    \end{subfigure}
    \caption{Information of the use and release of data used in NLP research papers}
    \label{fig:data_release}
\end{figure*}

%\FloatBarrier

\section{Survey Results}
\label{survey_results}
\begin{figure}[!htb]
    \centering
    \includegraphics[width=0.45\textwidth]{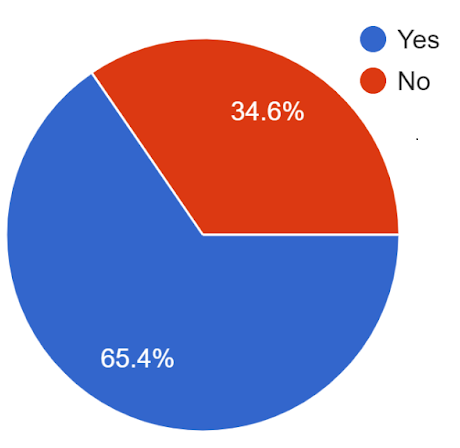}
    \caption{Distribution of researchers who published and did not publish data}
    \label{fig:NotPub}
\end{figure}

\begin{figure}[!htb]
    \centering
    \includegraphics[width=0.48\textwidth]{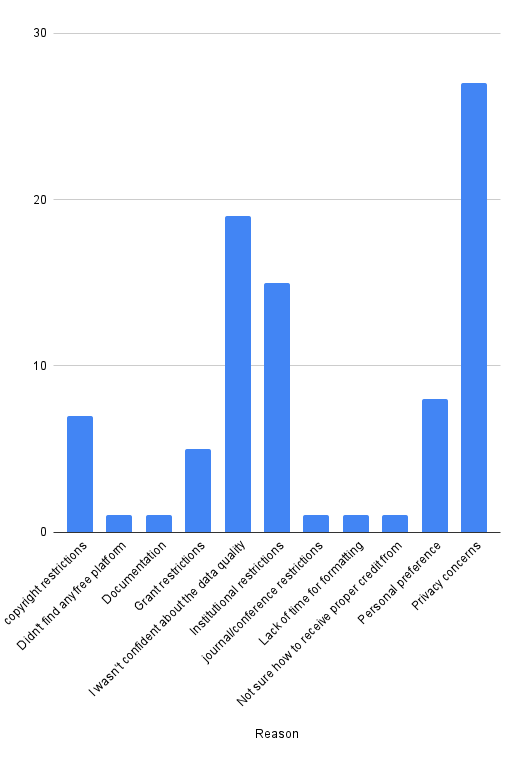}
    \caption{Reasons for not publishing the data}
    \label{fig:NotPubResons}
\end{figure}

\begin{figure}[!htb]
    \centering
    \includegraphics[width=0.45\textwidth]{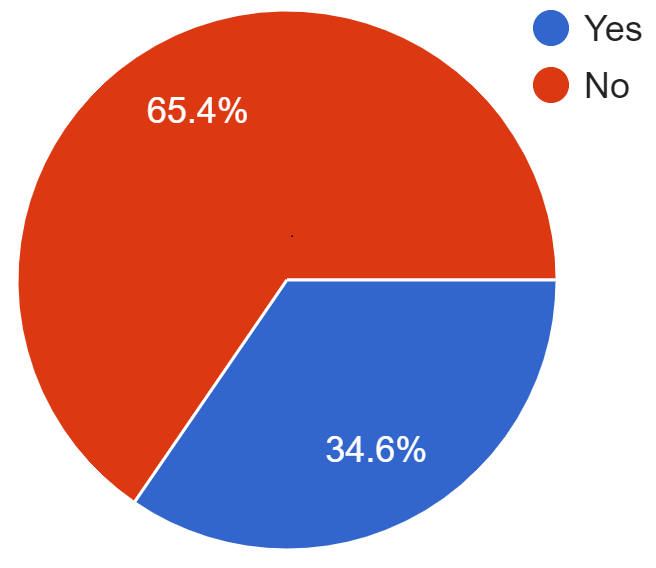}
    \caption{Openly released vs conditionally released}
    \label{fig:Open}
\end{figure}

\begin{figure}[!htb]
    \centering
    \includegraphics[width=0.48\textwidth]{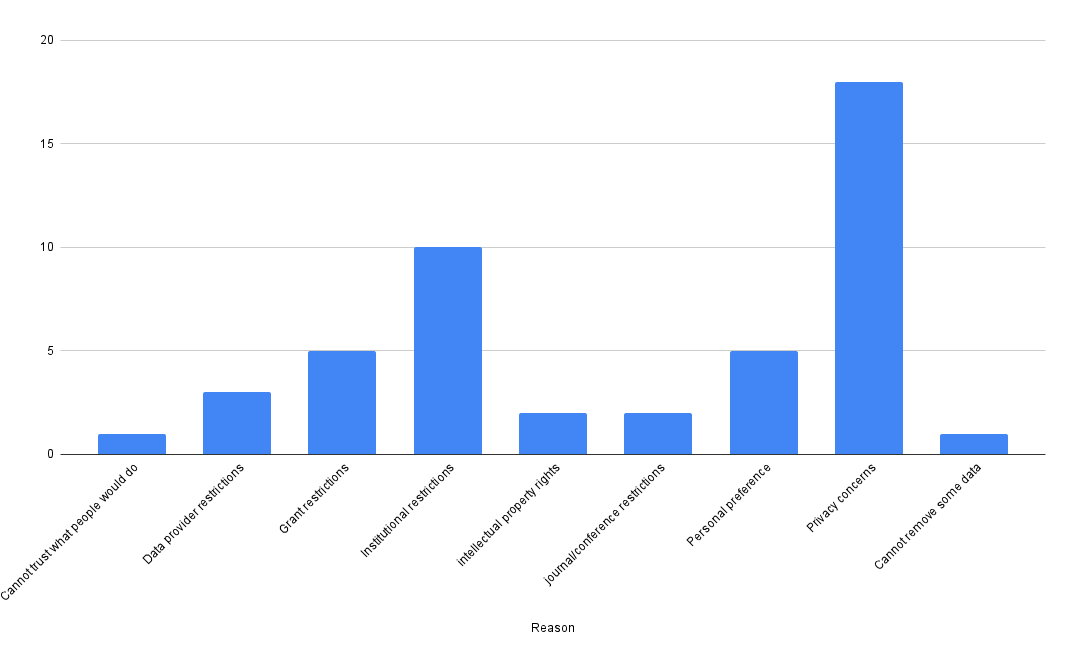}
    \caption{Reasons for conditionally releasing data}
    \label{fig:conditional}
\end{figure}

\begin{figure}[!htb]
    \centering
    \includegraphics[width=0.45\textwidth]{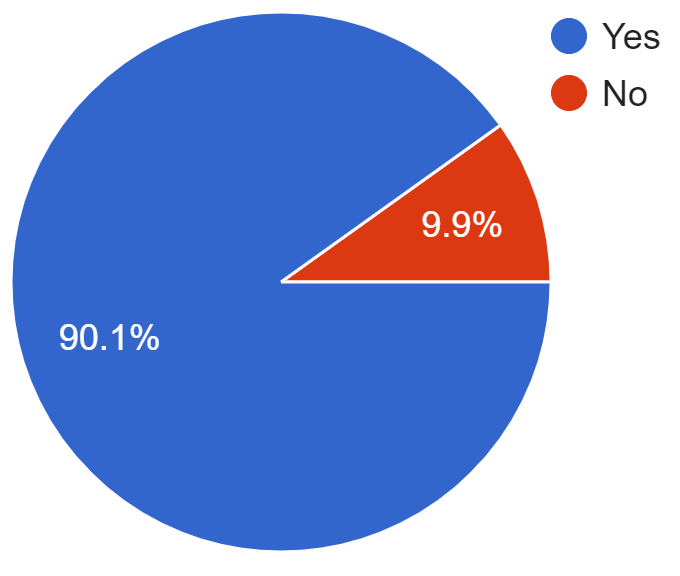}
    \caption{Distribution of researchers who made their data freely available}
    \label{fig:freelyAva}
\end{figure}

\begin{figure}[!htb]
    \centering
    \includegraphics[width=0.48\textwidth]{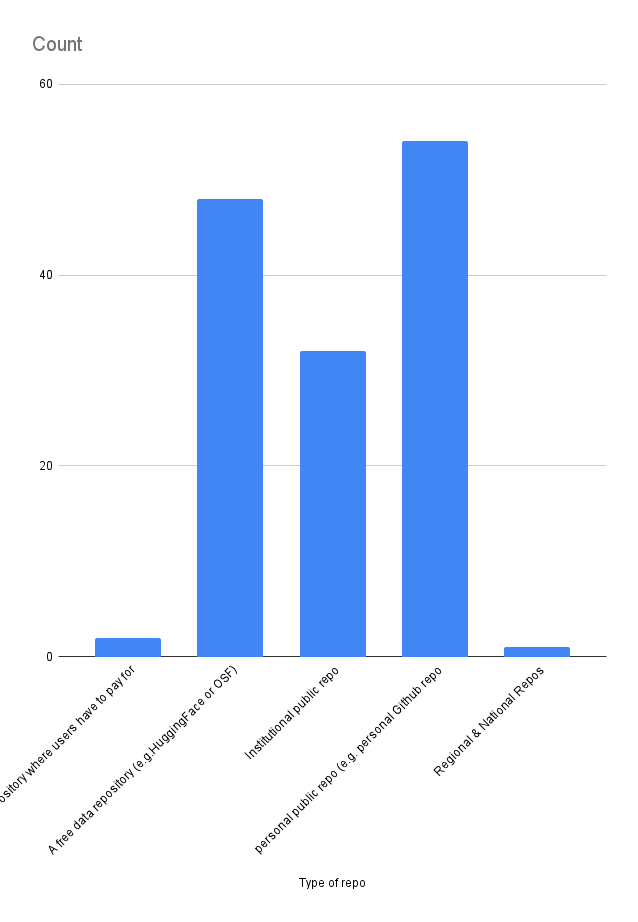}
    \caption{How datasets are publicly released}
    \label{fig:HowPub}
\end{figure}

\begin{figure}[!htb]
    \centering
    \includegraphics[width=0.48\textwidth]{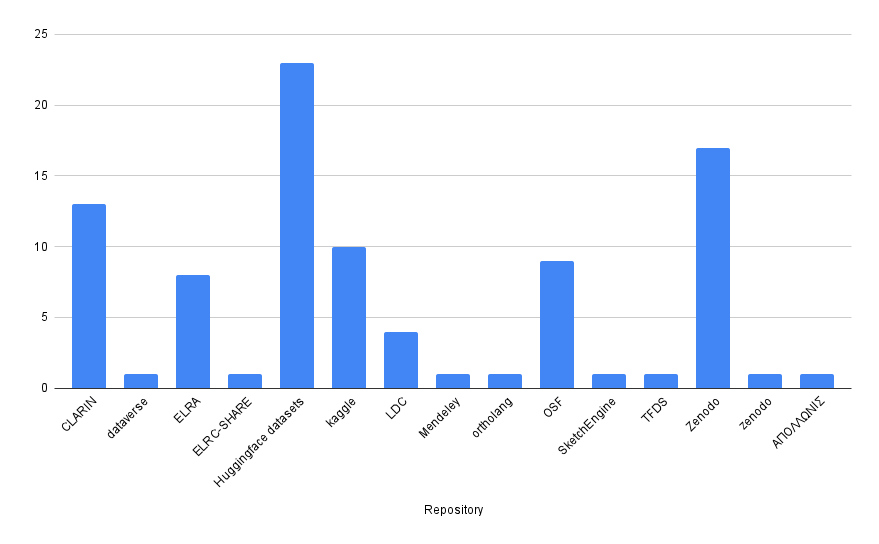}
    \caption{Where datasets are published}
    \label{fig:whePub}
\end{figure}

\begin{figure*}[!hbt]   
\centering
     \begin{subfigure}[!hbt]{\textwidth}
         \centering
         \includegraphics[width=\textwidth]{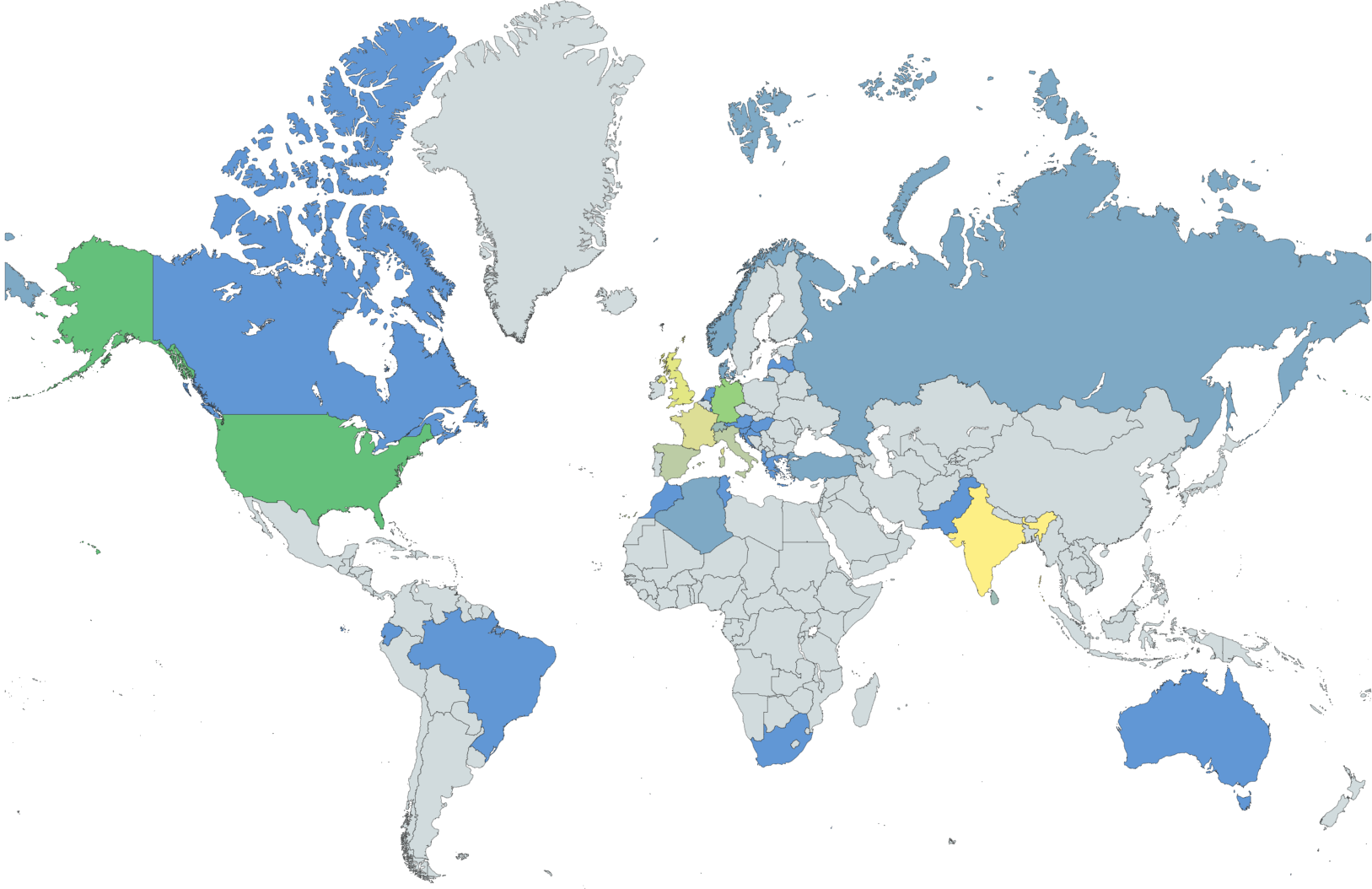}
    \end{subfigure}
    
    \begin{subfigure}[!hbt]{0.2\textwidth}
    \tiny
         \begin{tabular}{lrllrl}
\textbf{}             & \textbf{} & \textbf{}                &                &   &                          \\
United   States (USA) & 12        & \cellcolor[HTML]{63BE7B} & Australia      & 1 & \cellcolor[HTML]{5B9BD5} \\
Germany               & 10        & \cellcolor[HTML]{98CE7F} & Austria        & 1 & \cellcolor[HTML]{5B9BD5} \\
United   Kingdom      & 7         & \cellcolor[HTML]{E5E483} & Brazil         & 1 & \cellcolor[HTML]{5B9BD5} \\
India                 & 6         & \cellcolor[HTML]{FFEB84} & Canada         & 1 & \cellcolor[HTML]{5B9BD5} \\
France                & 5         & \cellcolor[HTML]{DEDB95} & Croatia        & 1 & \cellcolor[HTML]{5B9BD5} \\
Italy                 & 4         & \cellcolor[HTML]{BDCBA5} & Ecuador        & 1 & \cellcolor[HTML]{5B9BD5} \\
Spain                 & 4         & \cellcolor[HTML]{BDCBA5} & Greece         & 1 & \cellcolor[HTML]{5B9BD5} \\
Sri   Lanka           & 3         & \cellcolor[HTML]{9CBBB5} & Hungary        & 1 & \cellcolor[HTML]{5B9BD5} \\
Switzerland           & 3         & \cellcolor[HTML]{9CBBB5} & Latvia         & 1 & \cellcolor[HTML]{5B9BD5} \\
Algeria               & 2         & \cellcolor[HTML]{7BABC5} & Luxembourg     & 1 & \cellcolor[HTML]{5B9BD5} \\
Denmark               & 2         & \cellcolor[HTML]{7BABC5} & Morocco        & 1 & \cellcolor[HTML]{5B9BD5} \\
Norway                & 2         & \cellcolor[HTML]{7BABC5} & Netherlands    & 1 & \cellcolor[HTML]{5B9BD5} \\
Russia                & 2         & \cellcolor[HTML]{7BABC5} & Pakistan       & 1 & \cellcolor[HTML]{5B9BD5} \\
Turkey                & 2         & \cellcolor[HTML]{7BABC5} & Slovenia       & 1 & \cellcolor[HTML]{5B9BD5} \\
Albania               & 1         & \cellcolor[HTML]{5B9BD5} & South   Africa & 1 & \cellcolor[HTML]{5B9BD5} \\
                      &           &  & Tunisia        & 1 & \cellcolor[HTML]{5B9BD5}
\end{tabular}
    \end{subfigure}
    \caption{Countries at which the researchers who have uploaded their data sets have conducted their research}
    \label{fig:WorldMap}
\end{figure*}

%The survey questions are given in Table~\ref{tab:survey_question} and the analysis to the survey is shown in Figure~\ref{fig:survey_results}.
Given below are the survey questions that we have used:
\begin{enumerate}
    \item Have you ever kept a dataset you created ONLY in a private repo? Please select the most appropriate answer. (Results in Fig~\ref{fig:NotPub})
    \item If your answer was `yes' to the above question, please select all that applies. (Results in Fig~\ref{fig:NotPubResons})
    \item Have you ever made your dataset conditionally available? (e.g. signing NDA, expected a request to release data). Please  select the most appropriate answer.(Results in Fig~\ref{fig:Open})
    \item If your answer was `yes' to the above question,  please select all that applies. (Results in Fig~\ref{fig:conditional})
    \item Have you ever publicly made your dataset available? Please  select the most appropriate answer. (Results in Fig~\ref{fig:freelyAva})
    \item If yes, where did you publish your dataset?  Please select all that applies. (Results in Fig~\ref{fig:HowPub})
    \item If you have ever used a public repository (free or paid) to release data, what are they? select all that applies. (Results in Fig~\ref{fig:whePub})
    \item If you are not using data repositories such as  Huggingface, Kaggle and OSF, what are the reasons for that?   Please select all that applies (Results in Table~\ref{tab:ResonsForNot})
    \item Country that you are/were residing when you created most of your datasets (select the most relevant country) (Results in Fig~\ref{fig:WorldMap})
\end{enumerate}

Figure~\ref{fig:NotPub} shows a very positive trend - most researchers are releasing their dataset publicly. As per Figure~\ref{fig:NotPubResons}, the main reason for not publicly releasing the data is the privacy concerns. This is understandable, as text corpora deals with information written by/about people and organizations.  It is interesting to see that the second common reason for not releasing the dataset is the researcher not being confident about the dataset quality. This is a worrying situation, as the corresponding publication has already been made public and the claims in the paper may not be entirely correct. 

In their meta-study on parallel language data sets,~\citet{caswell2021quality} did observe that even the publicly available datasets have various quality issues. In that light, when these two ideas are put together, the conclusions we can draw here become more dire. If we are to hypothesise that the datasets that are released by the researchers that were confident of their data sets, and studies such as~\citet{caswell2021quality} find them lacking of quality, the work where the researchers themselves were not confident of the releasing data may be of highly questionable result. 
%If the quality of the released datasets are of unacceptable quality as~\citet{caswell2021quality} observes and the majority of the researchers opt to not-relase the dat for  
%
It is also worth encouraging researchers to publicly release their datasets, because some seem not to release the datasets just out of personal preference. 

Conditionally releasing the datasets also has a similar trend (see Figure~\ref{fig:Open}). Figure~\ref{fig:conditional} indicates that the reasons for conditionally releasing the datasets follows a similar trend to that of not releasing datasets. Institutional restrictions is also notable. We believe this is due to the institution investing in the dataset, or the dataset adding a competitive advantage to the institution.~\citet{de2021survey} also criticised the institutional barriers as a major reason for Sinhala NLP tools and data sets are not publicly shared. Our survey results in Figure~\ref{fig:conditional} re-affirms this observation but in a more generic manner, by the self-admission of NLP researchers on a wide range of languages.

Figure~\ref{fig:freelyAva} paints a very promising picture - about 90\% of the researchers have made their data publicly available at some point of time. What varies is how they publish their datasets. According to Figure~\ref{fig:HowPub}, most of the researchers still prefer to release their datasets via their personal repository (e.g.~Github repository of GoogleDrive). A considerable number released their datasets via their institutional repository, which could be due to institutional policies. It is worth noting that although it is lesser than those who release their data via their personal repositories, a decent number of researchers release their data via public repositories as well. This has a contradiction to what we found out by analysing LREC submissions, where only 9\% of the papers have indicated that the dataset has been released via a public repository. We suspect that this is due to the researchers adding their datasets first to their personal repository, and then to the public repository after publishing their paper.

The next noticeable fact is number of options that are available to publicly release a dataset (see Figure~\ref{fig:whePub}). Out of the 15 possible repositories, HuggingFace has been the most famous choice- this justifies our selection of the same to explain the impact of data repository in determining the resourcefulness of a language. The other famous repositories are Zenodo, CLARIN, Kaggle and OSF (in the given order). Interestingly, ELRA and LDC, the two repositories selected by~\citet{joshi-etal-2020-state} are further down in the preference list. 

In Table~\ref{tab:ResonsForNot}, we identify the reasons for researchers to not use the public repositories. It is surprising to see that there are several researchers who have not heard of such data repositories. A look into the individual responses did not indicate that these researchers belong to any particular geographical region. Given that there are 21 researchers who indicated that they cannot be bothered about adding data to public repositories, more awareness on the benefits of using public repositories should be carried out. Furthermore, availability of a repository that mitigates the limitations of the existing repositories would be a catalyst to encourage researchers. 

\begin{table}[!htb]
    \centering
    \tiny
    \begin{tabularx}{0.5\textwidth}{|X|r|}
    \hline
    Reason &  \makecell{Response\\Count} \\
    \hline
    \hline
    Accessing data through such repositories is difficult &	5 \\
    \hline
Control: it's easy to modify if it's personal/institute &	1  \\
\hline
Data was already released via my personal/institutional repo. so I could not be bothered to publish into another repo &	21 \\
\hline
Repository is maintained by a private company interested in Machine Learning &	2 \\
\hline
I do not trust those repositories would last long &	5 \\
\hline
Some repositories do not issue  DOI &	1 \\
\hline
I was not aware of such free data repositories &	13 \\
\hline
Such repos store older versions of datasets	& 1 \\
\hline
Too many different repositories. Unsure where the data will be found by other researchers &	1 \\
\hline
    \end{tabularx}
    \caption{Reasons for not using public repositories}
    \label{tab:ResonsForNot}
\end{table}

\iffalse
\begin{figure}[!htb]
    \centering
    \includegraphics[width=0.48\textwidth]{Figures/survey_7.png}
    \caption{Reasons for not using public repositories}
    \label{fig:ResonsForNot}
\end{figure}
\fi

Similarly, on the other end, these replies may also help those organisations and non-profits who maintain public repositories to augment the way they approach researchers to utilise their services. Specifically note the complaint of accessing data through such repositories being difficult. This could be taken as a call to improve the user interfaces and the overall experience of the repositories. The doubt of some researchers on how long the repositories would last is also an interesting point in this perspective. It seems given the choice between the institute of the researcher and a public repository run by a third party, some researchers are not confident of the continued existence of the repository. Thus this is a call for the repositories to inform the researchers of their policies on what happens to the hosted datasets upon a possible cessation of operations. Providing the researchers of such assurances about reliability, accessibility, and longevity may incentivise them to consider public data repositories in the future.

We show where each of the respondents of our survey marked as the country that they were residing when they created most of their datasets in Figure~\ref{fig:WorldMap}. It is unsurprising that the highest number of respondents are from the United States of America. The fact that personal contacts of the authors were also sent the survey explains the relative high number Sri Lanka has in the results. However the mot noticeable absentee is East Asia including China where a large portion of human population is concentrated and a considerable amount of language research is done. This might be an indication that researchers from these areas are under represented in the public mailing lists and private interest groups to which we sent our survey. We can postulate that one reason may be that aforementioned public mailing lists and private interest groups to which we sent out survey use English as the operational language. The researchers from East Asia (especially China) may use insular lists and groups that operate in the local language. This previously unforeseen divide may stand in the way of collaborations in the NLP field.

\section{Language Resource Increase Over Time}
\label{sec:WikiShift}

\begin{table*}[!hbt]
\centering
\renewcommand*{\arraystretch}{\tabComp}
\small
\begin{tabularx}{\textwidth}{|l|Z|ZZ|ZZ|ZZ|}
%\begin{tabular}{lrrrrrrr}
\hline
\multicolumn{2}{|c|}{Class} &  \multicolumn{2}{c|}{Nov 2021} &  \multicolumn{2}{c|}{Jul 2022} &  \multicolumn{2}{c|}{Difference} \\
\hline
Name &  Count &  Count &  Normalised &  Count &  Normalised &  Count &  Normalised\\
\hline
Small-Extinct       &    332 &           0 &             0.00 &       0 &         0.00 &     0 &       0.00 \\
Small-Endangered    &   2162 &          38 &             0.02 &      45 &         0.02 &     7 &       0.00 \\
Small-Stable        &   1168 &           1 &             0.00 &       3 &         0.00 &     2 &       0.00 \\
Small-Institutional &     28 &           0 &             0.00 &       0 &         0.00 &     0 &       0.00 \\
Mid-Extinct      &    0 &          0 &             0.00 &     0 &         0.00 &    0 &       0.00 \\
Mid-Endangered      &    458 &          86 &             0.19 &     101 &         0.22 &    15 &       0.03 \\
Mid-Stable          &   1700 &          24 &             0.01 &      34 &         0.02 &    10 &       0.01 \\
Mid-Institutional   &    208 &         228 &             1.10 &     310 &         1.49 &    82 &       0.39 \\
Large-Extinct      &    0 &          0 &             0.00 &     0 &         0.00 &    0 &       0.00 \\
Large-Endangered    &     14 &          27 &             1.93 &      31 &         2.21 &     4 &       0.29 \\
Large-Stable        &    133 &          51 &             0.38 &      76 &         0.57 &    25 &       0.19 \\
Large-Institutional &    217 &        3529 &            16.26 &    4140 &        19.08 &   611 &       2.82 \\
\hline
%\end{tabular}
\end{tabularx}
\caption{The number of datasets available in \textit{Huggingface} for the 12 Ethnologue language classes  in November 2021 compared with July 2022.}
\label{tab:HuggingShift}
\end{table*}

\begin{table*}[!hbt]
\centering
\renewcommand*{\arraystretch}{\tabComp}
\small
\begin{tabularx}{\textwidth}{|l|Z|ZZ|ZZ|ZZ|}
%\begin{tabular}{lrrrrrrr}
\hline
\multicolumn{2}{|c|}{Class} &  \multicolumn{2}{c|}{Nov 2021} &  \multicolumn{2}{c|}{Jul 2022} &  \multicolumn{2}{c|}{Difference} \\
\hline
Name &  Count &  Count &  Normalised &  Count &  Normalised &  Count &  Normalised\\
\hline
Small-Extinct       &    332 &                                    0 &                                      0.00 &      4176 &          12.58 &      4176 &      12.58 \\
Small-Endangered    &   2162 &                                 3849 &                                      1.78 &    180106 &          83.31 &    176257 &      81.52 \\
Small-Stable        &   1168 &                                 1036 &                                      0.89 &      2958 &           2.53 &      1922 &       1.65 \\
Small-Institutional &     28 &                                    0 &                                      0.00 &      2455 &          87.68 &      2455 &      87.68 \\
Mid-Extinct      &    0 &                                0 &                                     0.00 &    0 &        0.00 &    0 &    0.00 \\
Mid-Endangered      &    458 &                                18028 &                                     39.36 &    650903 &        1421.19 &    632875 &    1381.82 \\
Mid-Stable          &   1700 &                                 8903 &                                      5.24 &    171688 &         100.99 &    162785 &      95.76 \\
Mid-Institutional   &    208 &                               366882 &                                   1763.86 &   1058393 &        5088.43 &    691511 &    3324.57 \\
Large-Extinct      &    0 &                                0 &                                     0.00 &    0 &        0.00 &    0 &    0.00 \\
Large-Endangered    &     14 &                                    0 &                                      0.00 &     77070 &        5505.00 &     77070 &    5505.00 \\
Large-Stable        &    133 &                                22124 &                                    166.35 &   1085994 &        8165.37 &   1063870 &    7999.02 \\
Large-Institutional &    217 &                              1243317 &                                   5729.57 &  54612595 &      251670.94 &  53369278 &  245941.37 \\
\hline
%\end{tabular}
\end{tabularx}
\caption{The number of datasets available in \textit{Wikipedia} for the 12 Ethnologue language classes  in November 2021 compared with July 2022.}
\label{tab:WikiShift}
\end{table*}

Tables~\ref{tab:HuggingShift} and~\ref{tab:WikiShift} record the number of annotated and unannotated (respectively) dataset increase from November 2021 to July 2022. The \textit{Difference} column shows the growth in number and each of the normalised columns carries the value obtained by dividing the values in adjoining \textit{count} column by the the number in the \textit{count} column for the relevant class. Both tables show a similar trend, even after normalising to the class size - Large-institutional category has been added with more data. Similarly, the extinct languages seem to be forever forgotten. Annotated dataset count for Mid-institutional languages have increased by a noticeable number. On the other hand, focus on `small' languages is negligible, if not zero.  This trend of rich getting richer is a cause for concern for those who are interested in developing and using data sets to and from low-resourced languages as this shows that the average interest still lies with the few languages that are already enjoying an abundance of datasets.

In contrast, most categories show a growth in Wikipedia article counts. Particularly of interest is the mid-endangered category, which has a noticeable gain. This hints at some community efforts to increase the digital content for these languages that took place recently. As observed by~\citet{hoenen2021migration}, some members of the communities of endangered languages have taken to Wikipedia as a means of conserving traditional knowledge, and oral traditions in the source language.

\end{document}